\documentclass{article}

\pdfoutput=1

\PassOptionsToPackage{numbers, compress}{natbib}

\usepackage{xcolor}
\usepackage[preprint]{corl_2021} 
\usepackage{marginnote}
\usepackage{graphics}
\usepackage[pdftex]{graphicx}
\usepackage{wrapfig}
\DeclareGraphicsExtensions{.pdf,.png,.jpg}

\usepackage[font={small}]{caption}
\usepackage{subcaption}
\usepackage[rightcaption]{sidecap}
\usepackage{pbox}

\usepackage{mathtools}
\usepackage{amsmath, amssymb, amscd}
\usepackage{ wasysym } 
\usepackage{amsfonts}
\usepackage{mathptmx} 
\usepackage{gensymb} 
\usepackage{nicefrac}       

\DeclareMathAlphabet{\mathcal}{OMS}{lmsy}{m}{n}
\DeclareSymbolFont{largesymbols}{OMX}{cmex}{m}{n}
\usepackage{textcomp} 
\usepackage{dsfont}

\usepackage{array} 
\usepackage{tabularx}
\usepackage{multirow}
\usepackage{multicol}
\usepackage{booktabs}

\usepackage[
  separate-uncertainty = true,
  multi-part-units = repeat
]{siunitx}
\usepackage[T1]{fontenc} 
\usepackage[utf8]{inputenc}
\usepackage[english]{babel} 
\usepackage{units}
\usepackage{bm}
\usepackage{times} 
\usepackage{xspace}
\usepackage{flushend}
\usepackage{csquotes}
\usepackage{makeidx}

\usepackage{enumitem}

\usepackage{soul} 
\usepackage{subfiles} 

\usepackage[protrusion=true,expansion=true]{microtype}
\usepackage[yyyymmdd]{datetime}

\date{\protect\formatdate{1}{1}{2001}}

\usepackage{url}

\usepackage{soul} 

\newcommand{\tocite}[1]{%
\textcolor{red}{[cite:\ifthenelse{\equal{#1}{}}{}{#1}?]}
}

\newcommand{\ignore}[1]{}

\newcommand{\revision}[1]{#1}

\title{What Matters in Learning from Offline Human Demonstrations for Robot Manipulation}

\author{
  Ajay Mandlekar$^1$, 
  Danfei Xu$^1$, 
  Josiah Wong$^1$, 
  Soroush Nasiriany$^2$, 
  Chen Wang$^1$, \\ \\
\textbf{Rohun Kulkarni$^1$, 
  Li Fei-Fei$^1$, 
  Silvio Savarese$^1$, 
  Yuke Zhu$^2$, 
  Roberto Mart\'in-Mart\'in$^1$} \\ \\
  $^1$Stanford University, $^2$The University of Texas at Austin
}

\begin{document}
\maketitle


\begin{abstract}
Imitating human demonstrations is a promising approach to endow robots with various manipulation capabilities. 
While recent advances have been made in imitation learning and batch (offline) reinforcement learning, a lack of open-source human datasets and reproducible learning methods make assessing the state of the field difficult. 
In this paper, we conduct an extensive study of six offline learning algorithms for robot manipulation on five simulated and three real-world multi-stage manipulation tasks of varying complexity, and with datasets of varying quality.
Our study analyzes the most critical challenges when learning from offline human data for manipulation.
Based on the study, we derive a series of lessons including the sensitivity to different algorithmic design choices, the dependence on the quality of the demonstrations, and the variability based on the stopping criteria due to the different objectives in training and evaluation.
We also highlight opportunities for learning from human datasets, such as the ability to learn proficient policies on challenging, multi-stage tasks beyond the scope of current reinforcement learning methods, and the ability to easily scale to natural, real-world manipulation scenarios where only raw sensory signals are available.
We have open-sourced our datasets and all algorithm implementations to facilitate future research and fair comparisons in learning from human demonstration data. Codebase, datasets, trained models, and more available at \url{https://arise-initiative.github.io/robomimic-web/}

\end{abstract}



\section{Introduction}
\label{sec:intro}


Human supervision has been at the heart of the most significant recent advances in several domains such as computer vision~\cite{deng2009imagenet,krizhevsky2012imagenet,Redmon2016YouOL,Ronneberger2015UNetCN} and natural language processing~\cite{rajpurkar2018squad2,floridi2020gpt,Devlin2019BERTPO}. By intelligently extracting information from large-scale human-labeled datasets, autonomous machines have been able to reach near- or even super-human performance on decades-old problems such as image recognition and question answering. 
Roboticists have also attempted to tackle robot manipulation through learning from human datasets, using the paradigms of Imitation Learning~\cite{pomerleau1989alvinn, zhang2017deep, mandlekar2020learning} and Batch (Offline) Reinforcement Learning~\cite{lange2012batch, levine2020offline, cabi2019scaling}, where datasets consisting of robot arm trajectories, action labels at each timestep, and possibly reward labels, are used to train closed-loop policies.


As in other domains, large offline datasets offer several benefits such as scale, portability, and reproducible evaluations to measure progress.
Recently, there has been considerable progress in offline learning for robot manipulation from human demonstrations~\cite{zhang2017deep,florence2019self,mandlekar2020learning}. 
Despite these advances, the offline learning paradigm has not been nearly as disruptive in robotics as in other disciplines -- there is a large gap between autonomous robot manipulation capabilities and the wide range of tasks that humans can solve effortlessly using physical and cognitive intelligence. What has inhibited the use of large human-provided datasets to address this gap?



In contrast to other domains where supervised learning has been successful, robotic manipulation is a time-evolving dynamical system, requiring fine-grained real-time control to guide robot arms successfully through tasks -- consequently, data collection can present technical challenges requiring specialized systems~\cite{mandlekar2018roboturk}, which can explain why large-scale human-provided datasets~\cite{sharma2018multiple, mandlekar2019scaling} have not been very prevalent. Learning from such datasets can also present several challenges. Human demonstrations can differ from machine-generated datasets (a recent trend in benchmarks for offline policy learning~\cite{fu2020d4rl, gulcehre2020rl}) due to a non-Markovian decision process, since humans may not act purely based on the current observation. There can also be significant variance in both data quality and solution strategy when collecting data from multiple humans~\cite{mandlekar2020iris}. Differences from classic supervised learning, such as a mismatch between training and evaluation objectives (task success rate), can make selecting a final policy challenging~\cite{ross2011reduction, precup2000eligibility}, especially in real-world settings where evaluating each policy on a robot can be infeasible. Finally, offline learning is sensitive to state and action space coverage (dataset size) and agent design decisions.


Studying these challenges in the context of robot manipulation and human-provided datasets could be a stepping stone to closing the gap between robot and human manipulation capabilities. 
Unfortunately, a lack of suitable benchmark and human datasets have made studying this setting difficult. Prior works are either limited to studying simple 2D environments~\cite{toyer2020magical} or using data generated from hard-coded policies~\cite{james2020rlbench,zeng2020transporter}.
In this paper, we address this need by presenting a study of data-driven offline policy learning methods on several human-provided robot manipulation datasets. We collect task demonstrations from human teleoperators across a broad range of simulated and real world manipulation tasks and investigate several factors that play a role in learning from such data. 

\begin{figure}[t!]
\begin{subfigure}{0.20\linewidth}
\centering
\includegraphics[width=1.0\textwidth]{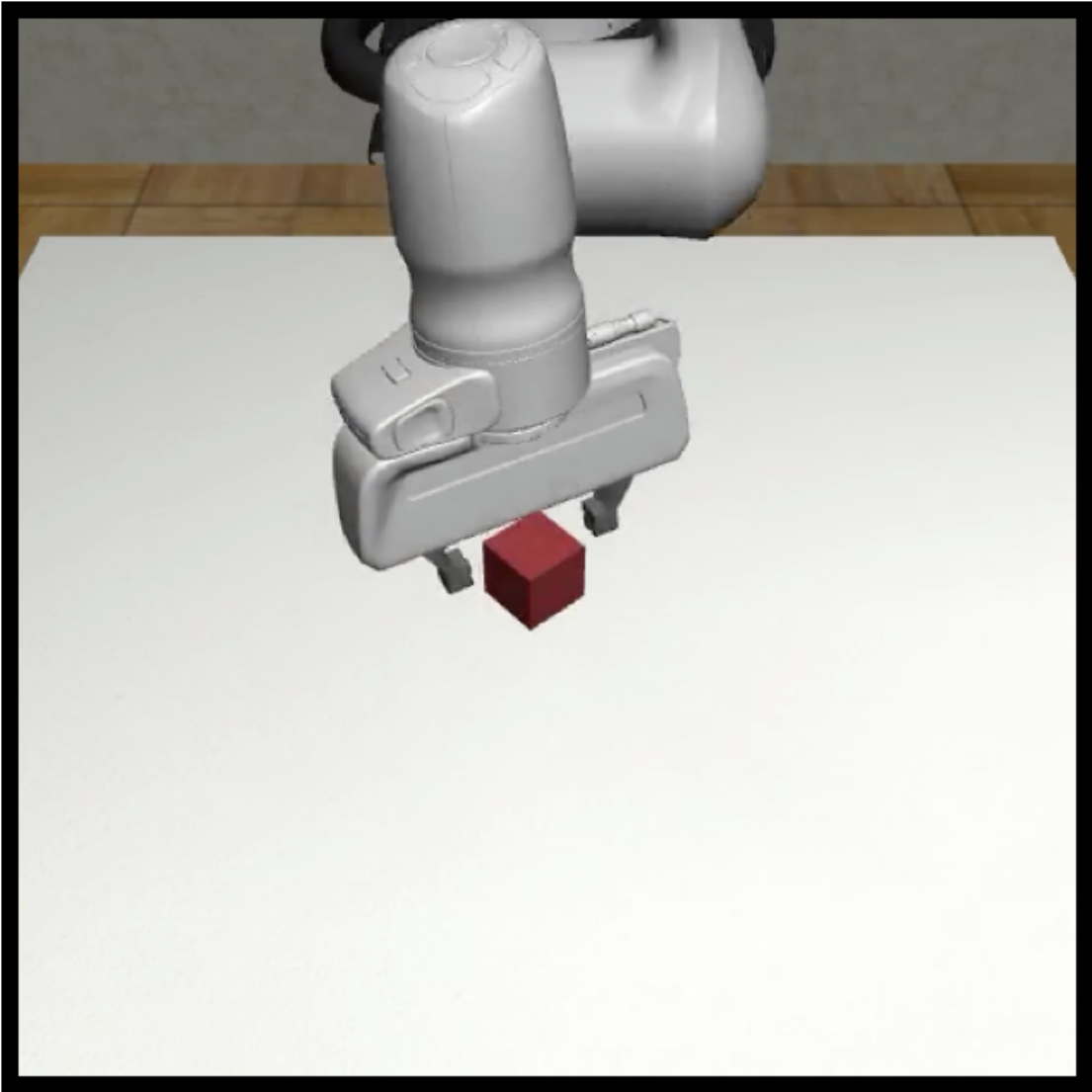}
\caption{Lift} 
\end{subfigure}
\hfill
\begin{subfigure}{0.20\linewidth}
\centering
\includegraphics[width=1.0\textwidth]{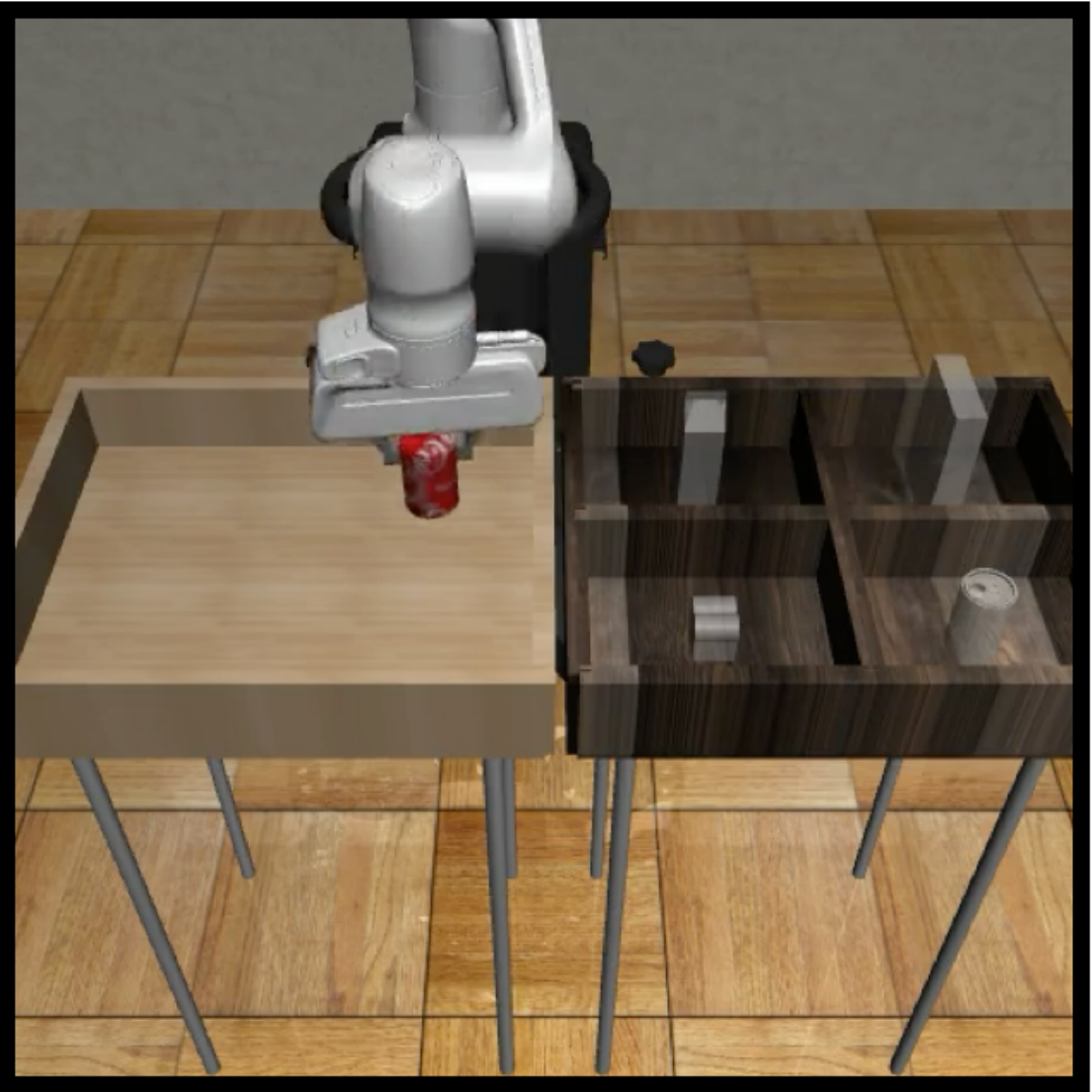}
\caption{Can} 
\end{subfigure}
\hfill
\begin{subfigure}{0.20\linewidth}
\centering
\includegraphics[width=1.0\textwidth]{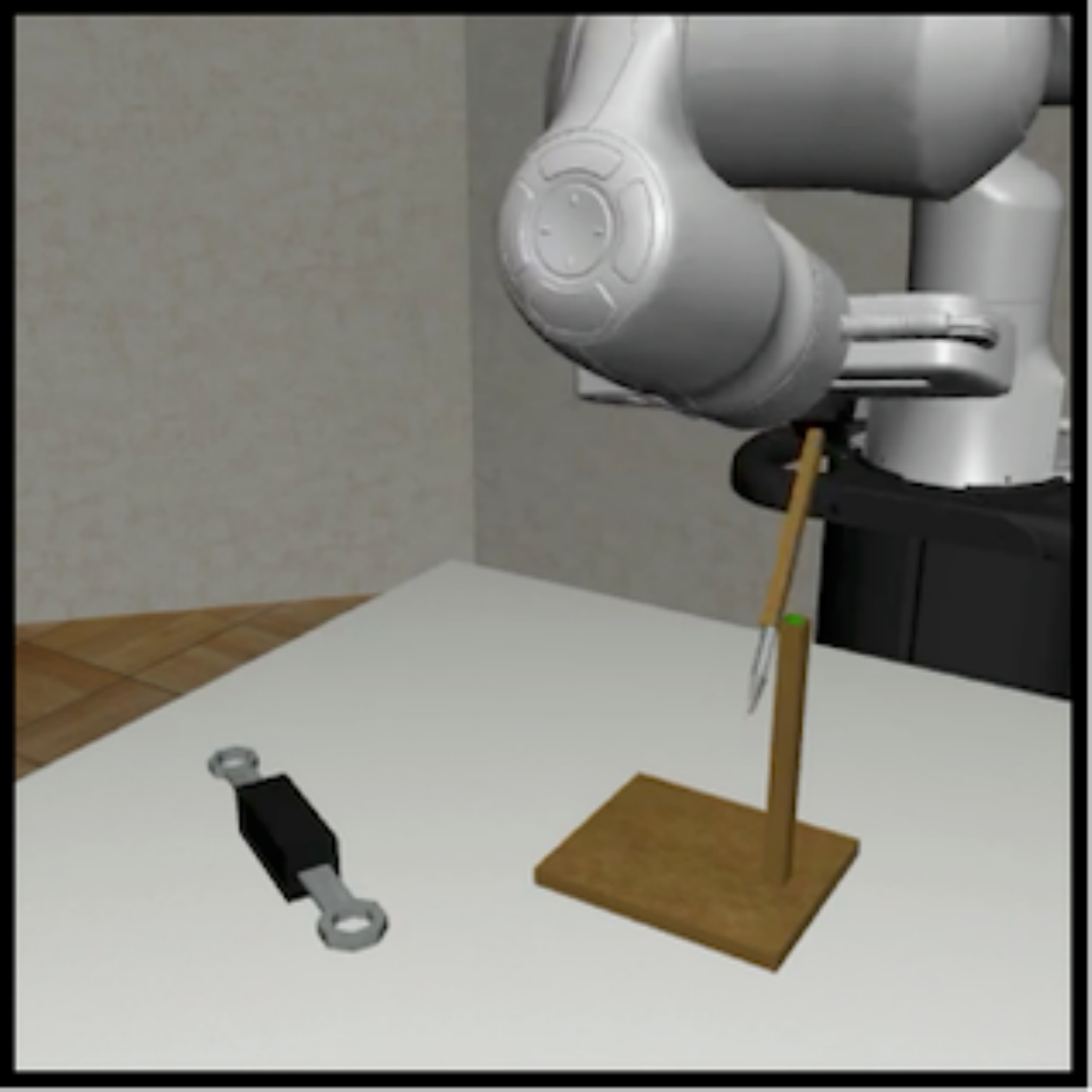}
\caption{Tool Hang} 
\end{subfigure}
\hfill
\begin{subfigure}{0.20\linewidth}
\centering
\includegraphics[width=1.0\textwidth]{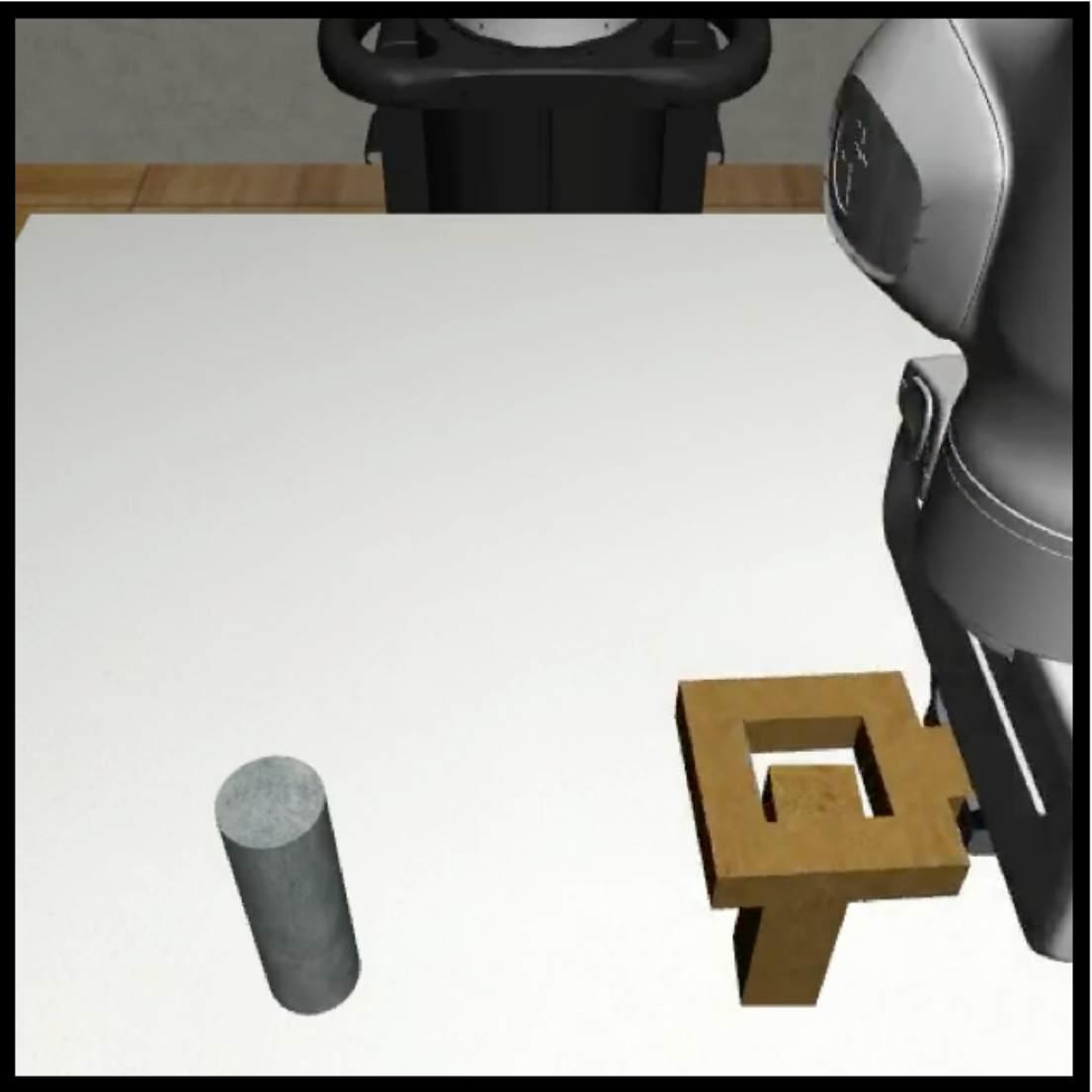}
\caption{Square} 
\end{subfigure}
\hfill
\begin{subfigure}{0.20\linewidth}
\centering
\includegraphics[width=1.0\textwidth]{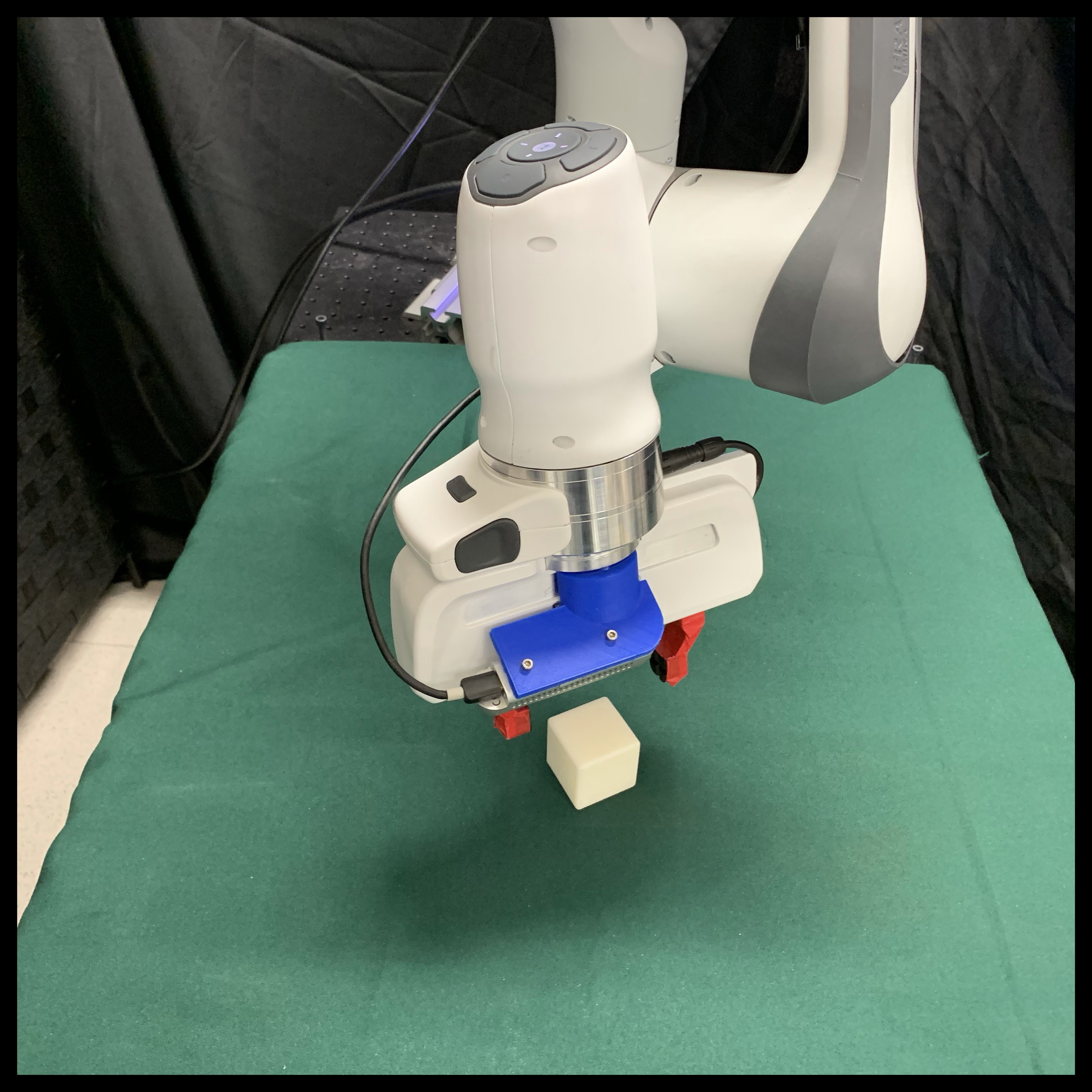}
\caption{Lift (Real)} 
\end{subfigure}
\hfill
\begin{subfigure}{0.20\linewidth}
\centering
\includegraphics[width=1.0\textwidth]{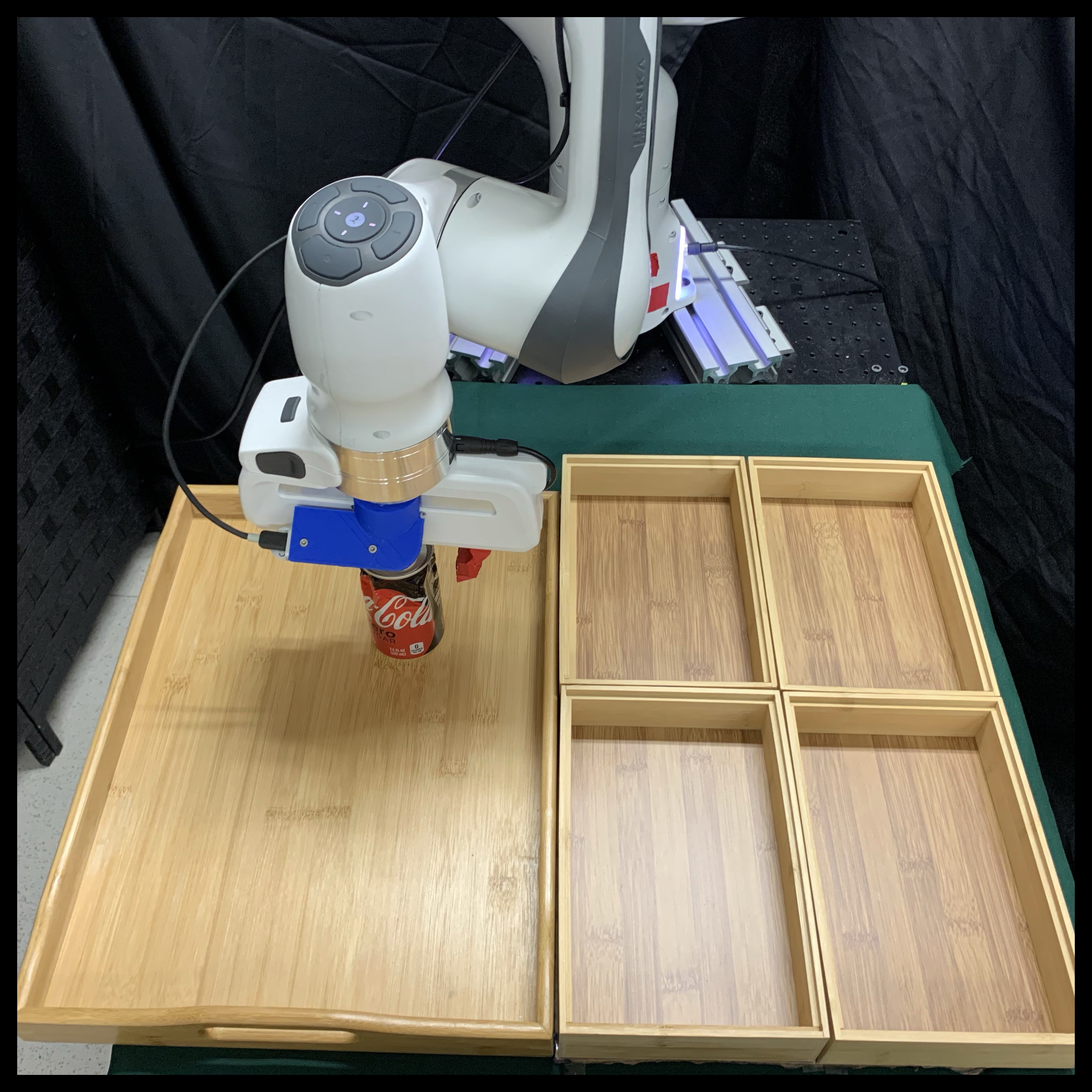}
\caption{Can (Real)} 
\end{subfigure}
\hfill
\begin{subfigure}{0.20\linewidth}
\centering
\includegraphics[width=1.0\textwidth]{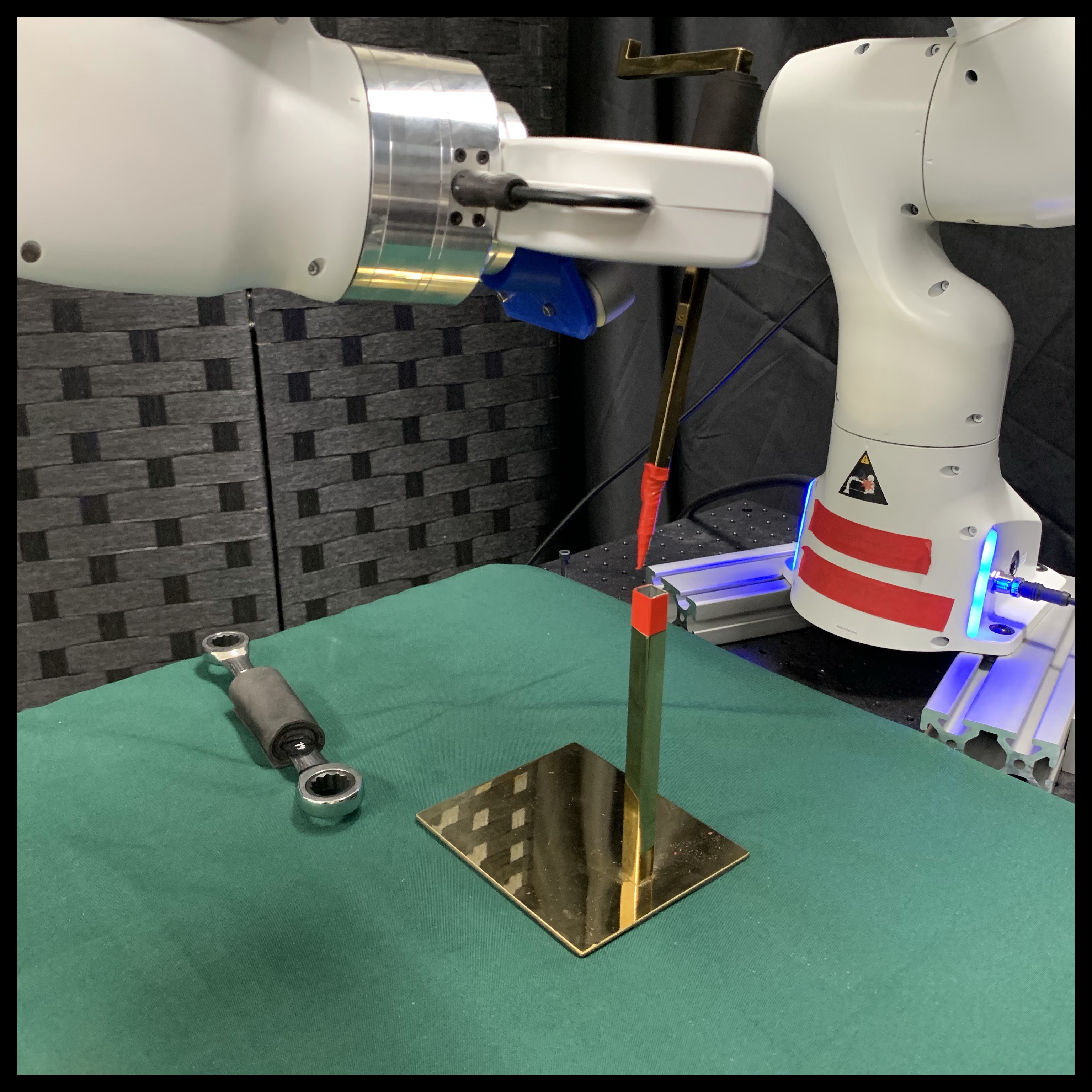}
\caption{Tool Hang (Real)} 
\end{subfigure}
\hfill
\begin{subfigure}{0.20\linewidth}
\centering
\includegraphics[width=1.0\textwidth]{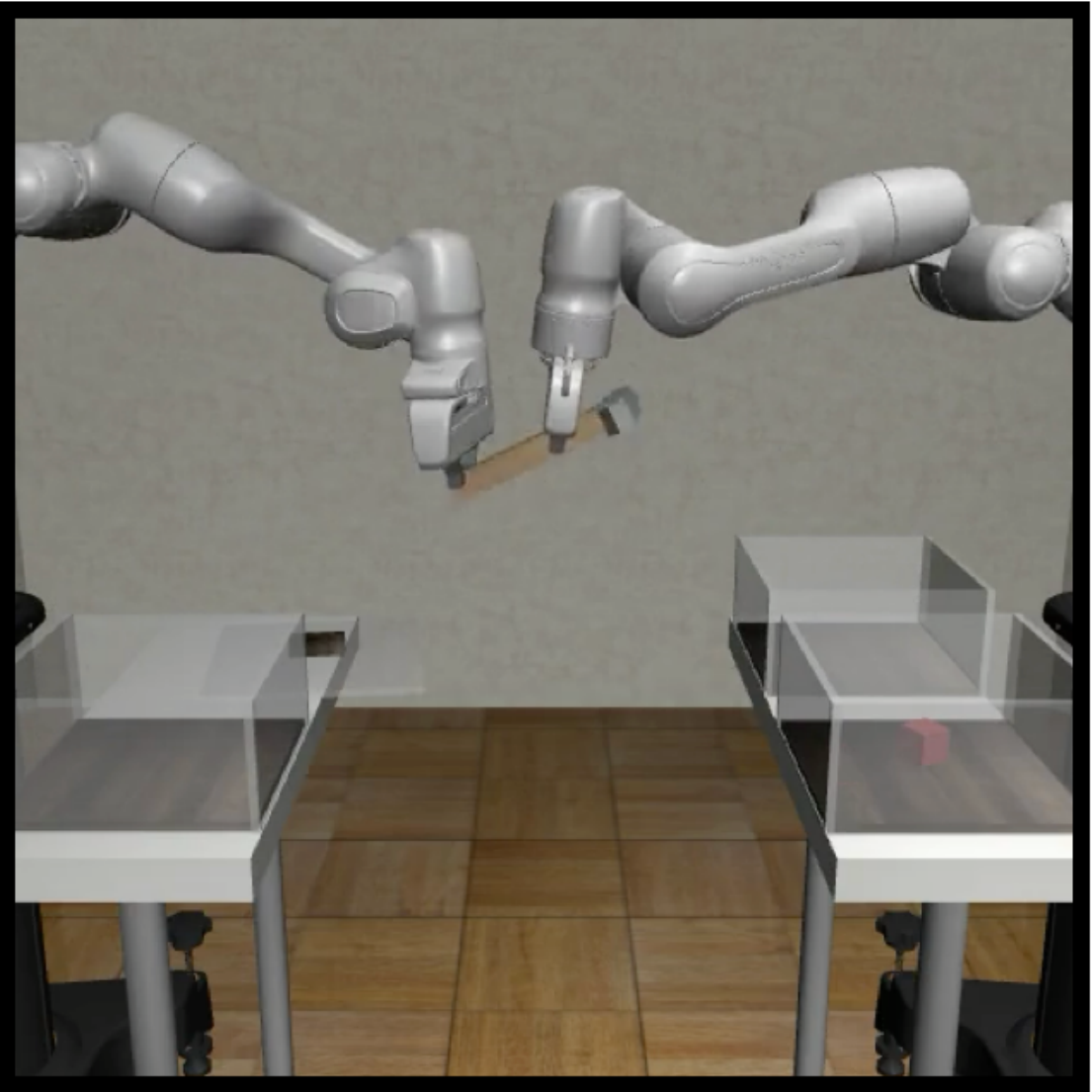}
\caption{Transport} 
\end{subfigure}
\caption{\textbf{Tasks.} We collect datasets across 6 operators of varying proficiency and evaluate offline policy learning methods on 8 challenging manipulation tasks that test a wide range of manipulation capabilities including pick-and-place, multi-arm coordination, and high-precision insertion and assembly.}
\label{fig:benchmark_tasks}
\vspace{-15pt}
\end{figure}

From our results, we point out several lessons to guide future research in leveraging human supervision for robot manipulation effectively. 
We find that history-dependent models can be extremely effective in learning from single and multi-human datasets while state-of-the-art batch RL algorithms struggle to learn from such datasets, and that the choice of observation space and hyperparameters play a substantial role in training proficient policies.
We also find that there is substantial promise for solving more complex tasks using large-scale human datasets and that our insights directly transfer to real-world scenarios, making this an important setting to explore further.



\section{Challenges in Offline Learning from Human Datasets}
\label{sec:challenges}

In this section, we outline five challenges in offline learning from human datasets that motivate different factors that we investigate in our study.

\textbf{(C1) Data from Non-Markovian Decision Process.} Human demonstrations can differ substantially from machine-generated demonstrations because humans may not act purely based on a single current observation. External factors (teleoperation device, past actions, history of episode) may all play a role. 
Prior work~\cite{mandlekar2020iris} has noted substantial benefits from leveraging models that are history-dependent and / or with temporal abstraction to learn from human demonstrations. We investigate various design choices related to such architectures in this study.

\textbf{(C2) Variance in Demonstration Quality from Multiple Humans.} Prior work~\cite{mandlekar2020iris, mandlekar2019scaling} has found that data collected from several humans can differ substantially in both demonstration proficiency and solution strategy. Differences in supervisor proficiency can manifest in many ways, such as large variations in trajectory length and noise in robot movement or mistakes (e.g. missed grasps). 
In our study, we evaluate offline policy learning algorithms on such datasets. While recent batch RL algorithms have shown an excellent ability to learn from mixed quality machine-generated datasets~\cite{fujimoto2019off,kumar2020conservative}, we empirically find that they fail to learn well from mixed quality human data.

\textbf{(C3) Dependence on Dataset Size.} Offline policy learning is sensitive to the state and action space coverage in the dataset, and by extension, the size of the dataset itself. In our study, we investigate how dataset sizes affect policy performance. This analysis is useful to understand the value of adding more data -- an important consideration since collecting human demonstrations can be costly.

\textbf{(C4) Mismatch between Training and Evaluation Objectives.} Unlike traditional supervised learning, where model selection can be achieved by using the model with the lowest validation loss~\cite{ross2011reduction}, offline policy learning often suffers from the fact that the training objective is only a surrogate for the true objective of interest (e.g. task success rate), and policy performance can change significantly from epoch to epoch. This makes it difficult to select the best trained model~\cite{gulcehre2020rl, paine2020hyperparameter, fu2021benchmarks}. In our study, we evaluate each policy checkpoint online in the environment in simulation, and report the best policy success rate per training run. We use these ground-truth values to understand the effectiveness of different selection criteria, and confirm that offline policy selection is an important problem, especially in real-world scenarios where large-scale empirical evaluation is difficult.

\textbf{(C5) High Sensitivity to Agent Design Decisions.} Prior studies on machine-generated datasets have shown that offline policy learning can be extremely sensitive to hyperparameter choices~\cite{gulcehre2020rl, paine2020hyperparameter}. In our study, we explore how agent design decisions affect policy performances, including the choice of agent architecture, agent observation space, and hyperparameter choices per algorithm. This results in several practical conclusions that should prove useful to researchers and practitioners alike. We further show that important design decisions made through our study in simulation directly translate to effective policy learning on real world tasks and datasets.

\section{Study Design}
\label{sec:setup}

\begin{table}[t!]
\centering
\resizebox{0.8\linewidth}{!}{

\begin{tabular}{ccccccc}
\toprule
\textbf{Dataset} & 
\begin{tabular}[c]{@{}c@{}}\textbf{BC}\end{tabular} &
\begin{tabular}[c]{@{}c@{}}\textbf{BC-RNN}\end{tabular} & 
\begin{tabular}[c]{@{}c@{}}\textbf{BCQ}\end{tabular} & 
\begin{tabular}[c]{@{}c@{}}\textbf{CQL}\end{tabular} & 
\begin{tabular}[c]{@{}c@{}}\textbf{HBC}\end{tabular} & 
\begin{tabular}[c]{@{}c@{}}\textbf{IRIS}\end{tabular} \\ 
\midrule

Lift (MG) & $65.3\pm2.5$ & $70.7\pm3.4$ & $\mathbf{91.3\pm1.9}$ & $64.0\pm2.8$ & $47.3\pm4.1$ & $\mathbf{96.0\pm1.6}$ \\
Can (MG) & $64.7\pm3.4$ & $68.7\pm2.5$ & $\mathbf{75.3\pm0.9}$ & $1.3\pm0.9$ & $40.7\pm3.4$ & $48.0\pm6.5$ \\
\midrule
Lift (PH) & $\mathbf{100.0\pm0.0}$ & $\mathbf{100.0\pm0.0}$ & $\mathbf{100.0\pm0.0}$ & $92.7\pm5.0$ & $\mathbf{100.0\pm0.0}$ & $\mathbf{100.0\pm0.0}$ \\
Can (PH) & $95.3\pm0.9$ & $\mathbf{100.0\pm0.0}$ & $88.7\pm0.9$ & $38.0\pm7.5$ & $\mathbf{100.0\pm0.0}$ & $\mathbf{100.0\pm0.0}$ \\
Square (PH) & $78.7\pm1.9$ & $\mathbf{84.0\pm0.0}$ & $50.0\pm4.9$ & $5.3\pm2.5$ & $\mathbf{82.6\pm0.9}$ & $78.7\pm2.5$ \\
Transport (PH) & $17.3\pm2.5$ & $\mathbf{71.3\pm6.6}$ & $7.3\pm3.3$ & $0.0\pm0.0$ & $48.6\pm3.8$ & $41.3\pm3.4$ \\
Tool Hang (PH) & $\mathbf{29.3\pm0.9}$ & $19.3\pm5.0$ & $0.0\pm0.0$ & $0.0\pm0.0$ & $\mathbf{30.0\pm7.1}$ & $11.3\pm2.5$ \\
\midrule
Lift (MH) & $\mathbf{100.0\pm0.0}$ & $\mathbf{100.0\pm0.0}$ & $\mathbf{100.0\pm0.0}$ & $56.7\pm40.3$ & $\mathbf{100.0\pm0.0}$ & $\mathbf{100.0\pm0.0}$ \\
Can (MH) & $86.0\pm4.3$ & $\mathbf{100.0\pm0.0}$ & $62.7\pm8.2$ & $22.0\pm5.7$ & $91.3\pm2.5$ & $92.7\pm0.9$ \\
Square (MH) & $52.7\pm6.6$ & $\mathbf{78.0\pm4.3}$ & $14.0\pm4.3$ & $0.7\pm0.9$ & $60.7\pm5.0$ & $52.7\pm5.0$ \\
Transport (MH) & $11.3\pm2.5$ & $\mathbf{65.3\pm7.4}$ & $2.6\pm0.9$ & $0.0\pm0.0$ & $14.0\pm1.6$ & $10.7\pm0.9$ \\

\bottomrule
\end{tabular}




}
\vspace{+3pt}
\caption{\textbf{Results on Low-Dimensional Observations.} We present success rates averaged over 3 seeds for each method across the low-dim Machine-Generated (MG), Proficient-Human (PH), and Multi-Human (MH) datasets. The results show that methods that model temporal correlations (BC-RNN, HBC, IRIS) exhibit strong performance on human datasets. Furthermore, while Batch RL algorithms like BCQ are proficient on machine-generated data, they perform poorly on human datasets.}
\label{table:core_low_dim_all}
\vspace{-25pt}
\end{table}
\begin{table}[t!]
\centering
\resizebox{0.8\linewidth}{!}{
\begin{tabular}{ccccccc}
\toprule
\textbf{Dataset} & 
\begin{tabular}[c]{@{}c@{}}\textbf{BC}\end{tabular} &
\begin{tabular}[c]{@{}c@{}}\textbf{BC-RNN}\end{tabular} & 
\begin{tabular}[c]{@{}c@{}}\textbf{BCQ}\end{tabular} & 
\begin{tabular}[c]{@{}c@{}}\textbf{CQL}\end{tabular} & 
\begin{tabular}[c]{@{}c@{}}\textbf{HBC}\end{tabular} & 
\begin{tabular}[c]{@{}c@{}}\textbf{IRIS}\end{tabular} \\ 
\midrule

Can-Worse & $56.7\pm2.5$ & $\mathbf{92.0\pm1.6}$ & $29.3\pm10.9$ & $4.0\pm3.3$ & $78.7\pm3.4$ & $77.3\pm1.9$ \\
Can-Okay & $72.0\pm2.8$ & $\mathbf{95.3\pm1.9}$ & $58.0\pm8.6$ & $22.0\pm4.3$ & $\mathbf{97.3\pm0.9}$ & $\mathbf{96.0\pm0.0}$ \\
Can-Better & $83.3\pm2.5$ & $\mathbf{99.3\pm0.9}$ & $62.0\pm5.9$ & $20.7\pm7.4$ & $\mathbf{96.7\pm0.9}$ & $\mathbf{96.0\pm0.0}$ \\
\midrule
Can-Worse-Okay & $74.7\pm5.7$ & $\mathbf{98.7\pm1.9}$ & $50.7\pm3.8$ & $18.7\pm2.5$ & $88.0\pm1.6$ & $87.3\pm1.9$ \\
Can-Worse-Better & $76.0\pm4.3$ & $\mathbf{100.0\pm0.0}$ & $48.0\pm4.9$ & $20.7\pm5.7$ & $90.0\pm1.6$ & $91.3\pm2.5$ \\
Can-Okay-Better & $90.7\pm1.9$ & $\mathbf{100.0\pm0.0}$ & $68.7\pm2.5$ & $30.7\pm7.7$ & $\mathbf{99.3\pm0.9}$ & $\mathbf{98.0\pm1.6}$ \\
\midrule
Square-Worse & $22.0\pm4.3$ & $39.3\pm3.8$ & $5.3\pm1.9$ & $0.0\pm0.0$ & $\mathbf{44.7\pm6.8}$ & $38.7\pm0.9$ \\
Square-Okay & $27.3\pm3.4$ & $45.3\pm2.5$ & $6.7\pm1.9$ & $0.0\pm0.0$ & $\mathbf{52.0\pm2.8}$ & $42.0\pm3.3$ \\
Square-Better & $58.7\pm2.5$ & $\mathbf{66.0\pm2.8}$ & $32.0\pm4.3$ & $0.7\pm0.9$ & $61.3\pm1.9$ & $60.0\pm1.6$ \\
\midrule
Square-Worse-Okay & $28.7\pm2.5$ & $\mathbf{55.3\pm0.9}$ & $8.7\pm1.9$ & $2.7\pm1.9$ & $50.7\pm4.1$ & $43.3\pm2.5$ \\
Square-Worse-Better & $46.7\pm5.7$ & $\mathbf{73.3\pm6.2}$ & $15.3\pm2.5$ & $1.3\pm0.9$ & $65.3\pm3.4$ & $56.7\pm3.4$ \\
Square-Okay-Better & $56.7\pm4.1$ & $\mathbf{74.0\pm2.8}$ & $22.0\pm4.3$ & $1.3\pm0.9$ & $63.3\pm4.1$ & $56.7\pm3.8$ \\
\midrule
Can-Paired & $64.0\pm9.1$ & $70.0\pm4.3$ & $44.7\pm1.9$ & $6.0\pm1.6$ & $70.7\pm5.2$ & $\mathbf{75.3\pm1.9}$ \\


\bottomrule
\end{tabular}
}
\vspace{+3pt}
\caption{\textbf{Results on Suboptimal Human Data.} We present success rates averaged over 3 seeds for each method across different subsets of the Multi-Human datasets, corresponding to mixtures of demonstrations from ``Better'', ``Adequate'', and ``Worse'' human operators, and finally on a diagnostic dataset with paired success and failure human trajectories for each starting initialization. Results indicate that BC-RNN is a strong baseline, and that Batch RL methods perform poorly across all datasets, even on the simple diagnostic dataset.}
\label{table:subopt}
\vspace{-25pt}
\end{table}



\subsection{Tasks}

We conducted our study across 5 simulated and 3 real world tasks. The tasks were chosen to test a broad range of manipulation capabilities. See Fig~\ref{fig:benchmark_tasks} and Appendix~\ref{app:task} for more details.

\textbf{Lift (sim + real)).} The robot arm must lift a small cube. 
This is the simplest task.

\textbf{Can (sim + real).} The robot must place a coke can from a large bin into a smaller target bin. 
Slightly more challenging than Lift, since picking the can is harder than picking the cube, and the can must also be placed into the bin.

\textbf{Square (sim).} The robot must pick a square nut and place it on a rod. 
Substantially more difficult than Lift and Pick Place Can due to the precision needed to pick up the nut and insert it on the rod.

\textbf{Transport (sim).} Two robot arms must transfer a hammer from a closed container on a shelf to a target bin on another shelf. One robot arm must retrieve the hammer from the container, while the other arm must clear the target bin by moving a piece of trash to the nearby receptacle. Finally, one arm must hand the hammer over to the other, which must place the hammer in the target bin. 

\textbf{Tool Hang (sim + real).} A robot arm must assemble a frame consisting of a base piece and hook piece by inserting the hook into the base, and hang a wrench on the hook. 
This is the most difficult task due to the multiple stages that each require precise, and dexterous, rotation-heavy movements. 


\subsection{Data Collection}

To study the effect of dataset source, we collected data from three different sources -- Machine-Generated, Proficient-Human, and Multi-Human (more details in Appendix~\ref{app:dataset}). 


\textbf{Machine-Generated (MG).} We collected these datasets by first training a state-of-the-art RL algorithm~\cite{haarnoja2018soft} on the Lift and Can task, taking agent checkpoints that are saved regularly during training, and collecting 300 rollout trajectories from each checkpoint. Consequently, these datasets are comprised of mixtures of expert and suboptimal data, and resemble datasets from common offline RL benchmarks~\cite{fu2020d4rl, gulcehre2020rl}. We excluded other tasks because they could not be solved by the RL algorithm even with substantial tuning. See the appendix for more details.

\textbf{Proficient-Human (PH) and Multi-Human (MH).} Datasets are collected by humans through RoboTurk~\cite{mandlekar2018roboturk, mandlekar2019scaling}, a remote teleoperation platform. 
The PH datasets consist of 200 demonstrations collected by a single, experienced teleoperator, while the MH datasets consist of 300 demonstrations, collected by 6 teleoperators of varying proficiency, each of which provided 50 demonstrations. The 6 teleoperators consisted of a ``better'' group of 2 experienced operators, an ``okay'' group of 2 adequate operators, and a ``worse'' group of 2 inexperienced operators. These data subsets in the Multi-Human data allowed us to investigate the ability of algorithms to deal with mixed quality human data. 

\textbf{Observation Modalities.} 
To study the effect of observation modalities, we capture a diverse set of sensor streams when collecting the dataset, including end-effector, gripper fingers, and joints, ground-truth object poses, and images from an external camera and wrist-mounted camera per robot arm \revision{(see Appendix~\ref{app:task})}. 
We have two observation spaces -- ``low-dim'' and ``image''. Both include end-effector poses and gripper finger positions, and only differ in whether ground-truth object information is used (low-dim) or whether that information is replaced by the available camera observations (image).


\subsection{Training and Evaluation Protocols}

There are several approaches to offline imitation learning~\cite{Ijspeert2002MovementIW, Finn2017OneShotVI, Billard2008RobotPB, Calinon2010LearningAR, zhang2017deep, mandlekar2020learning, zeng2020transporter, wang2021generalization, tung2020learning} and offline reinforcement learning~\cite{fujimoto2019off, kumar2020conservative, wu2019behavior, wang2020critic, siegel2020keep, kidambi2020morel, yu2020mopo, ghasemipour2020emaq, yu2021combo} (see Appendix~\ref{app:related} for more discussion on related work). 
We chose to evaluate 6 algorithms in this study -- Behavioral Cloning (BC), BC with an RNN policy (BC-RNN), Hierarchical Behavioral Cloning (HBC)~\cite{mandlekar2020learning}, Batch-Constrained Q-Learning (BCQ)~\cite{fujimoto2019off}, Conservative Q-Learning (CQL)~\cite{kumar2020conservative}, and IRIS~\cite{mandlekar2020iris}. BC-RNN, HBC, and IRIS have all been used in prior work to learn offline from teleoperated human demonstrations, while BCQ and CQL are commonly-used offline RL algorithms (see Appendix~\ref{app:algorithm}). \revision{We use binary task completion rewards for all our experiments.} Each agent is trained for $N$ epochs, where each epoch consists of $M$ gradient steps, and evaluated every $E$ epochs, by running 50 rollouts in the environment and reporting the success rate over a maximum horizon. For each agent, we report the maximum success rate over the coarse of training, and average over 3 seeds. For low-dim agents, $N=2000$, $M=100$, and $E=50$, and for image agents, $N=600$, $M=500$, and $E=20$ (see Appendix~\ref{app:dataset-training-setup}).

\section{Experiments}
\label{sec:experiments}

\begin{figure}[t!]
\begin{subfigure}{0.3\linewidth}
\centering
\includegraphics[width=1.0\textwidth]{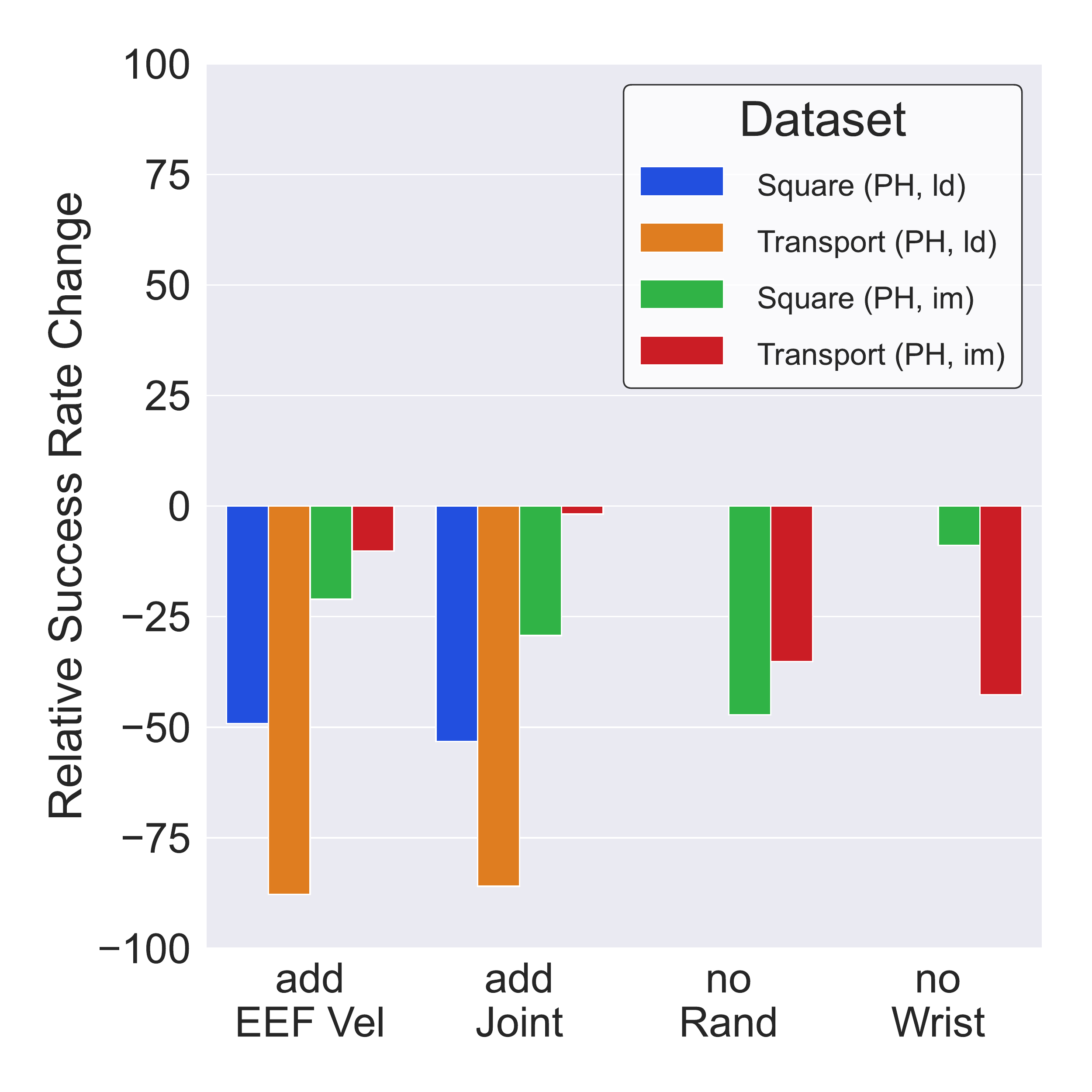}
\caption{Observation Space} 
\label{fig:obs}
\end{subfigure}
\hfill
\begin{subfigure}{0.3\linewidth}
\centering
\includegraphics[width=1.0\textwidth]{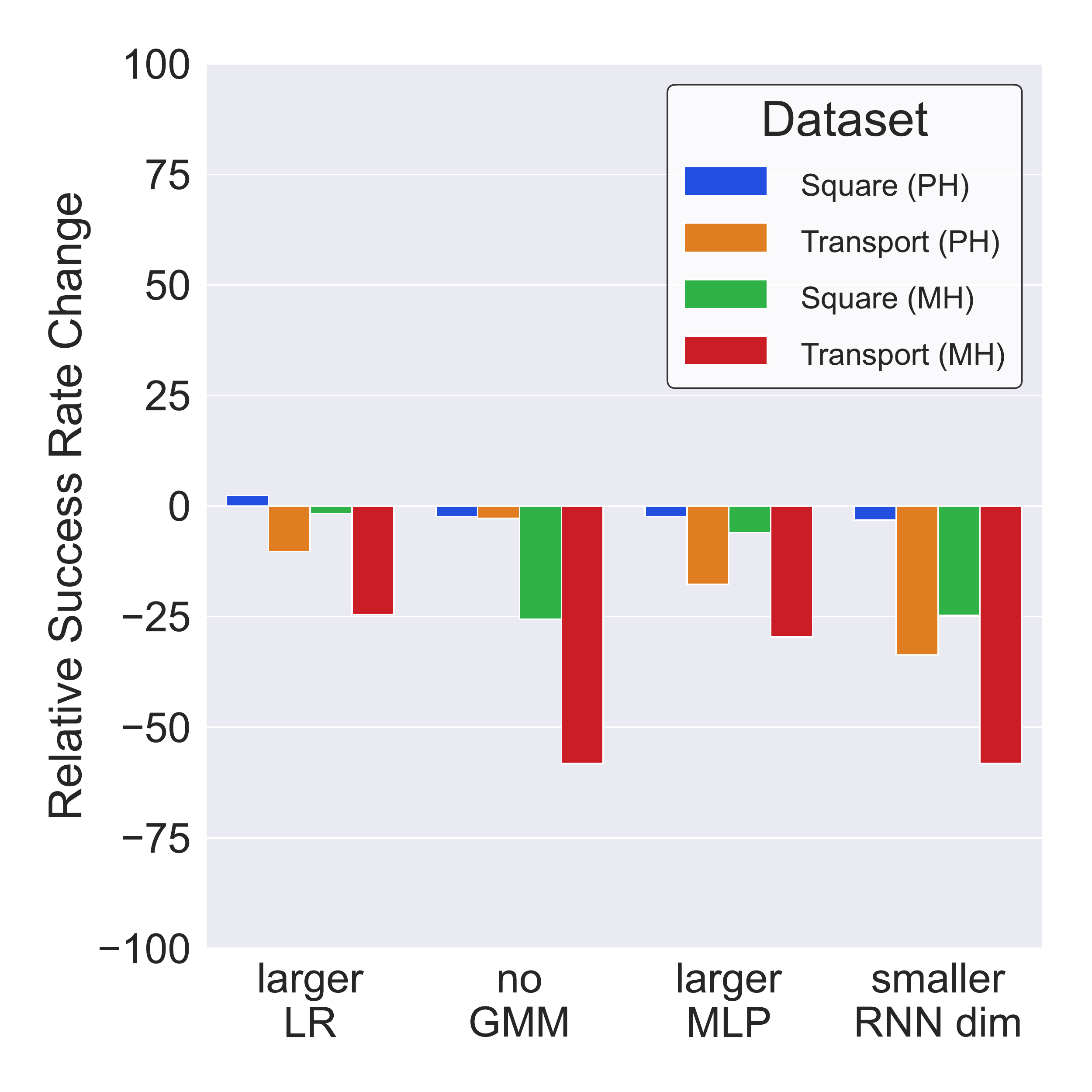}
\caption{Low-Dim Hyperparameters} 
\label{fig:hyper-low-dim}
\end{subfigure}
\hfill
\begin{subfigure}{0.3\linewidth}
\centering
\includegraphics[width=1.0\textwidth]{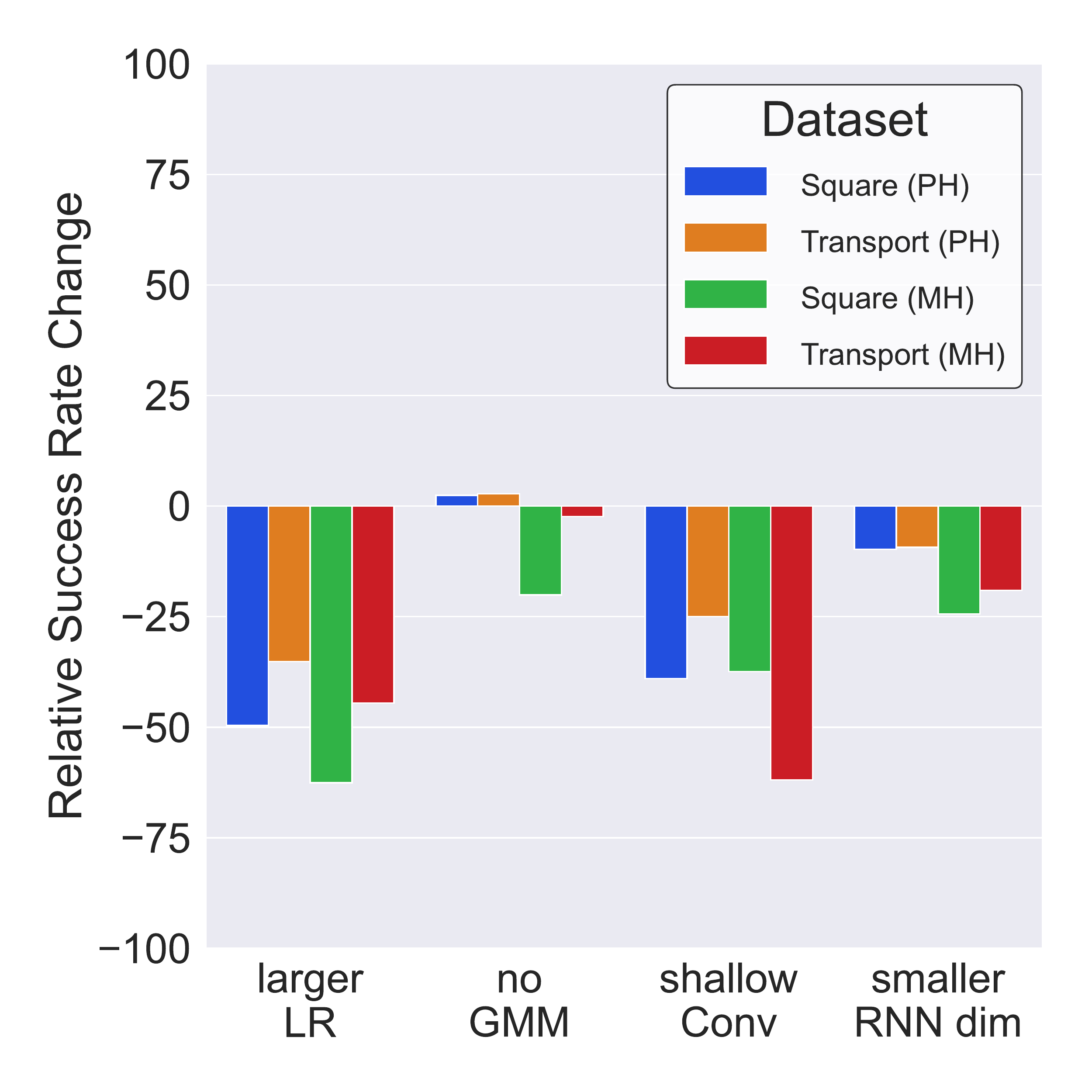}
\caption{Image Hyperparameters} 
\label{fig:hyper-image}
\end{subfigure}
\caption{\textbf{Effect of Observation Space and Hyperparameter Choice.} We show how the success rate that BC-RNN obtains can drop drastically due to changes to the observation space and hyperparameter settings.}
\label{fig:obs-hyper}
\vspace{-10pt}
\end{figure}
\begin{figure}[t!]
\begin{subfigure}{0.46\linewidth}
\centering
\includegraphics[width=1.0\textwidth]{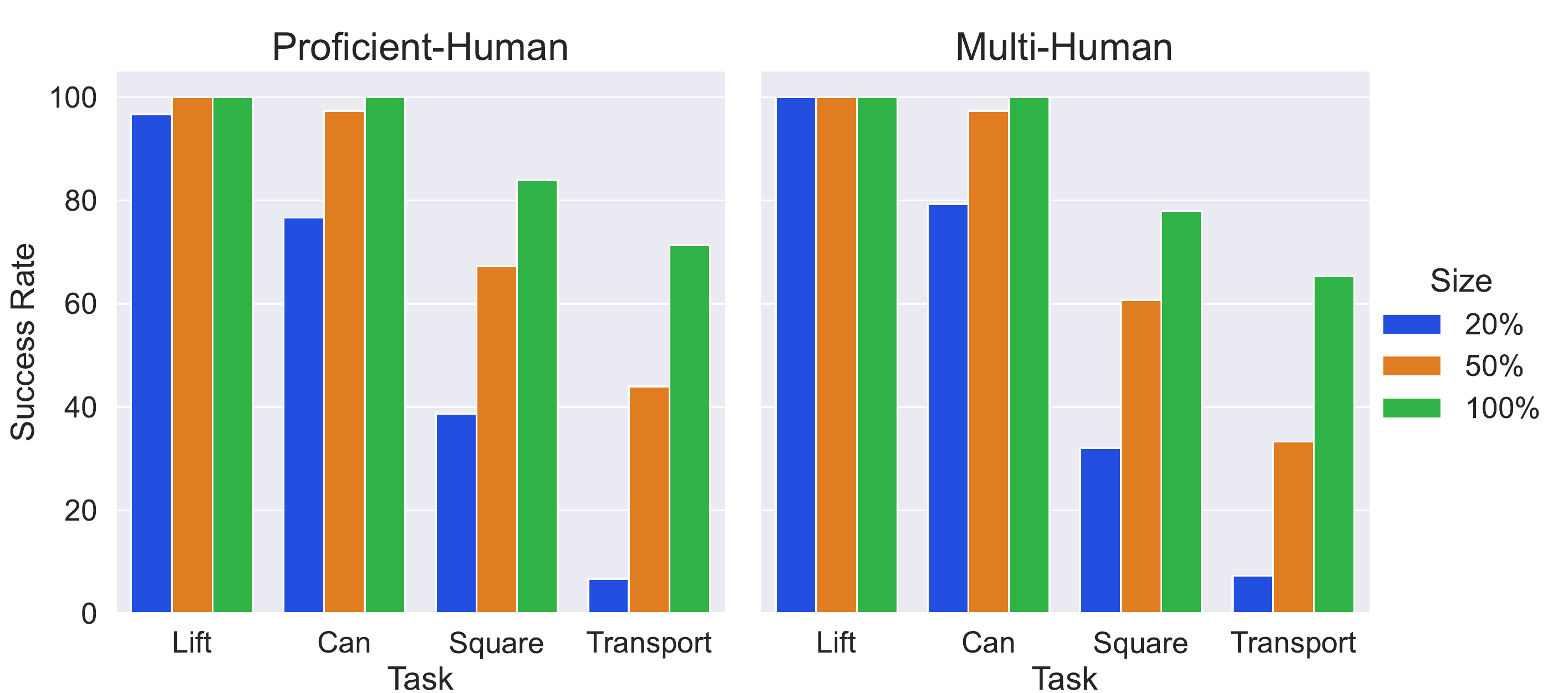}
\caption{Low-Dim} 
\end{subfigure}
\hfill
\begin{subfigure}{0.46\linewidth}
\centering
\includegraphics[width=1.0\textwidth]{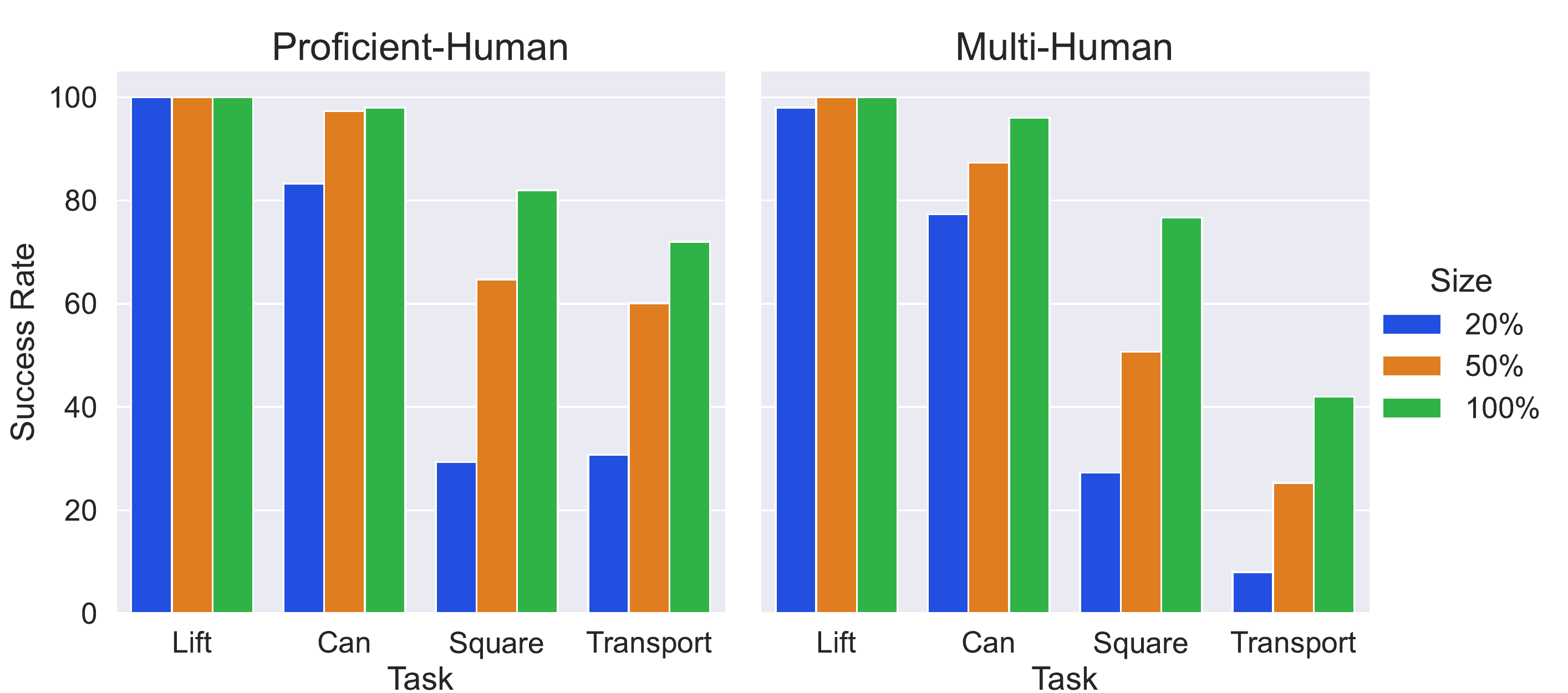}
\caption{Image} 
\end{subfigure}
\caption{\textbf{Effect of Dataset Size.} We study how the BC-RNN success rate changes when lowering the quantity of data to 20\% and 50\%. Results show that less complex tasks (Lift, Can) be learned with a fraction of the data, while more complex tasks might benefit from even larger human datasets.}
\label{fig:dataset_size}
\vspace{-15pt}
\end{figure}

In this section, we present each factor that we explored in our study, and note the relevant challenges from Sec.~\ref{sec:challenges} that each pertains to. 

\subsection{Algorithm Comparison on Single and Multi-Human Demonstrations (C1, C2)} 
\label{exp:human}

We trained and evaluated all algorithms on the Proficient-Human (PH) and Multi-Human (MH) datasets and report the average success rates across 3 seeds in Table~\ref{table:core_low_dim_all}.

\textbf{Observation history is crucial for good performance.} 
There is a substantial performance gap between BC-RNN and BC, which highlights the benefits of history-dependence. The performance gap is larger for longer-horizon tasks (e.g. $\sim55\%$ for Transport (PH) compared to $\sim5\%$ for Square (PH)) and for multi-human data compared to single-human data (e.g. $\sim25\%$ for Square (MH) compared to $\sim5\%$ for Square (PH)). Interestingly, results are lower for MH datasets compared to PH datasets, even though the MH datasets contain 100 more demos (300 demos vs. 200 demos). This most likely stems from the presence of suboptimal and multimodal data in the MH datasets.

\textbf{Batch RL algorithms perform poorly on Human Datasets.} Recent batch (offline) RL algorithms such as BCQ and CQL have demonstrated excellent results in learning from suboptimal and multimodal agent-generated datasets. Our results confirm the capacity of such algorithms to work well -- BCQ in particular performs strongly on our agent-generated MG datasets that consist of a diverse mixture of good and poor policies. Surprisingly though, neither BCQ nor CQL performs particularly well on these human-generated datasets. This puts the ability of such algorithms to learn from more natural dataset distributions into question (instead of those collected via RL exploration or pre-trained agents). There is an opportunity for future work in batch RL to resolve this gap.



\subsection{Learning from Suboptimal Human Data (C2)}
\label{exp:suboptimal}

To further investigate how algorithms deal with suboptimal human data, we split our MH datasets into smaller subsets based on the proficiency of the human operators. The MH-Better, MH-Okay, and MH-Worse are the 100 demo subsets corresponding to the 2 ``better'', 2 ``okay'', and 2 ``worse'' operators respectively, while MH Worse-Okay, MH Worse-Better, and MH Okay-Better are the 200 demo subsets corresponding to the mixture of the previous subsets. Similar data mixtures have been used for evaluations in batch RL~\cite{fu2020d4rl}.
Appendix~\ref{app:dataset} shows the average trajectory lengths in each data subset -- lower quality datasets contain demonstrations that take more time to solve the task.

\textbf{BC-RNN is a strong baseline on suboptimal human data, but there is room for improvement.} 
Table~\ref{table:subopt} shows that BC exhibits a large performance gap between the Better and Worse 100-demo subsets (roughly 27\% and 35\% for Can and Square respectively). Interestingly, BC-RNN is able to nearly eliminate this gap in performance on the Can task, but not on the Square task. However, BC-RNN outperforms BC on all datasets (7\%-35\% improvement). Comparing results on the 100 Better demonstrations and 100 Okay demonstrations to the 200 Worse-Better demonstrations and 200 Worse-Okay demonstrations further allows us to analyze how adding 100 ``worse'' demonstrations impacts the performance of each algorithm. Most algorithms decline in performance while BC-RNN is able to uniformly improve from the added data. Comparing the performance of BC-RNN on the 200-demo Square mixture datasets (55.3\%, 73.3\%, 74.0\%) to the high-quality 200-demo Square (PH) dataset (84.0\%) shows that there is still room for algorithms to improve on the use of this data.



\textbf{Diagnostic dataset shows that Batch RL struggles in simpler settings as well.} The final row of Table~\ref{table:subopt} shows additional results on a diagnostic dataset termed Can-Paired, where a single operator collected 2 demonstrations for each of 100 task initializations -- one successful demonstration, and one where the can is tossed outside of the bin (task failure), for a total of 200 demonstrations. There is a strong expectation for batch RL algorithms to be able to distinguish between actions leading to successful placement and actions leading to task failure, but even in this simple setting, most algorithms suffer, providing a pessimistic view of the state-of-the-art. The $5\%$ improvement that IRIS provides over BC-RNN suggests that introducing history-dependence into state-of-the-art batch RL algorithms might be a promising direction for future work.


\subsection{Effect of Observation Space (C5)}
\label{exp:obs}

\begin{figure}[t!]
\centering
\begin{subfigure}{0.38\textwidth}
\vskip 0pt
\centering
\includegraphics[width=\columnwidth]{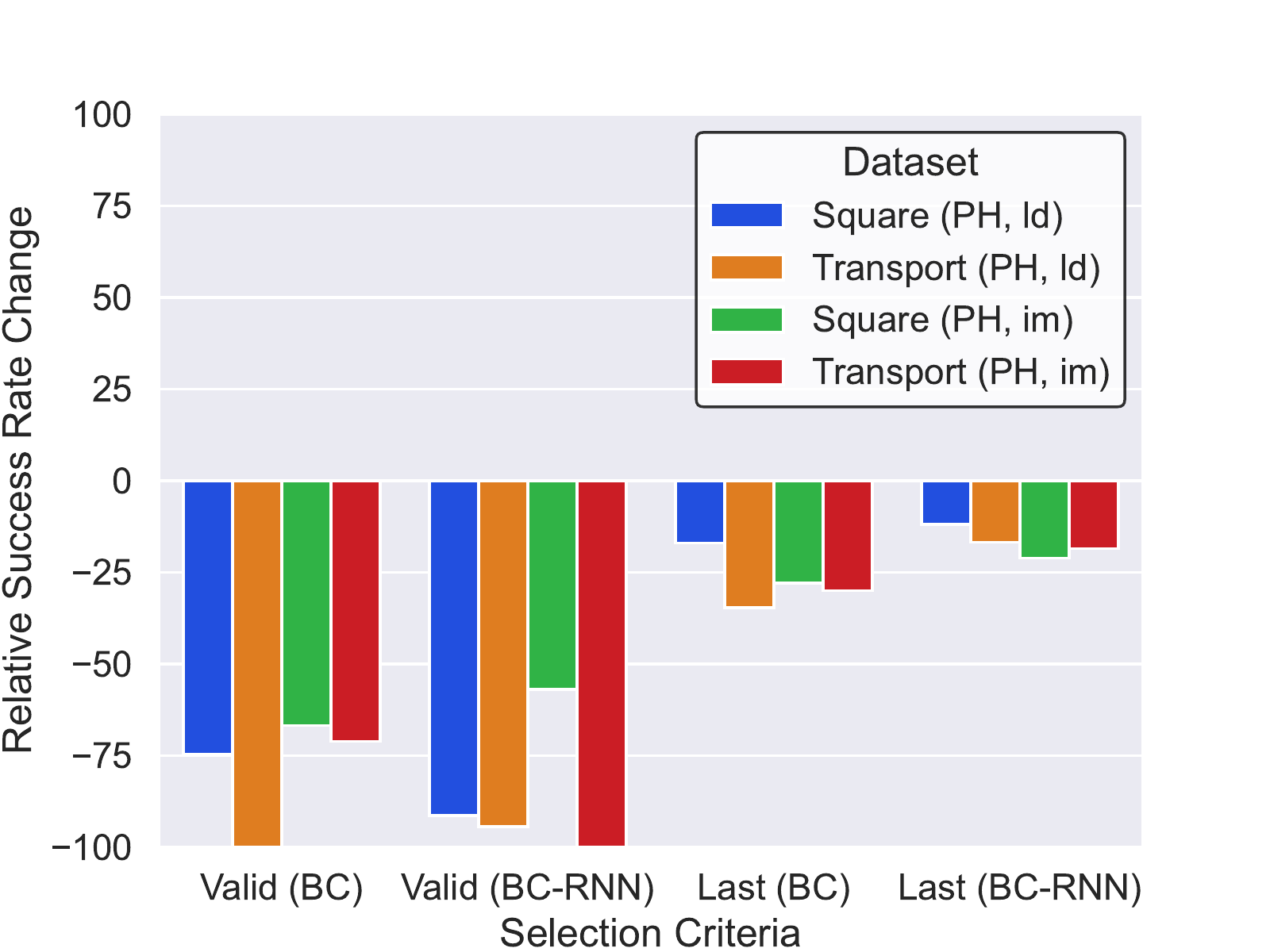}
\caption{\textbf{Effect of Policy Selection Criteria}}
\label{fig:valid}
\end{subfigure}
\hfill
\begin{minipage}[t]{0.6\textwidth}
\centering
\resizebox{\linewidth}{!}{
\begin{tabular}{ccccc}
\toprule
\textbf{Dataset} & 
\begin{tabular}[c]{@{}c@{}}\textbf{BC}\end{tabular} &
\begin{tabular}[c]{@{}c@{}}\textbf{BC-RNN}\end{tabular} & 
\begin{tabular}[c]{@{}c@{}}\textbf{BCQ}\end{tabular} & 
\begin{tabular}[c]{@{}c@{}}\textbf{CQL}\end{tabular} \\
\midrule
Lift (PH) & $\mathbf{100.0\pm0.0}$ & $\mathbf{100.0\pm0.0}$ & $\mathbf{98.0\pm1.6}$ & $52.0\pm13.0$ \\
Can (PH) & $\mathbf{97.3\pm1.9}$ & $\mathbf{98.0\pm0.9}$ & $86.7\pm2.5$ & $0.7\pm0.9$ \\
Square (PH) & $62.0\pm4.9$ & $\mathbf{82.0\pm0.0}$ & $41.3\pm4.1$ & - \\
Transport (PH) & $55.3\pm6.2$ & $\mathbf{72.0\pm4.3}$ & $0.7\pm0.9$ & - \\
Tool Hang (PH) & $20.0\pm5.9$ & $\mathbf{67.3\pm4.1}$ & $3.3\pm0.9$ & - \\
\midrule
Lift (MH) & $\mathbf{100.0\pm0.0}$ & $\mathbf{100.0\pm0.0}$ & $93.3\pm0.9$ & $11.3\pm9.3$ \\
Can (MH) & $85.3\pm0.9$ & $\mathbf{96.0\pm1.6}$ & $77.3\pm6.8$ & $0.0\pm0.0$ \\
Square (MH) & $46.0\pm1.6$ & $\mathbf{76.7\pm3.4}$ & $17.3\pm7.5$ & - \\
Transport (MH) & $18.7\pm2.5$ & $\mathbf{42.0\pm1.6}$ & $0.0\pm0.0$ & - \\
\bottomrule
\end{tabular}
}
\captionof{table}{\textbf{Results on Image Observations.}}
\label{table:core_image_all}
\end{minipage}
\caption{(left) \textbf{Effect of Policy Selection Criteria.} We compare how performance decreases when choosing the policy to evaluate by using the lowest validation loss, or when using the final trained checkpoint, with respect to the best policy performance. (right) \textbf{Results on Image Observations.} We present success rates for each method across the image observation human datasets. BC-RNN maintains nearly the same performance as learning from ground-truth observations, providing an optimistic view for learning with real-world raw sensory observations.}
\label{fig:valid-core-image}
\vspace{-20pt}
\end{figure}



\textbf{Learning from image observations can match low-dim agent performance.}
In Table~\ref{table:core_image_all}, we present policy learning results when using image observations instead of ground-truth object locations -- an important setting for real-world policy learning. BC-RNN still maintains superior performance improvements over BC on the complex Square and Transport tasks, and with the exception of Transport (MH), maintains nearly the same performance as learning from ground-truth observations. This result provides an optimistic view for learning with real-world raw sensory observations. 

\textbf{Features used for robot proprioception can matter.} 
In Fig~\ref{fig:obs}, we study the effect of adding end effector velocities to the observations (+ EEF Vel), and joint positions and velocities to the observations (+ Joint). 
Surprisingly, we find that including end effector velocity information, and joint information hurts agents trained on low-dim observations substantially (49\%-88\% relative performance drop), while image-based agents are more tolerant to the inclusion of this extra information (2\%-29\% relative performance drop). We hypothesize that performance drops might be due to overfitting to the presence of this extra information not needed for solving these tasks. Thus, practitioners should take care to engineer the robot observation space and exclude possibly irrelevant information -- information-hiding can be a powerful paradigm for training proficient robots~\cite{lee2020learning}.

\textbf{Image randomization and wrist observations can be crucial for manipulation tasks.} 
In Fig~\ref{fig:obs}, we report performance drops from removing pixel shift image randomization (- Rand) and the wrist camera (- Wrist) from image-based agents to understand their importance. We see that removing randomization results in 47\% and 35\% relative performance drops on Square and Transport respectively, and removing wrist images results in 9\% and 43\% relative drops. Consequently, both wrist camera images and image randomization play a substantial role in producing performant policies. We confirm the importance of each for visuomotor imitation in the real world as well (see Sec~\ref{exp:real}). Wrist observations likely help the robot improve gripper alignment for grasping and randomization helps the policy develop invariance for portions of the image that are not important for action prediction.


\subsection{Effect of Hyperparameter Choice (C5)}
\label{exp:hyper}


In this section, we take our default hyperparameters for BC-RNN and study the effect of changing a subset of them to report practical recommendations for learning from human datasets (see Appendix~\ref{app:offline-rl} for BCQ and CQL). We present our results in Fig~\ref{fig:hyper-low-dim} (low-dim) and Fig~\ref{fig:hyper-image} (image).

\textbf{(larger LR)} Increasing the learning rate from 1e-4 to 1e-3 affects the performance of image-agents substantially (drop of 35\%-63\%), while low-dim agents are more tolerant to the change.
\textbf{(no GMM)} Using a deterministic policy instead of learning a GMM action distribution results in significant relative performance drops on the MH datasets (especially low-dim Transport, with a drop of 58\%).
\textbf{(larger MLP)} Using a larger MLP size at each RNN timestep reduces performance uniformly, suggesting that it is possible to overfit to dataset actions if network architectures are too large.
\textbf{(shallow Conv)} Using a shallow convolutional network~\cite{finn2016deep} instead of the ResNet backbone~\cite{he2016deep} for encoding image observations reduces performance significantly -- with relative drops of 25\%-62\%, suggesting that large-capacity visual encoders are crucial for visuomotor imitation.
\textbf{(smaller RNN dim)} Reducing the size of the RNN hidden dimension from 400 to 100 (low-dim) and 1000 to 400 (image) uniformly decreases performance (drops of 3\%-58\%), showing the importance of a large RNN hidden dimension.
\textbf{(Recommendations)} We recommend tuning the LR (especially for image agents) and network structure (MLP size, size of RNN dim) carefully. Opting to use a GMM policy and a ResNet encoder appears to be uniformly better.

\subsection{Selecting a Policy to Evaluate (C4)}
\label{exp:mismatch}

Model selection in offline policy learning can be challenging -- for this reason, in our simulation experiments, we evaluated every policy checkpoint online and reported the best one. This is not feasible for real-world settings, making offline policy selection desirable. In Fig~\ref{fig:valid}, we show that this can be non-trivial, by showing the relative performance drop when selecting the policy using the best loss on validation data (common in supervised learning), and when using the final training checkpoint as well (common in offline RL~\cite{agarwal2020optimistic, fu2020d4rl, kumar2020conservative}) -- in both cases, the selected policy is significantly worse than the best one (10\% to 100\% decrease). See Appendix~\ref{app:policy-selection} for more detailed results and discussion. This motivates the need for better offline evaluation metrics. 

\subsection{Effect of Dataset Size (C3)}
\label{exp:size}

To study how dataset size impacts performance, we formed smaller 20\% and 50\% subsets of our human datasets by sampling trajectories. We evaluate low-dim and image BC-RNN agents across these subsets in Table~\ref{table:ds_size_low_dim_se} and Table~\ref{table:ds_size_low_dim_mh}. There are several promising results here. We first note that less complex tasks (Lift, Can) can yield proficient policies (75\%-100\% success rate) using a small fraction of the data (20\%). Second, while policies trained on more complex tasks (Square, Transport) suffer substantially when using 50\% or 20\% of the data, the converse is also true -- adding more data (e.g. moving from 20\% to 50\% or 50\% to 100\% size) can result in significant policy improvement. This confirms the value of using large human datasets as a means to obtain proficient policies for challenging and complex manipulation tasks.





\subsection{Applicability to Real-World Settings}
\label{exp:real}

\revision{Here, we show that design decisions made in simulation can potentially transfer to real world settings.} We collected 3 additional real-world datasets with a Franka robotic arm -- Lift (Real), Can (Real), and Tool Hang (Real). 
Each consists of 200 trajectories collected by one operator. We train BC-RNN and report the final policy checkpoint success rate, over 30 rollouts, due to the time-consuming nature of real world policy evaluation. We also emphasize that no real-world hyperparameter tuning took place, so our results are a lower bound.
We were able to train proficient \textbf{Lift (96.7\%}) and \textbf{Can (73.3\%)} policies, and the \textbf{Tool Hang (3.3\%)} policy is able to generate some task successes, despite the extremely difficult nature of the task. Furthermore, as in Sec.~\ref{exp:obs}, we validate the importance of pixel shift randomization and the wrist camera by ablating each component on the Can task, and show that including both is the difference between a proficient and non-proficient real-world policy -- \textbf{Can (- Rand) (26.7\%)}, \textbf{Can (- Wrist) (43.3\%)}.

\section{Discussion}
\label{sec:lessons}

In this section, we summarize the lessons from our study and make recommendations for future work.

\textbf{(L1) Models with temporal abstraction can be extremely effective in learning from human datasets.} In Sec~\ref{exp:human} and Sec~\ref{exp:suboptimal}, we demonstrated that history-dependent models (BC-RNN, HBC, and IRIS) are particularly effective in learning from human datasets compared to algorithms that do not take temporal context into account. 

\textbf{(L2) Need to improve the ability of batch (offline) RL to learn from suboptimal human datasets.} Sec~\ref{exp:suboptimal} and Appendix~\ref{app:offline-rl} demonstrated that state-of-the-art batch RL algorithms are excellent at learning from suboptimal machine-generated datasets but much worse at learning from suboptimal human datasets. They even struggled with a diagnostic dataset with paired good and bad human demonstration trajectories while IRIS was able to improve slightly on BC-RNN, suggesting that combining history-dependence with value learning might be a good place to start for improving batch RL methods~\cite{ajay2020opal, pertsch2020accelerating, chen2021decision, janner2021reinforcement}. This also demonstrates a need to start benchmarking new batch RL algorithms on human datasets instead of purely on machine-generated datasets. 

\textbf{(L3) Improving offline policy selection is important for real world settings.} Sec~\ref{exp:mismatch} demonstrated the need for better ways to select an evaluation policy in an offline manner. We hope that our datasets can help supplement other efforts~\cite{fu2021benchmarks}. 


\textbf{(L4) Observation space plays a large role and hyperparameters matter.} Sec~\ref{exp:obs} demonstrates that policies trained on low-dim observations can be very sensitive to the choice of robot proprioception, while pixel shift randomization and wrist camera images are critical for effective visuomotor policy learning. The choice of observation space for imitation merits careful consideration -- other work has also confirmed the importance of feature representations used for offline policy learning~\cite{yang2021representation, nachum2021provable}. Sec~\ref{exp:hyper} and Appendix~\ref{app:offline-rl} made practical recommendations for choosing hyperparameters to learn from human data.


\textbf{(L5) There is substantial promise for solving more complex tasks using large-scale human datasets.}
Sec~\ref{exp:size} showed that adding more data can result in significant policy improvement on complex tasks. Table~\ref{table:core_image_all} and Sec~\ref{exp:real} shows that we could learn proficient policies on the Tool Hang task, our most complex task, without any hyperparameter tuning on the task or dataset. Together, these results show the potential of large human datasets as a means to solve challenging and complex manipulation tasks.


\textbf{(L6) Study results transfer to real-world settings.}
In Sec~\ref{exp:real}, we showed that we could directly apply hyperparameters that were tuned on simulated tasks directly to real-world datasets and tasks. This provides promise for using our tasks, datasets, and codebase to enable reproducible evaluation in simulation, while also being confident that conclusions can transfer to real-world settings. 


Going forward, we hope that the datasets, tasks, code, and subsequent insights of our study will serve researchers and practitioners alike.

\clearpage
\acknowledgments{We would like to thank Albert Tung for helping with the RoboTurk data collection system, Jim Fan for providing timely lab cluster support, and Helen Roman for helping order items for the physical robot tasks. Ajay Mandlekar acknowledges the support of the Department of Defense (DoD) through the NDSEG program. We acknowledge the support of Toyota Research Institute (“TRI”); this article solely reflects the opinions and conclusions of its authors and not TRI or any other Toyota entity. We acknowledge the support of the US Army Research Office (award W911NF-15-1-0479) and the National Science Foundation (award CNS-1955523). This work relates to Department of Navy award N00014-14-1-0671 issued by the Office of Naval Research.}


\bibliography{benchmark}  

\newpage
\appendix
\part*{Appendix}

\renewcommand\thesection{\Alph{section}}

\counterwithin{figure}{section}

\section{Additional Related Works}
\label{app:related}

\textbf{Human Supervision for Robotics:} Prior work has tried using human supervision in the form of human demonstrations provided both by teleoperating the robot~\cite{zhang2017deep, mandlekar2020learning} and via kinesthetic teaching~\cite{hersch2008dynamical, kormushev2011imitation, akgun2012trajectories}, corrections to actions taken by the robot~\cite{ross2011reduction, bajcsy2017learning, bajcsy2018learning, mandlekar2020human}, and positive and negative feedback for robot actions~\cite{zhang2019leveraging, loftin2016learning, macglashan2017interactive, cabi2019scaling, christiano2017deep, singh2019end}.

\textbf{Offline Learning from Demonstrations}: Imitation Learning (IL) is a popular paradigm for training policies from a set of demonstrations. Offline Imitation Learning typically consists of variants of Behavioral Cloning (BC)~\cite{pomerleau1989alvinn}, where a policy is trained to output the same actions as the ones taken by the demonstrator in each state. Offline IL has been used extensively in robotic manipulation~\cite{Ijspeert2002MovementIW, Finn2017OneShotVI, Billard2008RobotPB, Calinon2010LearningAR, zhang2017deep, mandlekar2020learning, zeng2020transporter, wang2021generalization, tung2020learning}. Typically, IL assumes that the demonstration data is optimal. By contrast, Batch (Offline) Reinforcement Learning~\cite{lange2012batch, levine2020offline} is a method to learn from demonstrations that can consist of both good and bad quality data, by leveraging reward annotations in the datasets. Prior algorithms~\cite{fujimoto2019off, kumar2020conservative, wu2019behavior, wang2020critic, siegel2020keep, kidambi2020morel, yu2020mopo, ghasemipour2020emaq, yu2021combo} are mostly evaluated on datasets generated by several RL-trained policies of varying quality. In this study, we evaluate both offline IL and offline RL algorithms on datasets collected from one or more humans, which can both break the assumption of optimality in IL, and present interesting challenges for offline RL methods compared to commonly used RL-generated datasets. 

\textbf{Empirical Studies in Reinforcement and Imitation Learning}: 
Prior work has benchmarked Reinforcement Learning (RL) algorithms in continuous control domains~\cite{duan2016benchmarking}, run extensive evaluations of model-based RL algorithms~\cite{wang2019benchmarking} and on-policy RL methods~\cite{andrychowicz2020matters}, and shown how Deep RL algorithms can be extremely sensitive to hyperparameter choices and can make reproducing results challenging~\cite{henderson2018deep}. There are fewer empirical studies in imitation learning. The MAGICAL benchmark~\cite{toyer2020magical} consists of 2D environments that test the ability of IL algorithms to generalize. Both RLBench~\cite{james2020rlbench} and Ravens~\cite{zeng2020transporter} provide several simulated robot manipulation tasks and expert demonstrations for imitation learning, but demonstrators are pre-programmed and rely on ground-truth simulator state. 
Simitate~\cite{memmesheimer2019simitate} is an imitation learning benchmark suite consisting of real-world motion trajectories collected by humans, and carefully translated into simulation using extensive sensor instrumentation in the real world. 
By contrast, our datasets are collected via remote teleoperation from humans in both simulated and real-world settings, allowing fast and easy demonstration collection for a wide range of tasks without assuming privileged instrumentation. 
Recently, Hussenot et al.~\cite{hussenot2021hyperparameter} presented an empirical study on the importance of hyperparameter selection in imitation learning -- it consists of an extensive evaluation of both offline and online imitation learning algorithms using modest-sized datasets (\textasciitilde10s of trajectories). This work is complementary to ours -- we also confirm the importance of hyperparameter selection when learning offline from human-provided datasets (\textasciitilde100s of trajectories).

\revision{Prior work has also established benchmarks for motion-based learning from demonstration methods~\cite{rana2020benchmark, lemme2015open}, which directly model entire robot arm trajectories. Lemme et al.~\cite{lemme2015open} evaluates motion generation methods by having a robot arm reproduce several point-to-point motions from demonstrations. This differs from our focus -- tabletop manipulation tasks where a robot must interact with one or more objects. Lemme et al.~\cite{lemme2015open} also evaluate the robustness of the learned motions by introducing perturbations -- similar mechanisms could be used to understand the robustness of the policies trained in this study. While such evaluations are important for deploying policies in real-world settings, we leave this for future work. Similar to our study, recent work by Rana et al.~\cite{rana2020benchmark} collected demonstration trajectories across many humans and tasks. However, their datasets consist of robot end effector trajectories, while our datasets and learning methods focus on leveraging additional modalities such as object poses and camera images to train policies that can solve tasks across several scene configurations. Rana et al.~\cite{rana2020benchmark} also used crowdsourced humans to evaluate learned robot motions with subjective metrics such as safety, while we primarily evaluate our policies using task success rate. Subjective measures like safety are important for real-world policy deployment, but this is also left for future work.}

\textbf{Robot Manipulation Benchmarks}: Several benchmark robot manipulation tasks have been proposed before~\cite{fan2018surreal, lin2020softgym, yu2020meta, robosuite2020, james2020rlbench, ahmed2020causalworld}. Benchmark datasets~\cite{fu2020d4rl, gulcehre2020rl} have also been proposed recently for batch (offline) reinforcement learning. While these datasets span a variety of domains (locomotion, control, autonomous driving, video games, robot manipulation), most of the datasets are generated by autonomous agents trained with online reinforcement learning. Unfortunately, this means that most of these datasets are limited to tasks that RL methods can solve from scratch. By contrast, we focus on datasets collected by one or more humans, allowing us to increase the complexity of the tasks we consider. Furthermore, our study explores how learning from human data can be substantially different than agent datasets.

\newpage
\section{Dataset Details}
\label{app:dataset}

\begin{table}[h!]
  \small
  \centering
  \begin{tabular}{cccccc}
    \toprule
    \textbf{Dataset Type} & \begin{tabular}[c]{@{}c@{}}\textbf{Lift}\end{tabular} &
    \begin{tabular}[c]{@{}c@{}}\textbf{Can}\end{tabular} & 
    \begin{tabular}[c]{@{}c@{}}\textbf{Square}\end{tabular} & 
    \begin{tabular}[c]{@{}c@{}}\textbf{Transport}\end{tabular} &
    \begin{tabular}[c]{@{}c@{}}\textbf{Tool Hang}\end{tabular} \\
    \midrule
    
Machine-Generated (MG) & $150\pm0$ & $150\pm0$ & - & - & - \\
Proficient-Human (PH) & $48\pm6$ & $116\pm14$ & $151\pm20$ & $469\pm54$ & $480\pm88$  \\
Multi-Human (MH) & $104\pm44$ & $209\pm114$ & $269\pm123$ & $653\pm201$ & -  \\
\midrule
MH-Better & $72\pm24$ & $143\pm29$ & $185\pm46$ & $461\pm56$ & -  \\
MH-Okay & $94\pm30$ & $181\pm47$ & $265\pm78$ & $636\pm128$ & -  \\
MH-Worse & $145\pm40$ & $304\pm148$ & $357\pm150$ & $778\pm221$ & -  \\
MH-Worse-Okay & $119\pm44$ & $242\pm126$ & $311\pm128$ & $710\pm115$ & -  \\
MH-Worse-Better & $109\pm49$ & $224\pm134$ & $271\pm140$ & $734\pm297$ & -  \\
MH-Okay-Better & $83\pm29$ & $162\pm44$ & $225\pm76$ & $597\pm83$ & -  \\
    \bottomrule
  \end{tabular}
  \vspace{+3pt}
  \caption{\textbf{Average Trajectory Lengths by Dataset.} The table shows the average trajectory length (mean and standard deviation) for each dataset variant. This was used to determine evaluation rollout horizons for each dataset. The length is a proxy for the quality of the dataset -- less proficient humans took more time to demonstrate the task.}
  \label{table:dataset_stats}
\end{table}
\begin{table}[h!]
  \small
  \centering
  \begin{tabular}{cccccc}
    \toprule
    \textbf{Dataset Type} & \begin{tabular}[c]{@{}c@{}}\textbf{Lift}\end{tabular} &
    \begin{tabular}[c]{@{}c@{}}\textbf{Can}\end{tabular} & 
    \begin{tabular}[c]{@{}c@{}}\textbf{Square}\end{tabular} & 
    \begin{tabular}[c]{@{}c@{}}\textbf{Transport}\end{tabular} &
    \begin{tabular}[c]{@{}c@{}}\textbf{Tool Hang}\end{tabular} \\
    \midrule
    
Machine-Generated(MG) & $400$ & $400$ & - & - & - \\
Proficient-Human(PH) & $400$ & $400$ & $400$ & $700$ & $700$ \\
Multi-Human(MH) & $500$ & $500$ & $500$ & $1100$ & - \\
MH-Better & $500$ & $500$ & $500$ & $1100$ & - \\
MH-Okay & $500$ & $500$ & $500$ & $1100$ & - \\
MH-Worse & $500$ & $500$ & $500$ & $1100$ & - \\
MH-Worse-Okay & $500$ & $500$ & $500$ & $1100$ & - \\
MH-Worse-Better & $500$ & $500$ & $500$ & $1100$ & - \\
    \bottomrule
  \end{tabular}
  \vspace{+3pt}
  \caption{\textbf{Evaluation Rollout Length by Dataset.} The table shows the evaluation rollout horizon for each dataset. These were determined based on the average rollout length of trajectories in each dataset.}
  \label{table:dataset_horizon}
\end{table}

\subsection{Data Collection}

\textbf{Machine-Generated (MG) Datasets.} We trained RL agents to solve tasks and we subsequently collected data from the trained agents to form our MG datasets.
We only considered the Lift and Can tasks, as the other tasks were exceedingly difficult for RL to solve.
Our RL algorithm is based on the Soft Actor-Critic~\citep{haarnoja2018soft} implementation in \href{https://github.com/vitchyr/rlkit}{RLkit}.
We trained the algorithm with episode lengths of $150$ and a batch size of $1024$, and used dense rewards to facilitate exploration.
We trained on the Lift and Can tasks for 2.4 million and 7.2 million environment steps respectively.
For each task we saved agent checkpoints every 600k timesteps in training, which amounts to 5 checkpoints for Lift and 13 checkpoints for Can.
Most checkpoints achieved $0\%$ average task success rate but the last few checkpoints reached an average task success rate of $\sim 70-80\%$.
For generating the datasets, we loaded each saved checkpoint and generated 300 rollouts of a fixed length of 150 timesteps.
We annotated the transitions in the rollouts with sparse task completion reward and a ``done'' signal of \texttt{True} when the corresponding next state in the transition represented task success.
Overall the Lift dataset consists of 225k transitions, and the Can dataset consists of 585k transitions.

\textbf{Proficient-Human (PH) and Multi-Human (MH) Datasets.} Datasets were collected by using the RoboTurk platform~\cite{mandlekar2018roboturk, mandlekar2019scaling}. Our six human operators used smartphones to control robot arms hosted on a simulation server and were provided with video streams in a local web browser. The six operators were located at distances ranging from tens to several thousands of miles from the simulation server, and consisted of two "better" quality operators, two "okay" operators, and two "worse" operators. Since the Transport task requires controlling two arms, we used Multi-Arm RoboTurk~\cite{tung2020learning} to allow pairs of operators to collect data. The Transport (PH) dataset consists of 300 demonstrations collected jointly by the two proficient operators, while the Transport (MH) dataset consists of 6 sets of 50 demonstrations, where each set consists of a pairing of demonstrators (Better-Better, Okay-Okay, Worse-Worse, Worse-Better, Okay-Better, Worse-Okay). As explained in Sec~\ref{exp:suboptimal}, we also further split the MH datasets into smaller subsets to investigate how suboptimal human data affects performance. Unlike the other tasks, all MH data subsets consist of exactly 50 demonstrations, corresponding to a specific pairing of demonstrators. Table~\ref{table:dataset_stats} shows the average trajectory length by dataset, which is a proxy for the quality of the dataset -- less experienced operators produced longer trajectories. 

\textbf{Can-Paired Dataset.} A single experienced operator collected 2 demonstrations for each of 100 task initializations on the Can task, resulting in 200 total demonstrations. Each pair of demonstrations consists of a "good" trajectory, where the can is picked up and placed in the correct bin, and a "bad" trajectory, where the can is picked up, and tossed outside of the robot workspace. Since the task initializations are identical, and the first part of each trajectory leading up to the can grasp is similar, there is a strong expectation for algorithms that deal with suboptimal data, to be able to filter the good trajectories from the bad ones, and achieve near-perfect performance (for reference, BC-RNN can achieve near-perfect performance on the 50\% subset of Can (PH), which corresponds to 100 good quality demos, see Fig~\ref{fig:dataset_size}).

\textbf{Preparing Demonstration Subsets.} In order to prepare smaller datasets used in Sec~\ref{exp:size} and Fig~\ref{fig:dataset_size}, we sampled a fixed portion (20\% and 50\%) of the trajectories uniformly per human that provided data. This ensured that the smaller size datasets were not biased towards higher or lower quality data. We also split all datasets and data subsets into training (90\%) and validation (10\%) subsets, using the same methodology. Models were not trained on the validation subsets.

\subsection{Training Setup}
\label{app:dataset-training-setup}

As mentioned in Sec~\ref{sec:setup}, each agent is trained for $N$ epochs, where each epoch consists of $M$ gradient steps, and evaluated every $E$ epochs, by running 50 rollouts in the environment and reporting the success rate over a maximum horizon. All networks are trained using Adam optimizers~\cite{kingma2014adam}. The average trajectory length in each dataset (Table~\ref{table:dataset_stats}) was used to determine an appropriate evaluation rollout horizon for each dataset (Table~\ref{table:dataset_horizon}). For each agent, we report the maximum success rate over the coarse of training, and average over 3 seeds. For agents trained with low-dimensional observations, $N=2000$, $M=100$, and $E=50$, and for image observations, $N=600$, $M=500$, and $E=20$. With the exception of the MG datasets, agents were only trained over 90\% training subsets, with 10\% held out as validation.

\newpage
\section{Problem Setup and Algorithm Details}
\label{app:algorithm}

Consider a robot manipulation task, formulated as an infinite-horizon discrete-time Markov Decision Process (MDP), $\mathcal{M} = (\mathcal{S}, \mathcal{A}, \mathcal{T}, R, \gamma, \rho_0)$, where $\mathcal{S}$ is the state space, $\mathcal{A}$ is the action space, $\mathcal{T}(\cdot | s, a)$ is the state transition distribution, $R(s, a, s')$ is the reward function, $\gamma \in [0, 1)$ is the discount factor, and $\rho_0(\cdot)$ is the initial state distribution. At every step, an agent observes the state $s_t$, uses a policy $\pi$ to choose an action, $a_t = \pi(s_t)$, and observes the next state, $s_{t+1} \sim \mathcal{T}(\cdot | s_t, a_t)$, and reward, $r_t = R(s_t, a_t, s_{t+1})$. The goal is to learn an policy $\pi$ that maximizes the expected return: $\mathbb{E}[\sum_{t=0}^{\infty} \gamma^t R(s_t, a_t, s_{t+1})]$. In this study, we assume access to an offline dataset of trajectories $\mathcal{D} = \{(s_0^i, a_0^i, r_0^i, s_1^i, ..., s_{T^i}^i)\}_{i=1}^N$~\cite{levine2020offline} and that a policy $\pi_{\theta}$ must be learned offline, without collecting any additional samples from the MDP. 

\revision{\textbf{Rewards and Dones.} Unless otherwise mentioned, the rewards used in this study are binary task completion rewards, $R(s, a, s') = \mathds{1}[s' \in \mathcal{G}]$, where $\mathcal{G} \subset \mathcal{S}$ is the set of all states where the task is considered to be solved. Similarly, the done signal, which indicates the end of an episode, is considered to be true for a transition $(s, a, r, s')$ if the task is solved in state $s'$ or if it is the last transition in a dataset trajectory. The reward and done signals are only used by Batch (Offline) RL algorithms (BCQ, CQL, IRIS).} 

We now outline the Imitation Learning and Batch (Offline) Reinforcement Learning algorithms used in our study.

\subsection{Behavioral Cloning (BC, BC-RNN, HBC)} Behavioral Cloning~\cite{pomerleau1989alvinn} (BC) is a common method for learning a policy from a set of demonstrations. It trains a policy $\pi_{\theta}(s)$ to clone the actions in the dataset via the objective: 
\[
\arg\min_{\theta} \mathbb{E}_{(s, a) \sim \mathcal{D}} ||\pi_{\theta}(s) - a||^2. 
\]

BC-RNN is a variant of BC that uses a Recurrent Neural Network (RNN) as the policy network -- this allows the policy to model temporal dependencies in the data through the recurrent hidden state. The network is trained on length-$T$ temporal sequences of data $(s_t, a_t, ..., s_{t + T}, a_{t + T})$. The network predicts the sequence of actions using the sequence of states as input. At test-time, the RNN policy network is unrolled one-step at a time, $a_t, h_{t+1} = \pi_{\theta}(s_t, h_t)$ where $h$ is the RNN hidden state. The hidden state is refreshed every $T$ steps.

Hierarchical Behavioral Cloning (HBC) trains hierarchical policies and has been shown to be effective in learning from offline human demonstrations~\cite{mandlekar2020iris, mandlekar2020learning, tung2020learning}. HBC consists of a low-level policy that is conditioned on future observations $s_g \in \mathcal{S}$ (termed \textit{subgoals}) and outputs action sequences to try and achieve them, and a high-level policy that predicts future subgoals from the current observation. The architecture and training procedure for the low-level policy $\pi^L_{\theta}(s, s_g)$ is nearly identical to BC-RNN -- the only difference is the subgoal conditioning (during training, this is the final observation of the sampled sequence). The high-level policy $\pi^H_{\theta}(s)$ is trained to predict subgoal observations $s_{t+T}$ that occur $T$ timesteps in the future from the current observation $s_t$, and is often a conditional Variational Autoencoder (cVAE)~\cite{kingma2013auto} that learns a conditional distribution $\pi^H(s_{t+T} | s_t)$. At test-time, the high-level policy is queried for a new subgoal every $T$ timesteps, and the low-level policy is unrolled subsequently for $T$ timesteps using the predicted subgoal.

\subsection{Batch Constrained Q-Learning (BCQ)}
\label{app:algorithm-bcq}

Batch Constrained Q-Learning (BCQ)~\cite{fujimoto2019off} is a commonly used algorithm for batch (offline) reinforcement learning~\cite{lange2012batch, levine2020offline}. It maintains a Q-network $Q_{\psi}(s, a)$, a generative action model $p_{\omega}(a | s)$ (in the original implementation, this is a cVAE~\cite{kingma2013auto}), and (optionally) a perturbation actor network $\pi_{\theta}(s, a)$. Fujimoto et al.~\cite{fujimoto2019off} noted that target value estimation in batch RL can suffer from overestimation error due to querying the Q-network on actions unseen in the dataset. To address this, BCQ modifies the way target value estimates are constructed for the temporal difference Q-network loss by approximately constraining the Q-network maximization to actions seen in the dataset. 

To form the target value estimate, actions are sampled using the generative model and perturbed using the perturbation actor $A = \{a_i + \pi_{\theta}(s, a_i) | a_i \sim p_{\omega}(\cdot | s)\}_{i=1}^N$, and then used to maximize the Q-network at the next state $Q_{\text{target}} = r + \gamma \max_{a_i \in A} Q_{\psi}^{'}(s', a_i)$. The Q-network is trained by minimizing the temporal difference loss $(Q_{\psi}(s, a) - Q_{\text{target}})^2$. As in DDPG~\cite{lillicrap2015continuous}, the perturbation actor is trained to maximize Q-values via the loss $-Q_{\psi}(s, a + \pi_{\theta}(s, a)) | a \sim p_{\omega}(\cdot | s)$. Using the perturbation actor to modify the samples is optional -- we find that removing it is often beneficial (see Appendix~\ref{app:offline-rl}), as in other prior work~\cite{ghasemipour2020emaq}. At test-time, $N$ actions are sampled from the generative action model, perturbed by the actor (if present), and the action with the highest Q-value is selected.

\subsection{Conservative Q-Learning (CQL)}

Conservative Q-Learning (CQL)~\cite{kumar2020conservative} is a recent batch (offline) RL algorithm that addresses the overestimation issue of Q-values directly.
While other offline RL algorithms place constraints on the policy to stay within the support of data, CQL places an implicit constraint on the Q-function that lower-bounds its values. 
Specifically a Q-value regularizer is added to the policy evaluation objective to ensure that the estimated Q-values under the policy $\pi_{\theta}(s, a)$ do not overestimate the Q-values under the data distribution $\mu(a|s)$:
\[
Q^{k+1} \leftarrow \operatorname*{argmin}_Q  \frac{1}{2}(Q(s, a) - Q_{\text{target}})^2 + \alpha \Big( \mathbb{E}_{s\sim\mathcal{D}, a\sim\mu(a|s)}[Q(s, a)] - \mathbb{E}_{s\sim\mathcal{D}, a\sim\pi_{\theta}(a | s)}[Q(s, a)] \Big)
\]
\citet{kumar2020conservative} proved that this formulation lower bounds the true Q function and has theoretical improvement guarantees.
CQL is simple to implement and has shown state-of-the-art empirical results on various offline datasets. 

\subsection{Implicit Reinforcement without Interaction at Scale (IRIS)}

Implicit Reinforcement without Interaction at Scale (IRIS)~\cite{mandlekar2020iris} is a batch (offline) RL algorithm proposed for learning from large robot manipulation datasets collected by multiple humans. It is identical to Hierarchical Behavioral Cloning (HBC) except that the high-level policy consists of both a cVAE subgoal sampler and a value function trained using Batch Constrained Q-Learning (BCQ). Similar to BCQ, the high-level policy selects the subgoal by sampling $N$ subgoals from the cVAE, and picking the state with the highest value estimate. Since the low-level policy is unimodal, all modeling of suboptimal and diverse data takes place in the high-level policy, at a reduced frequency (every $T$ timesteps), enabling temporal abstraction.
\newpage
\section{Hyperparameters}
\label{app:hypers}

\subsection{Network Architecture}
\label{app:network-arch}

Here we briefly overview the general architecture design. Please refer to later sections for more detailed hyperparameter choices.

\textbf{General Network Details.} All Multi-Layer Perceptrons (MLPs) use ReLU activations. All Recurrent Neural Networks (RNNs) are 2-layer LSTMs, where the final layer hidden states are fed to downstream modules.

\textbf{Encoding Observations.} All networks have an observation encoder that processes observation dictionaries into a single vector. The encoder takes image observations, passes each through an observation-specific encoder into a low-dimensional vector, and finally concatenates the encoded image vectors with the low-dimensional observation vectors. For example, visuomotor policies typically contain two image encoders, one for the frontview camera, and one for the wrist camera. Each image encoder consists of a ResNet-18 network~\cite{he2016deep} followed by a spatial-softmax layer~\cite{finn2016deep}.

\textbf{Observation-Conditioned Network Structure.} Here we describe the general structure of networks that take observations as inputs. This includes policies, as well as value and state-action value functions, and comprises the majority of all networks used by the algorithms. First, observations are encoded (if they are images) and concatenated together into a flat vector. Next, in the case of networks that use an RNN (policies for BC-RNN, HBC, and IRIS), the flat encoded observations are sent through the RNN in order to produce hidden state outputs. Finally, outputs are passed to an MLP consisting of one or more layers in order to predict the item of interest. In the case of policy networks, the output either consists of raw actions, or the parameters of an action distribution (e.g. Gaussian Mixture Model parameters), while value functions output a single scalar. Policy action predictions (raw action predictions and mean parameter predictions) are passed through a tanh layer for normalization to [-1, 1]. 

\subsection{Hyperparameter Selection Procedure}

For each algorithm, we tuned hyperparameters separately for the Machine-Generated (MG) datasets, the Proficient-Human (PH) datasets, and the Multi-Human (MH) datasets. We used both the Lift (MG) and Can (MG) datasets for selecting a single set of hyperparameters for use on all MG datasets. We used the Square (PH) and Transport (PH) datasets for selecting a single set of hyperparameters for use on all PH datasets. We used the Square (MH) and Transport (MH) datasets for selecting a single set of hyperparameters for use on all MH datasets. We used Weights and Biases~\cite{wandb} to conduct hyperparameter tuning.

Note that the Tool Hang (sim) and all real tasks were purely used for evaluation purposes -- \textbf{no hyperparameter tuning took place on these tasks}. This was to see whether our insights from hyperparameter tuning above could transfer to our hardest simulation task, and directly to real robot datasets.

Also note that we excluded HBC and IRIS from image-based training, due to the dependence of these algorithms on subgoal reconstructions, which could be problematic for high-dimensional images.


\textbf{BC and BC-RNN.}
We scanned the following hyperparameters for BC and BC-RNN:
\begin{itemize}
    \item \textbf{Learning rate}: We compared $1e-3$ and $1e-4$ for both BC and BC-RNN. We found lower learning rate to perform better consistently.
    \item \textbf{Actor network dimension}: We scanned different dimensions for the action output network (MLP): $[300, 400], [1000, 1000]$, and no MLP (directly output from RNN). We found higher capacity works better for BC, and no MLP works better for RNN. Our hypothesis is that RNN already has enough capacity to learn the tasks, and larger actor network results in overfitting. 
    \item \textbf{GMM for action output}: Stochastic policy (GMM output) performs better than deterministic policy, although the gap is smaller for BC-RNN than for BC. We scanned different number of modes: $\{5, 10, 100\}$ standard deviation minimum clipping:  $\{1e-2, 1e-4, 1e-6\}$ and observed no significant difference in performance.
    \item \textbf{Sequence length (RNN)}: We compared sequence length $\{10, 30, 50\}$. We found longer sequence length does not improve performance significantly. We opt for short sequence length for training efficiency.
    \item \textbf{RNN Hidden Dim}: We compared different hidden dimension sizes for the RNN (LSTM): $\{100, 400, 1000, 2000\}$. We observed that 400 and 1000 tend to work well for most tasks. 
\end{itemize}

\textbf{BCQ.}
As in the original implementation~\cite{fujimoto2019off}, we used two critic networks. When using the perturbation actor, we also left the scale unchanged from the original (0.05). When using the VAE action sampler, we used a latent dimension of 14. We scanned the following hyperparameters:
\begin{itemize}
    \item \textbf{Learning rate}: For the learning rate of the critic (CLR) and the action sampler (ASLR), we compared $1e-3$ and $1e-4$. We found smaller learning rate for the action sampler usually has better performance on human datasets, while there was not much difference between the two learning rates for the critic.
    \item \textbf{Actor/Critic network dimension}: We compared the networks with layer dimensions $[300, 400]$ and $[1024, 1024]$. We found there is no difference for them with Machine-Generated (MG) and Proficient-Human (PH) datasets. For Multi-Human (MH) dataset, larger network dimension $[1024, 1024]$ has better results.
    \item \textbf{Perturbation actor}: We compare the results of whether use the perturbation actor for the BCQ action sampling\cite{lillicrap2015continuous}. As is shown in Table. \ref{table:hyper-bcq-actor}, we found the performance drastically decreases when using the actor. By contrast, with the machine-generated data (MG), we found BCQ with actor enabled usually has better performance.
    \item \textbf{Action sampler (VAE vs GMM)}: We compared two types of action sampler - the VAE and GMM and found that the VAE generally has better performance than the GMM sampler. However, the GMM sampler also allows for a direct comparison with BC -- see Table~\ref{table:hyper-bcq-gmm} for results.
    \item \textbf{VAE KL}: We scanned $\{5e-1, 5e-2, 5e-3, 5e-4\}$ the weight of KL for the VAE action sampler and found that larger KL weights ${5e-1, 5e-2}$ show better performance.
    \item \textbf{VAE Layer Dims:}: For the VAE action sampler, we compared the encoder / decoder / prior layer dimensions $[300, 400]$ and $[1024, 1024]$. We found there is no significant difference for them in low-dimensional tasks. With image observation input, larger dimension $[1024, 1024]$ outperforms $[300, 400]$.
    \item \textbf{VAE prior}: While the VAE prior is commonly a normal prior $N(0, 1)$, it can also be learned as part of the KL loss. We compared using the normal prior to using a state-dependent GMM prior, whose parameters are output by an MLP. We found that the normal prior $N(0, 1)$ generally performs better than the learned GMM prior, except for the Multi-Human(MH) dataset. Therefore, we opted the learned GMM prior only for Multi-Human(MH) dataset and keep $N(0, 1)$ prior for the others.
    \item \textbf{Tau (target network update rate)}: We compared $5e-3$ and $5e-4$ and found the lower value of $5e-4$ to give better results.
    \item \textbf{Num action samples (train/test)}: During training, BCQ requires generating a number of samples from the action sampler to do the Bellman backup, while at test-time, BCQ requires generating action samples to approximately maximize the critic over the samples to choose an appropriate action. We compared $[10, 100]$ and $[100, 1000]$ for the number of action samples during training and testing and found that $[10, 100]$ works better.
\end{itemize}

\textbf{CQL.}
In contrast to other algorithms, we primarily tuned CQL hyperparameters on the low-dim Lift (MG) dataset, due to poor performance on the harder Square (PH) and Transport (PH) datasets. In subsequent experiments, we found that these hyperparameter settings worked better than other choices on other datasets as well. For all experiments we used a discount factor of $\gamma = 0.99$, a target network update rate of $\tau=0.005$, and actor / critic network layer sizes of $[300, 400]$. We scanned the following hyperparameters:
\begin{itemize}
    \item \textbf{Learning rate}: For low-dim experiments we scanned an extensive sweep of learning rates for the Q network and the policy, spanning values of $\{1e-5, 3e-5, 1e-4, 3e-4, 1e-3, 3e-3, 1e-2\}$. For the Q network we found that $1e-3$ performed best --- lower values slowed down learning while higher values led to unstable learning.
    For the policy we found that $3e-4$ and $1e-3$ worked well, though $3e-4$ performed slightly better.
    For image experiments we performed a smaller sweep and found that $1e-4$ for both the Q network and policy performed relatively well compared to other learning rates.
    \item \textbf{Deterministic backup}: As suggested by the CQL implemented from \citet{kumar2020conservative}, we experimented with a deterministic Bellman backup objective --- i.e.\ removing the additional entropy term from the target Q value backup.
    We found that the deterministic backup outperformed the non-deterministic backup, though at times the gains were marginal.
    We subsequently chose to deterministic variant.
    \item \textbf{BC start steps}: We experimented with replacing the policy loss objective for CQL with the behavior cloning objective for the first $40000$ gradient steps of training. We found that the performance of the agent degraded after these initial gradient steps, so we set this hyperparameter to $0$ in all subsequent experiments.
    \item \textbf{Batch size}: For low-dim experiments, we found that increasing the batch size from the default value of $100$ to $1024$ can lead to significant gains in stability and performance. For image experiments we were bottlenecked by GPU memory and we used a batch size of $8$.   
    \item \textbf{Lagrange variant}: We found that the Lagrange variant consistently performed better than the non-Lagrange variant and was less sensitive to hyperparameters. We therefore decided to use the Lagrange variant in all of our experiments.
    \item \textbf{Lagrange threshold} $\tau$: We found that a threshold of $\tau=1$ often caused the dual weight to diverge to large values. We subsequently experimented with higher values of $5,10,25$ and found that the algorithm was relatively stable with all of these with no major difference in performance. We subsequently chose $\tau=5$. 
\end{itemize}

\textbf{HBC.}
We scanned the following hyperparameters for HBC:
\begin{itemize}
    \item \textbf{Learning Rate:} We scanned ${1e-3, 1e-4}$ for both the policy and goal learning rate, and generally found the higher learning rates to perform better.
    \item \textbf{VAE vs AE:} We compared using a VAE vs. only an AE for the planner network, and found the best performing VAE model outperformed the best performing AE model.
    \item \textbf{VAE KL:} We compared using ${5e-1, 5e-2, 5e-3, 5e-4}$, and found that $5e-4$ worked the best.
    \item \textbf{VAE Prior:} We compared using ${\mathcal{N}(0, 1), \text{Learned GMM}}$ priors, and found that, when tuned, the GMM prior worked the best.
    \item \textbf{VAE Latent Dim:} We compared ${2, 16, 100}$, and found that $16$ worked the best.
    \item \textbf{VAE Layer Dims:} We compared ${[300, 400], [1024, 1024]}$ as the encoder / decoder / prior layer dimensions, and found that $[1024, 1024]$ worked the best.
    \item \textbf{RNN Hidden Dim:} We compared ${100, 400, 1000, 2000}$ and found that $400$ worked the best.
    \item \textbf{Actor MLP Dims:} We compared ${[], [300, 400], [1024, 1024]}$ and found that no hidden layers ($[]$) worked the best.
\end{itemize}

\textbf{IRIS.}
We first initialize our scan with the best hyperparameters from HBC and BCQ. Note that from our BCQ scan, we use $\textbf{Action Sampler Learning Rate} = 1e-4$ and $\textbf{Num Action Samples (Train/Test)} = 10 / 100$. We also use $\textbf{Tau} = 5e-4$ and no value actor for PH and MH datasets and $\textbf{Tau} = 5e-3$ with value actor enabled for MG datasets. We then proceeded to scan over the following hyperparameters:

\begin{itemize}
    \item \textbf{Learning Rate:} We compared ${1e-3, 1e-4}$ for the policy, goal, and value learning rates individually, and generally found the higher learning rates to perform better. The exception is the multi-human setting, where we found a lower Value LR to work better.
    \item \textbf{Value KL Weight:} We compared ${0.5, 0.05}$, and found that both values generally performed similarly.
    \item \textbf{Value Actor:} We compared using a value actor ${\text{True}, \text{False}}$ on the MG datasets, and found that an actor can improve results on the Can MG dataset.
\end{itemize}

\subsection{Final Hyperparameters}
\label{app:final-hyper}

We present our finalized set of hyperparameters in Tables \ref{tab:hypers-bc-ld} and \ref{tab:hypers-bc-im} for BC, Tables \ref{tab:hypers-bc-rnn-ld} and \ref{tab:hypers-bc-rnn-im} for BC-RNN, Tables \ref{tab:hypers-bcq-ld} and \ref{tab:hypers-bcq-im} for BCQ, Table \ref{tab:hypers-hbc-ld} for HBC, Table \ref{tab:hypers-iris-ld} for IRIS, and Tables \ref{tab:hypers-cql-ld} and \ref{tab:hypers-cql-im} for CQL. Each column shows the hyperparameters for learning from a dataset setting (shared across environments): PH for proficient human, MH for multiple humans, MG for machine-generated. \colorbox{gray!15}{-} means the hyperparameter is inherited from the default hyperparameters shown on the left-most column. For further details on the training setup shared by all algorithms (such as the number of gradient steps and epochs), see Appendix~\ref{app:dataset-training-setup}.


\begin{table*}[h!]
  \fontsize{8}{10}\selectfont
  \caption{\textbf{BC Hyperparameters - Low-Dim (LD)}}
  \label{tab:hypers-bc-ld}
  \centering
  \begin{tabular}{ccccc}
    \toprule
    \multirow{2}{*}{Hyperparameter}
    & \multirow{2}{*}{Default}
    & \multicolumn{3}{c}{Dataset} \\
    & & PH & MH & MG \\
    \midrule
    LR & $1e-4$ & - & - & $1e-3$\\
    Actor MLP Dims & $[1024, 1024]$ & - & - & -\\
    GMM Num Modes & $5$ & - & - & no-gmm\\
    \bottomrule
  \end{tabular}
\end{table*}

\begin{table*}[h!]
  \fontsize{8}{10}\selectfont
  \caption{\textbf{BC Hyperparameters - Image (IM)}}
  \label{tab:hypers-bc-im}
  \centering
  \begin{tabular}{ccccc}
    \toprule
    \multirow{2}{*}{Hyperparameter}
    & \multirow{2}{*}{Default}
    & \multicolumn{3}{c}{Dataset} \\
    & & PH & MH & MG \\
    \midrule
    LR & $1e-4$ & - & - & - \\
    Actor MLP Dims & $[1024, 1024]$ & - & - & -\\
    GMM Num Modes & $5$ & - & - & no-gmm \\
    Image Encoder & ResNet-18 & - & - & - \\
    SpatialSoftmax~\cite{finn2016deep} (num-KP)& $64$ & - & - & -  \\
    \bottomrule
  \end{tabular}
\end{table*}

\begin{table*}[h!]
  \fontsize{8}{10}\selectfont
  \caption{\textbf{BC-RNN Hyperparameters - Low-Dim (LD)}}
  \label{tab:hypers-bc-rnn-ld}
  \centering
  \begin{tabular}{ccccc}
    \toprule
    \multirow{2}{*}{Hyperparameter}
    & \multirow{2}{*}{Default}
    & \multicolumn{3}{c}{Dataset} \\
    & & PH & MH & MG \\
    \midrule
    LR & $1e-4$ & - & - & - \\
    Actor MLP Dims & $[]$ & - & - & -\\
    RNN Hidden Dim & $400$ & - & - & -\\
    RNN Seq Len & $10$ & - & - & - \\
    GMM Num Modes & $5$ & - & - & no-gmm \\
    \bottomrule
  \end{tabular}
\end{table*}

\begin{table*}[h!]
  \fontsize{8}{10}\selectfont
  \caption{\textbf{BC-RNN Hyperparameters - Image (IM)}}
  \label{tab:hypers-bc-rnn-im}
  \centering
  \begin{tabular}{ccccc}
    \toprule
    \multirow{2}{*}{Hyperparameter}
    & \multirow{2}{*}{Default}
    & \multicolumn{3}{c}{Dataset} \\
    & & PH & MH & MG \\
    \midrule
    LR & $1e-4$ & - & - & - \\
    Actor MLP Dims & $[]$ & - & - & -\\
    RNN Hidden Dim & $1000$ & - & - & -\\
    RNN Seq Len & $10$ & - & - & - \\
    GMM Num Modes & $5$ & - & - & no-gmm \\
    Image Encoder & ResNet-18 & - & - & - \\
    SpatialSoftmax~\cite{finn2016deep} (num-KP)& $64$ & - & - & -  \\
    \bottomrule
  \end{tabular}
\end{table*}

\begin{table*}[h!]
  \fontsize{8}{10}\selectfont
  \caption{\textbf{BCQ Hyperparameters - Low Dim (LD)}}
  \label{tab:hypers-bcq-ld}
  \centering
  \begin{tabular}{ccccc}
    \toprule
    \multirow{2}{*}{Hyperparameter}
    & \multirow{2}{*}{Default}
    & \multicolumn{3}{c}{Dataset} \\
    & & PH & MH & MG \\
    \midrule
    CLR & $1e-4$ & - & - & $1e-3$ \\
    ASLR & $1e-4$ & - & - & $1e-3$ \\
    Critic Dims & $[300, 400]$ & - & $[1024, 1024]$ & - \\
    Actor Dims & $[300, 400]$ & - & $[1024, 1024]$ & - \\
    Perturb Actor & False & - & - & True \\
    Action Sampler & VAE & - & - & - \\
    VAE KL & $5e-2$ & - & $5e-1$ & $5e-1$\\
    VAE Dims & $[300, 400]$ & - & $[1024, 1024]$ & -\\
    VAE Prior & $N(0, 1)$ & - & GMM & -\\
    Tau & $5e-4$ & - & - & $5e-3$ \\
    Num Action Samples & $[10, 100]$ & - & - & - \\
    \bottomrule
  \end{tabular}
\end{table*}

\begin{table*}[h!]
  \fontsize{8}{10}\selectfont
  \caption{\textbf{BCQ Hyperparameters - Image (IM)}}
  \label{tab:hypers-bcq-im}
  \centering
  \begin{tabular}{ccccc}
    \toprule
    \multirow{2}{*}{Hyperparameter}
    & \multirow{2}{*}{Default}
    & \multicolumn{3}{c}{Dataset} \\
    & & PH & MH & MG \\
    \midrule
    CLR & $1e-4$ & $1e-3$ & - & $1e-3$ \\
    ASLR & $1e-4$ & - & - & $1e-3$ \\
    Critic Dims & $[300, 400]$ & - & $[1024, 1024]$ & - \\
    Actor Dims & $[300, 400]$ & - & $[1024, 1024]$ & - \\
    Perturb Actor & False & - & - & - \\
    Action Sampler & VAE & - & - & - \\
    VAE KL & $5e-2$ & - & - & $5e-1$\\
    VAE Dims & $[1024, 1024]$ & - & - & -\\
    VAE Prior & $N(0, 1)$ & - & GMM & -\\
    Tau & $5e-4$ & - & - & $5e-3$ \\
    Num Action Samples & $[10, 100]$ & - & - & - \\
    \bottomrule
  \end{tabular}
\end{table*}


\begin{table*}[h!]
  \fontsize{8}{10}\selectfont
  \caption{\textbf{HBC Hyperparameters - Low Dim (LD)}}
  \label{tab:hypers-hbc-ld}
  \centering
  \begin{tabular}{ccccc}
    \toprule
    \multirow{2}{*}{Hyperparameter}
    & \multirow{2}{*}{Default}
    & \multicolumn{3}{c}{Dataset} \\
    & & PH & MH & MG \\
    \midrule
    Planner LR & $1e-3$ & - & - & -\\
    Planner VAE KL & $5e-4$ & - & - & - \\
    Planner VAE GMM Prior & True & - & - & - \\
    Planner VAE GMM Latent Dim & $16$ & - & - & - \\
    Planner VAE MLP Dims & $[1024, 1024]$ & - & -  & - \\
    Actor LR & $1e-3$ & - & - & - \\
    Actor RNN Hidden Dim & $400$ & - & - & 100 \\
    Actor MLP Dims & $[]$ & - & - & $[1024, 1024]$ \\
    \bottomrule
  \end{tabular}
\end{table*}


\begin{table*}[h!]
  \fontsize{8}{10}\selectfont
  \caption{\textbf{IRIS Hyperparameters - Low Dim (LD)}}
  \label{tab:hypers-iris-ld}
  \centering
  \begin{tabular}{ccccc}
    \toprule
    \multirow{2}{*}{Hyperparameter}
    & \multirow{2}{*}{Default}
    & \multicolumn{3}{c}{Dataset} \\
    & & PH & MH & MG \\
    \midrule
    Planner LR & $1e-3$ & - & - & - \\
    Planner VAE KL & $5e-4$ & - & - & - \\
    Planner VAE GMM Prior & True & - & - & - \\
    Planner VAE GMM Latent Dim & $16$ & - & - & - \\
    Planner VAE MLP Dims & $[1024, 1024]$ & - & - & - \\
    Actor LR & $1e-3$ & - & - & - \\
    Actor RNN Hidden Dim & $400$ & - & - & - \\
    Actor MLP Dims & $[]$ & - & - & - \\
    Value LR & $1e-3$ & - & $1e-4$ & - \\
    Value KL & $0.5$ & - & $0.05$ & - \\
    \bottomrule
  \end{tabular}
\end{table*}

\begin{table*}[h!]
  \fontsize{8}{10}\selectfont
  \caption{\textbf{CQL Hyperparameters - Low Dim (LD)}}
  \label{tab:hypers-cql-ld}
  \centering
  \begin{tabular}{ccccc}
    \toprule
    \multirow{2}{*}{Hyperparameter}
    & \multirow{2}{*}{Default}
    & \multicolumn{3}{c}{Dataset} \\
    & & PH & MH & MG \\
    \midrule
    Q network LR & $1e-3$ & - & - & - \\
    Policy LR & $3e-4$ & - & - & -\\
    Deterministic Backup & True & - & - & - \\
    BC Start Steps & 0 & - & - & - \\
    Batch Size & $1024$ & - & - & - \\
    Lagrange & True & - & - & - \\
    Lagrange Threshold $\tau$ & $5.0$ & - & - & - \\
    Actor MLP Dims & $[300, 400]$ & - & - & - \\
    \bottomrule
  \end{tabular}
\end{table*}

\begin{table*}[h!]
  \fontsize{8}{10}\selectfont
  \caption{\textbf{CQL Hyperparameters - Image (IM)}}
  \label{tab:hypers-cql-im}
  \centering
  \begin{tabular}{ccccc}
    \toprule
    \multirow{2}{*}{Hyperparameter}
    & \multirow{2}{*}{Default}
    & \multicolumn{3}{c}{Dataset} \\
    & & PH & MH & MG \\
    \midrule
    Q network LR & $1e-4$ & - & - & - \\
    Policy LR & $1e-4$ & - & - & - \\
    Deterministic Backup & True & - & - & -\\
    BC Start Steps & 0 & - & - & - \\
    Batch Size & $8$ & - & - & - \\
    Lagrange & True & - & - & - \\
    Lagrange Threshold $\tau$ & $5.0$ & - & - & - \\
    Actor MLP Dims & $[300, 400]$ & - & - & - \\
    \bottomrule
  \end{tabular}
\end{table*}


\subsection{Additional Details on Hyperparameter Choice Study for BC-RNN}

In Sec~\ref{exp:hyper}, Fig~\ref{fig:hyper-low-dim}, and Fig~\ref{fig:hyper-image}, we presented results that showcase how changing a subset of BC-RNN hyperparameters can result in large performance decreases. In this section, we provide more details for each change.

\textbf{Larger LR.} We changed the policy learning rate from the default 1e-4 to 1e-3.

\textbf{No GMM.} By default, all BC-RNN policies on PH and MH learned a Gaussian Mixture Model (GMM) distribution. Here, we replace the GMM distribution with a direct action prediction.

\textbf{Larger MLP.} As noted in Appendix~\ref{app:network-arch} above, there is an MLP that transforms RNN hidden states into action (or action distribution) predictions. By default, we use a single linear layer, but here we try adding two layers of size 1024. 

\textbf{Shallow Conv.} As noted in Appendix~\ref{app:network-arch} above, all image encoders are ResNet-18 networks. Here, we tried replacing the ResNet with the shallow convolutional network from Finn et al.~\cite{finn2016deep}.

\textbf{Smaller RNN Dim.} By default, we use a hidden layer size of 400 for low-dimensional datasets, and 1000 for image datasets. Here, we tried reducing the dimension to 100 for low-dim, and 400 for image.

\newpage
\section{Additional Details on Task Environments}
\label{app:task}

\begin{figure}[h!]

\begin{subfigure}{0.16\linewidth}
\centering
\includegraphics[width=1.0\textwidth]{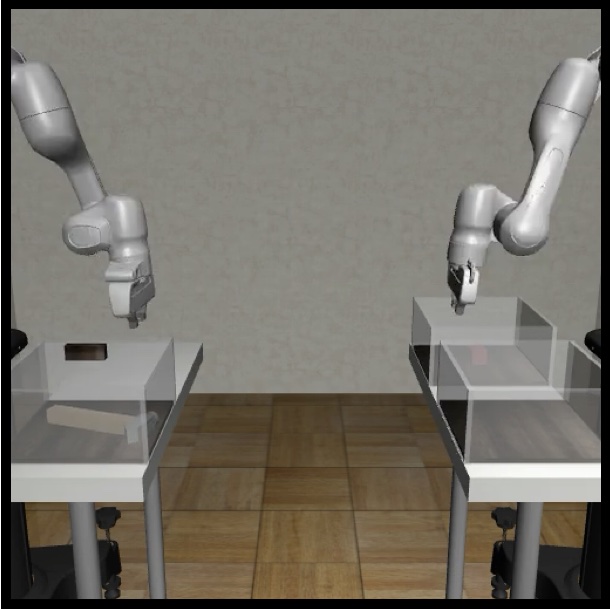}
\end{subfigure}
\hfill
\begin{subfigure}{0.16\linewidth}
\centering
\includegraphics[width=1.0\textwidth]{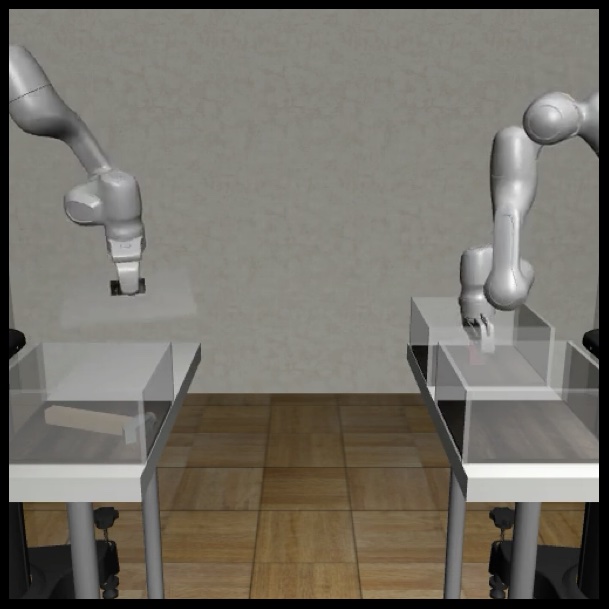}
\end{subfigure}
\hfill
\begin{subfigure}{0.16\linewidth}
\centering
\includegraphics[width=1.0\textwidth]{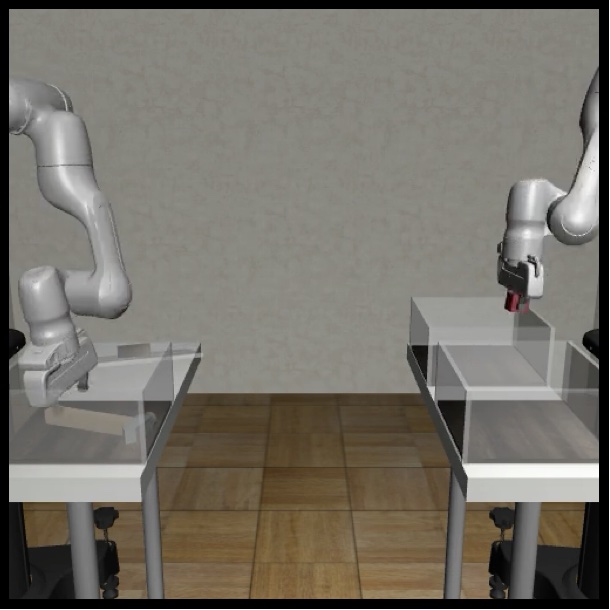}
\end{subfigure}
\hfill
\begin{subfigure}{0.16\linewidth}
\centering
\includegraphics[width=1.0\textwidth]{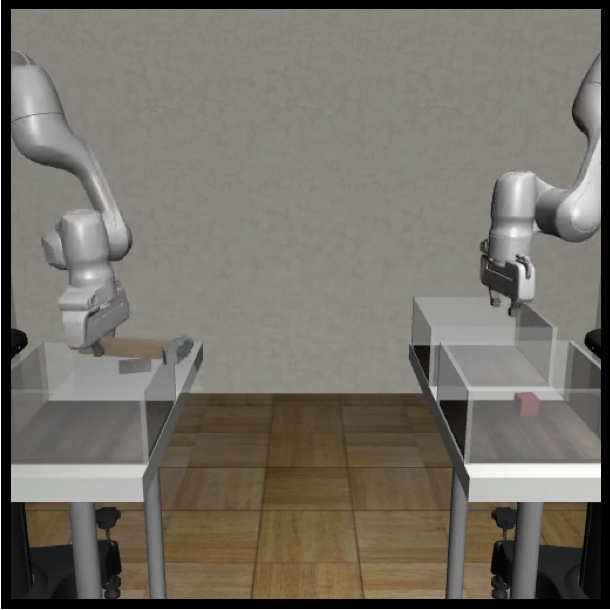}
\end{subfigure}
\hfill
\begin{subfigure}{0.16\linewidth}
\centering
\includegraphics[width=1.0\textwidth]{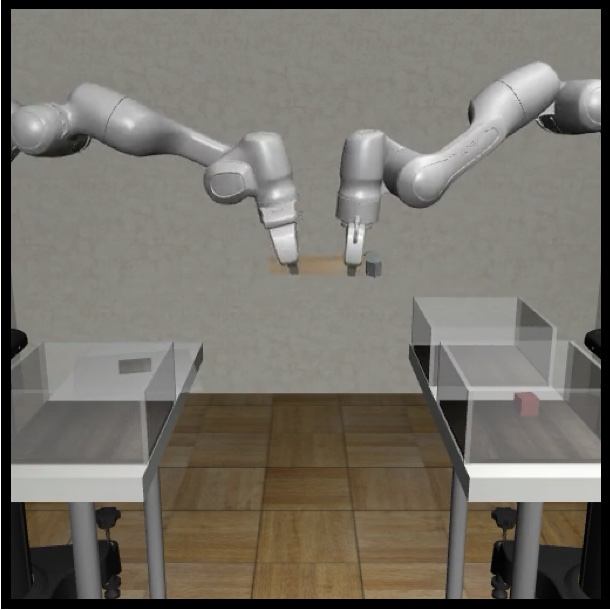}
\end{subfigure}
\hfill
\begin{subfigure}{0.16\linewidth}
\centering
\includegraphics[width=1.0\textwidth]{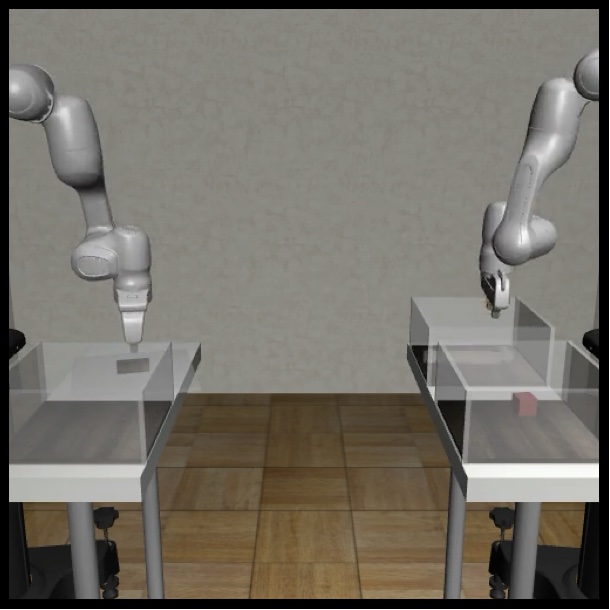}
\end{subfigure}
\hfill

\begin{subfigure}{0.16\linewidth}
\centering
\includegraphics[width=1.0\textwidth]{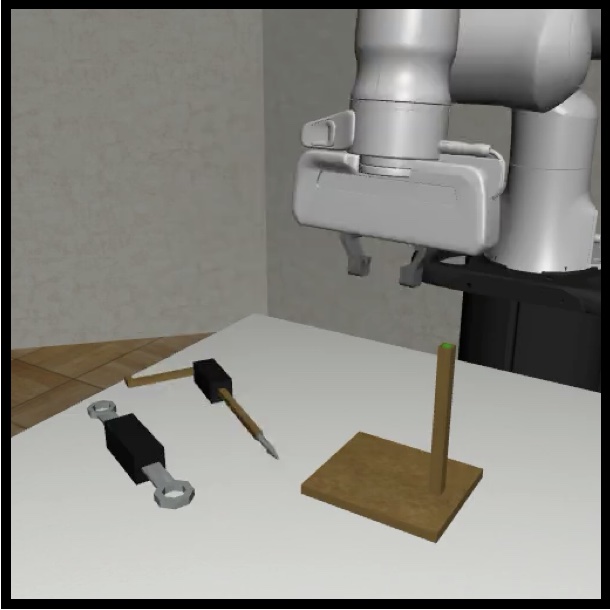}
\end{subfigure}
\hfill
\begin{subfigure}{0.16\linewidth}
\centering
\includegraphics[width=1.0\textwidth]{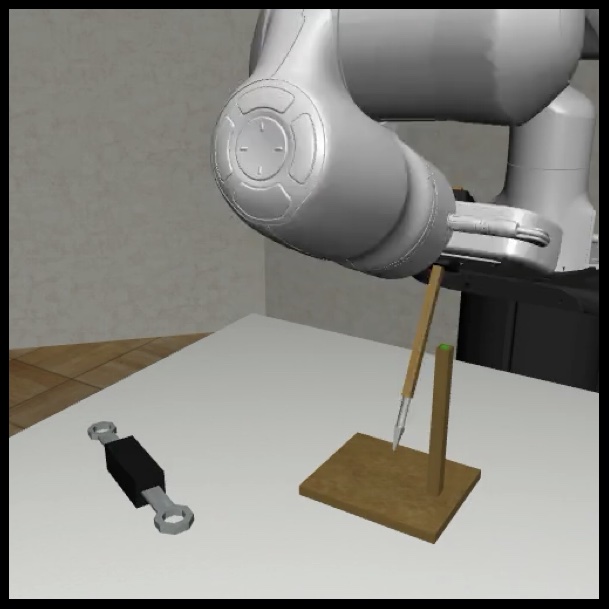}
\end{subfigure}
\hfill
\begin{subfigure}{0.16\linewidth}
\centering
\includegraphics[width=1.0\textwidth]{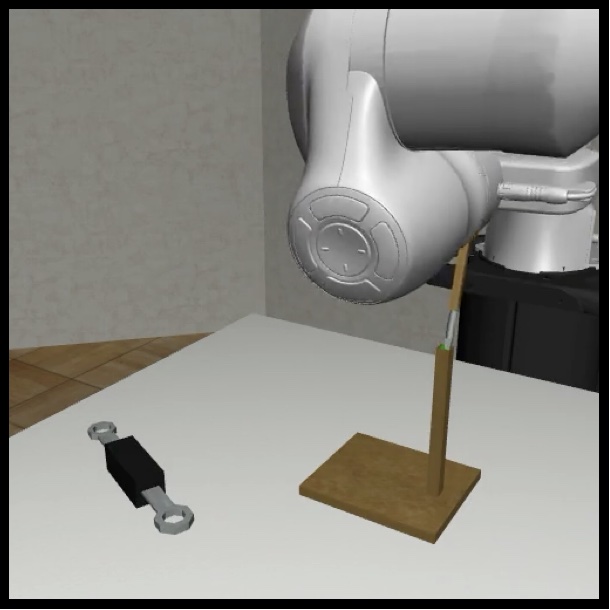}
\end{subfigure}
\hfill
\begin{subfigure}{0.16\linewidth}
\centering
\includegraphics[width=1.0\textwidth]{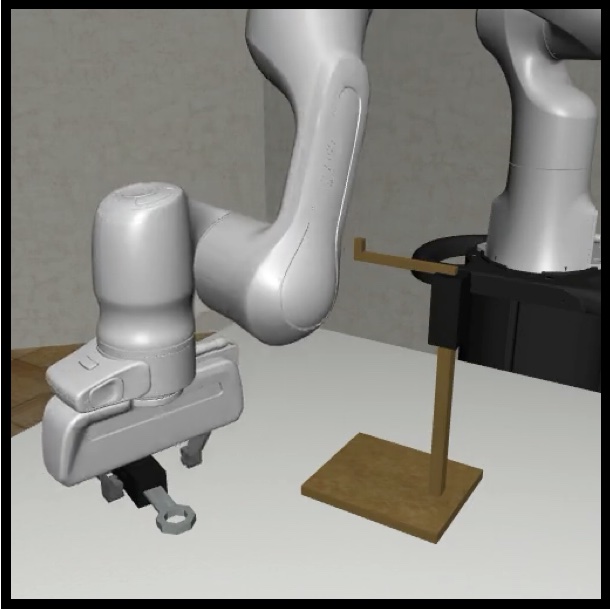}
\end{subfigure}
\hfill
\begin{subfigure}{0.16\linewidth}
\centering
\includegraphics[width=1.0\textwidth]{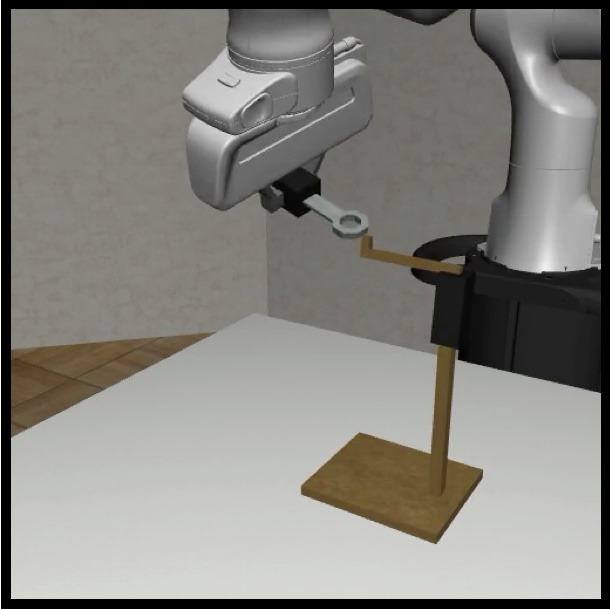}
\end{subfigure}
\hfill
\begin{subfigure}{0.16\linewidth}
\centering
\includegraphics[width=1.0\textwidth]{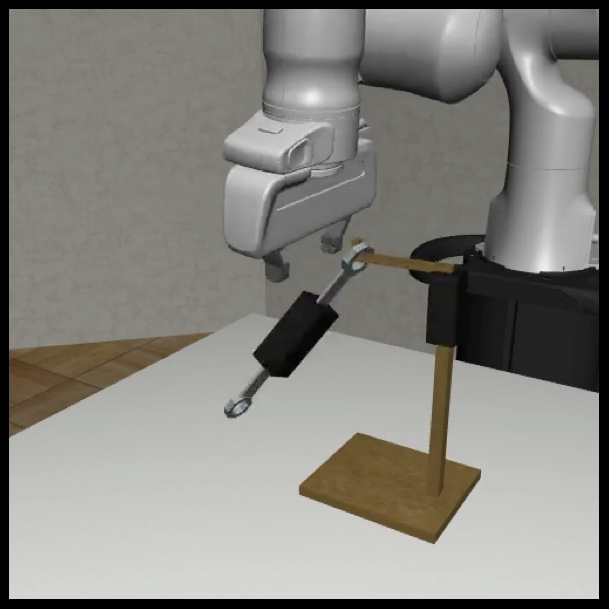}
\end{subfigure}
\hfill

\begin{subfigure}{0.16\linewidth}
\centering
\includegraphics[width=1.0\textwidth]{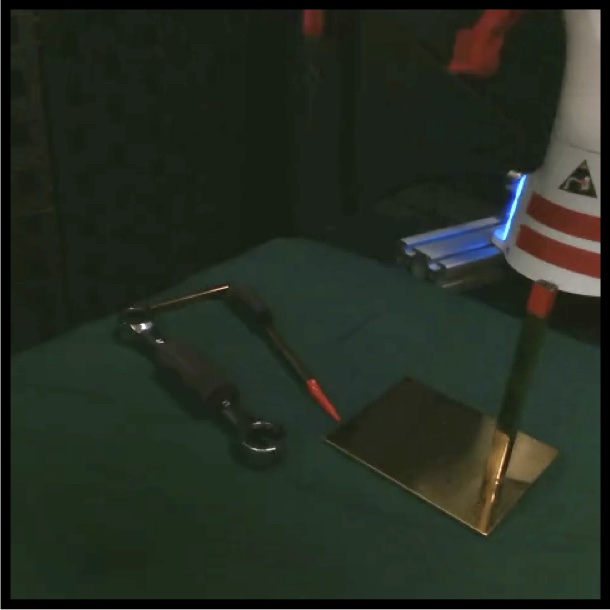}
\end{subfigure}
\hfill
\begin{subfigure}{0.16\linewidth}
\centering
\includegraphics[width=1.0\textwidth]{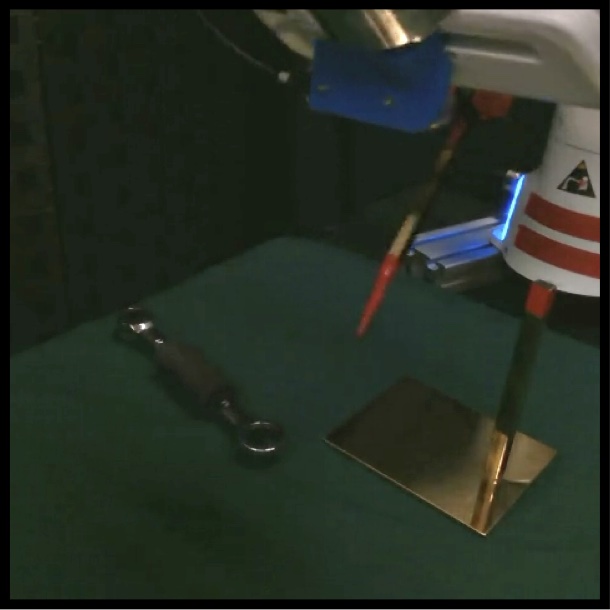}
\end{subfigure}
\hfill
\begin{subfigure}{0.16\linewidth}
\centering
\includegraphics[width=1.0\textwidth]{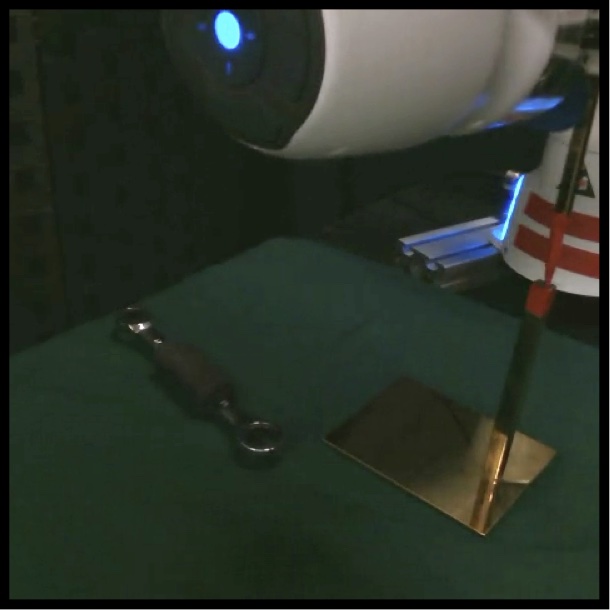}
\end{subfigure}
\hfill
\begin{subfigure}{0.16\linewidth}
\centering
\includegraphics[width=1.0\textwidth]{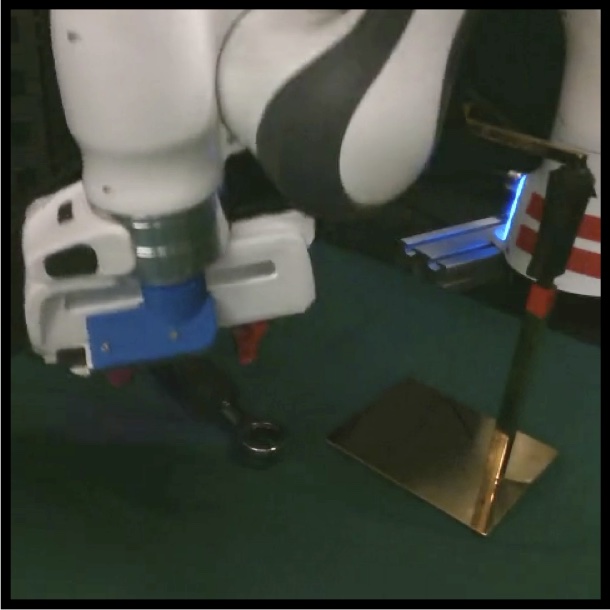}
\end{subfigure}
\hfill
\begin{subfigure}{0.16\linewidth}
\centering
\includegraphics[width=1.0\textwidth]{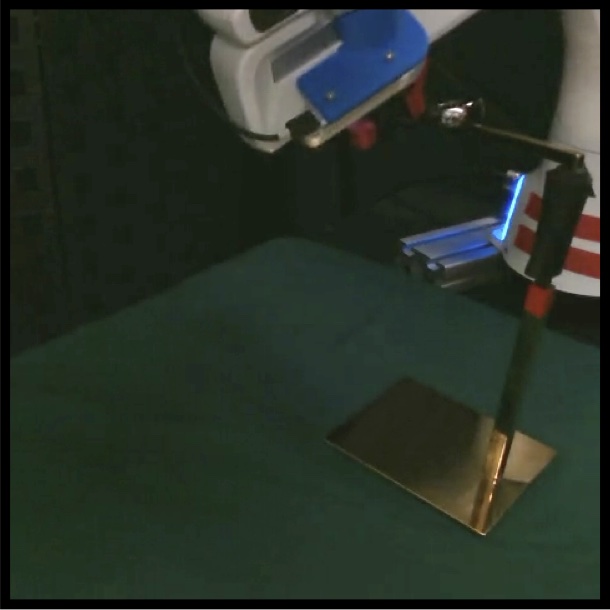}
\end{subfigure}
\hfill
\begin{subfigure}{0.16\linewidth}
\centering
\includegraphics[width=1.0\textwidth]{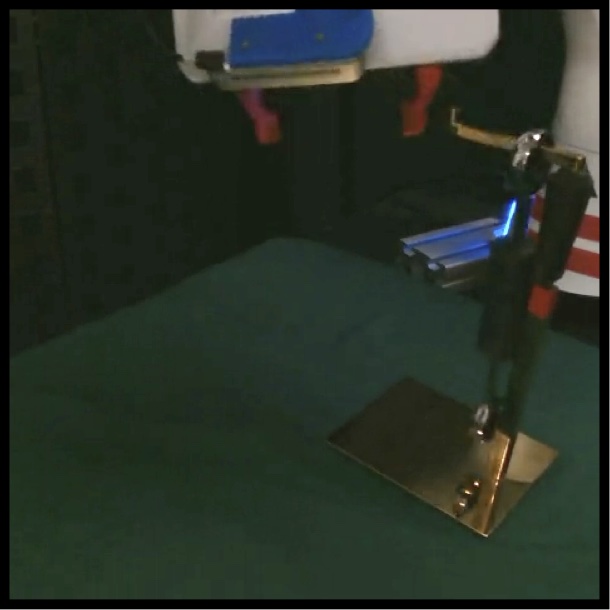}
\end{subfigure}
\hfill

\caption{\textbf{Task Demonstrations.} We showcase example demonstration trajectories for the Transport (top), Tool Hang (middle), and Tool Hang (Real) (bottom) tasks, to provide a better sense of each stage of these tasks.}
\label{fig:task_long}
\end{figure}

\subsection{Simulation Tasks}
\label{app:task-sim}

All simulation tasks were designed using MuJoCo~\cite{todorov2012mujoco} and the robosuite framework~\cite{robosuite2020}. We used Panda robotic arms in both simulation and the real world for this study. The action space for the agent is a 7-dimensional vector for each arm where the first 3 coordinates are the desired translation from the current end effector position, the next 3 coordinates encoder the desired delta rotation from the current end effector rotation, and the final coordinate controls the opening and closing of the gripper fingers. The delta rotation is encoded in axis-angle form, where the norm of the 3-vector is the angle, and normalizing the 3-vector produces the axis. The policy outputs actions at a rate of 20 Hz. Policy actions are transformed into end effector target poses and sent to an operational space controller~\cite{khatib1987unified} that outputs the robot joint torques at 500 Hz to try and achieve the desired cartesian poses.

All simulation environments will be released along with datasets and codebase upon publication. We next describe the low-dimensional object observations and initial state randomization for each task. 

\textbf{Lift.} Object observations (10-dim) consist of the absolute cube position and cube quaternion (7-dim), and the cube position relative to the robot end effector (3-dim). The cube pose is randomized at the start of each episode with a random z-rotation in a small square region at the center of the table.

\textbf{Can.} Object observations (14-dim) consist of the absolute can position and quaternion (7-dim), and the can position and quaternion relative to the robot end effector (7-dim). The can pose is randomized at the start of each episode with a random z-rotation anywhere inside the left bin.

\textbf{Square.} Object observations (14-dim) consist of the absolute square nut position and quaternion (7-dim), and the square nut position and quaternion relative to the robot end effector (7-dim). The square nut pose is randomized at the start of each episode with a random z-rotation in a square region on the table.

\textbf{Transport.} Fig~\ref{fig:task_long} (top) shows a full demonstration of the task. Object observations (41-dim) consist of the absolute position and quaternion of the hammer (7-dim), the absolute position and quaternion of the trash cube (7-dim), the absolute position and quaternion of the lid handle (7-dim), the target bin position (3-dim), the trash bin position (3-dim), the relative positions of the hammer and the lid handle with respect to the first arm end effector (6-dim), the relative positions of the hammer and trash cube with respect to the second arm end effector (6-dim), a binary indicator for the hammer reaching the target bin (1-dim), and a binary indicator for the trash reaching the trash bin (1-dim). The position of all bins, the lid, the trash cube, and the hammer are randomized in small squares at the start of each episode. The z-rotation of the trash cube and the hammer are also randomized with a full range of 108 and 60 degrees respectively.

\textbf{Tool Hang.} Fig~\ref{fig:task_long} (middle) shows a full demonstration of the task. Object observations (44-dim) consist of the absolute position and quaternion and relative pose and quaternion with respect to the end effector of the base frame (14-dim), the insertion hook (14-dim), and the ratcheting wrench (14-dim), as well as binary indicators for whether the stand was assembled (1-dim) and whether the tool was successfully placed on the stand (1-dim). The position of the insertion hook and ratcheting wrench and z-rotation (range of 40 degrees) are randomized in a small square at the beginning of the episode.

\subsection{Real World Task Setup} 

We first describe details about the robot workspace and setup. Next, we discuss the materials needed to construct each physical task. We took care to approximately match the visual appearance, the physical dimensions, and the task initialization randomizations of the real tasks to those in simulation.

\textbf{Workspace and Setup.} Our physical robot workspace consists of a Franka Emika Panda robotic arm, a front-view Intel RealSense SR300 camera, and a wrist-mounted Intel RealSense D415 camera. The robot arm and the front-view camera are attached rigidly to the table, while the wrist-view camera points towards the space in between the robot gripper fingers. Demonstration data was collected from the robot sensors and the two cameras at approximately 20 Hz. Similar to simulation, the robot is controlled using an operational space controller, using the same action space as the one in simulation (see above). 

\textbf{Lift (Real).} A white 3D-printed cube that measures 4 cm in all dimensions was used for this task, with a small initialization square that roughly corresponds to the one for the simulation task.

\textbf{Can (Real).} We purchased \href{https://www.amazon.com/dp/B08BX8Q4J6/ref=cm_sw_em_r_mt_dp_YAEEQJ091HF655034SPV?_encoding=UTF8\&psc=1}{this tray} for the left bin (where the can starts in each episode), and four of \href{https://www.amazon.com/dp/B000QWALJC/ref=cm_sw_em_r_mt_dp_CJ5JNS7A0PS3V00VEPZM}{these small boxes} to construct the right bin.
We used a 7.5 oz Coca Cola Zero Sugar Diet Soda Can (empty, and stuffed with some paper) as the coke can.

\textbf{Tool Hang (Real).} Fig~\ref{fig:task_long} (bottom) shows a full demonstration of the task on the real robot. We purchased \href{https://www.amazon.com/dp/B07QF7WZXH/ref=cm_sw_em_r_mt_dp_0HBCG99TSZW07T0N0VCT}{this handbag stand}, and sawed the base rod and the hook rod in half. We also outfitted the bottom of the hook rod with a soldering iron tip using 2-part epoxy in order to make the hook more amenable to insertion. We also purchased this \href{https://www.amazon.com/dp/B07V43MDN8/ref=cm_sw_em_r_mt_dp_KQGNZ32DJN096ZHMJ6H7}{ratcheting wrench tool set} and used the 17mm-19mm wrench for the task.

\subsection{Observation Space Details}

\begin{figure}[h!]


\centering
\includegraphics[width=0.6\textwidth]{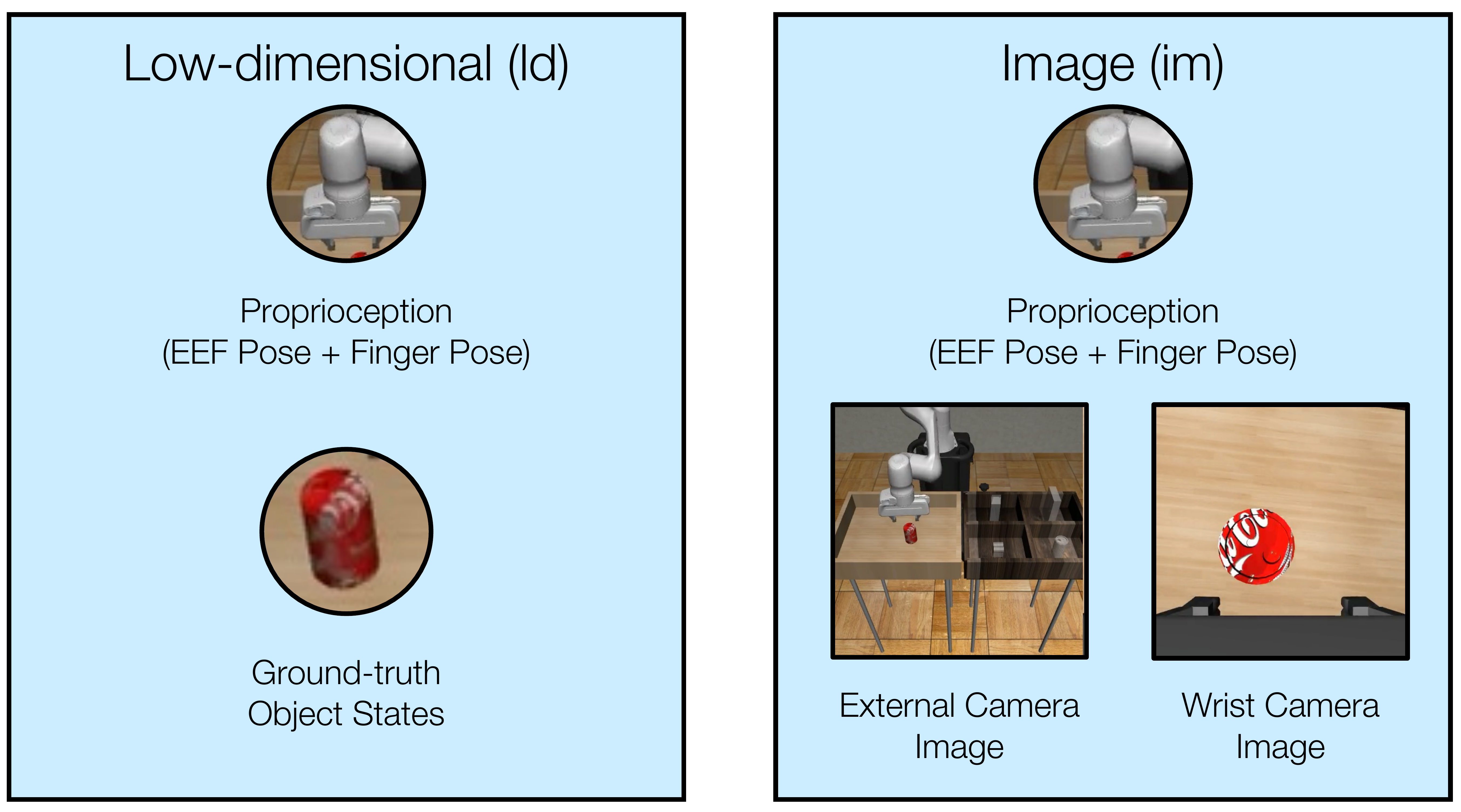}

\caption{\textbf{Observation Spaces.} The figure shows the low-dimensional and image observation spaces used for our study. Proprioception, and camera images are provided for each robot arm in the environment.}
\label{fig:obs_diag}
\end{figure}

In this section, we provide additional details related to the observation spaces used to train agents (shown in Fig~\ref{fig:obs_diag}). Both low-dim agents and image agents receive proprioception observations (9-dim per arm) consisting of the end effector position (3-dim), quaternion (4-dim), and gripper finger positions (2-dim). The low-dim agents also receive object observations (described above, see Appendix~\ref{app:task-sim}), while image agents receive an external camera image and a wrist camera image per robot arm. We first provide further details on cameras and image sizes used per task, then we discuss pixel shift randomization details (shown to be crucial for visuomotor learning, see Sec~\ref{exp:obs} and Fig~\ref{fig:obs}), and finally, we provide more details on the experiments in Sec~\ref{exp:obs}.

\textbf{Image Observations per Task.} For all tasks except Transport, we provide two image observations -- one from a front-view camera and one from a wrist-mounted camera. For the Transport task, we provide four image observations to the agent -- two from shoulder-view cameras per arm, and two from wrist-mounted camera on each arm. The front-view and shoulder-view cameras are the same cameras used by human operators to provide task demonstrations. All simulation tasks, with the exception of Tool Hang, provide 84 by 84 images. All real robot tasks, with the exception of Tool Hang, provide 120 by 120 images. Tool Hang, in both simulation and real world provides 240 by 240 images (due to the need for high-precision control). On the real robot, raw camera frames are read from the camera at a full resolution of 640 by 480, then a center crop of 480 x 480 is applied, and finally, images are resized to the appropriate resolution.

\textbf{Additional Details for Pixel Shift Randomization.} To implement random pixel shifts~\cite{laskin2020reinforcement, kostrikov2020image, young2020visual, zhan2020framework} for input image observations, we take large random crops from the source images when feeding image observations to any network. For each input image of width $W$ and height $H$, we randomly crop a region of width $w$ and height $h$, where $W - w$ and $H - h$ is small. For $(H, W) = (84, 84)$ we use $(h, w) = (76, 76)$, for $(H, W) = (120, 120)$ we use $(h, w) = (108, 108)$ and for $(H, W) = (240, 240)$ we use $(h, w) = (216, 216)$.

\textbf{Additional Details for Observation Space Study.} Here, we describe the observations added for the experiments presented in Sec~\ref{exp:obs} and Fig~\ref{fig:obs}. When adding EEF Vel observations, we added linear end effector velocity (3-dim per arm), angular end effector velocity (3-dim per arm), and gripper finger velocities (2-dim per arm). When adding Joint observations, we encoded the joint positions using cosine (7-dim per arm) and sine (7-dim per arm), and also provided joint velocities (7-dim per arm). 

\newpage
\section{Learning Curves}
\label{app:learning-curves}

In this section, we present learning curves that show success rate versus epoch for BC-RNN on the Proficient-Human (PH) and Multi-Human datasets. Notice that epoch-to-epoch performance can vary drastically, even though the number of evaluation rollouts per checkpoint is high (50), suggesting that this is caused by the mismatch between training and evaluation objectives (C4). See Appendix~\ref{app:policy-selection} for more discussion. 

\begin{figure}[h!]

\begin{subfigure}{0.24\linewidth}
\centering
\includegraphics[width=1.0\textwidth]{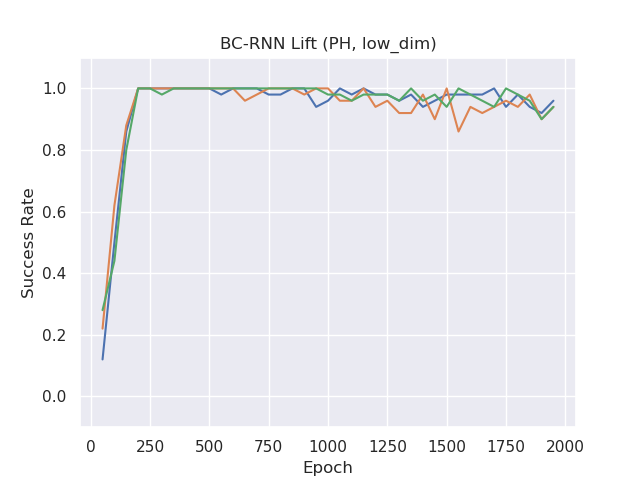}
\caption{Lift (PH, ld)} 
\label{fig:lc-lift-ph-ld}
\end{subfigure}
\hfill
\begin{subfigure}{0.24\linewidth}
\centering
\includegraphics[width=1.0\textwidth]{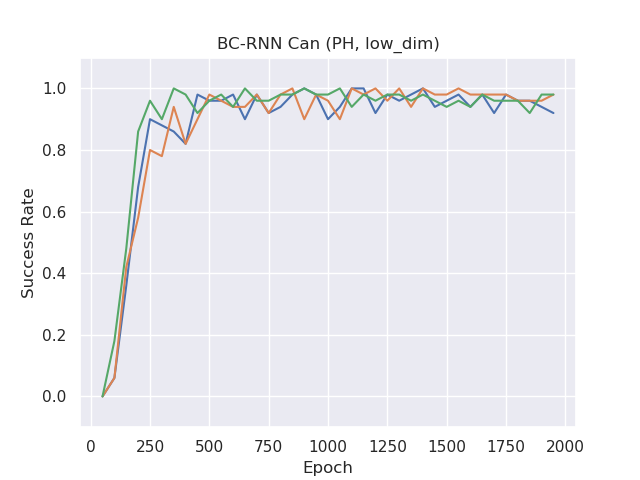}
\caption{Can (PH, ld)} 
\label{fig:lc-can-ph-ld}
\end{subfigure}
\hfill
\begin{subfigure}{0.24\linewidth}
\centering
\includegraphics[width=1.0\textwidth]{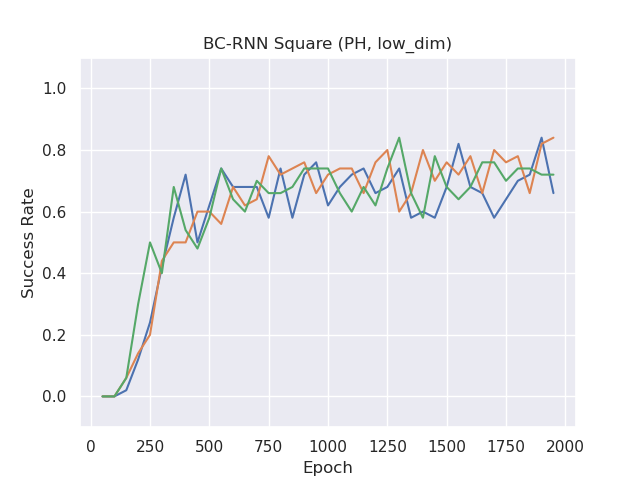}
\caption{Square (PH, ld)} 
\label{fig:lc-square-ph-ld}
\end{subfigure}
\hfill
\begin{subfigure}{0.24\linewidth}
\centering
\includegraphics[width=1.0\textwidth]{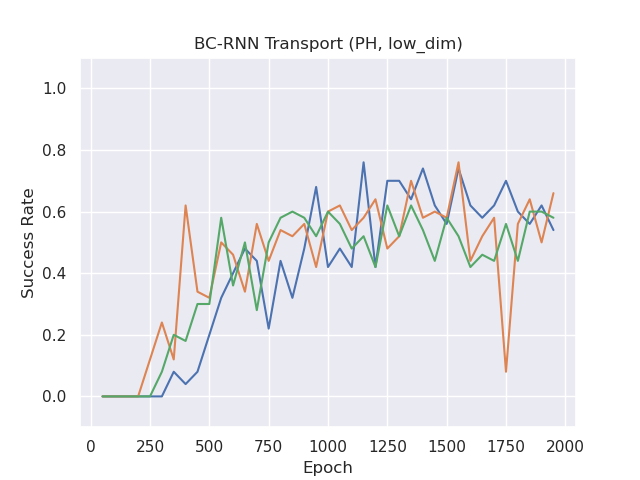}
\caption{Transport (PH, ld)} 
\label{fig:lc-transport-ph-ld}
\end{subfigure}
\hfill

\begin{subfigure}{0.24\linewidth}
\centering
\includegraphics[width=1.0\textwidth]{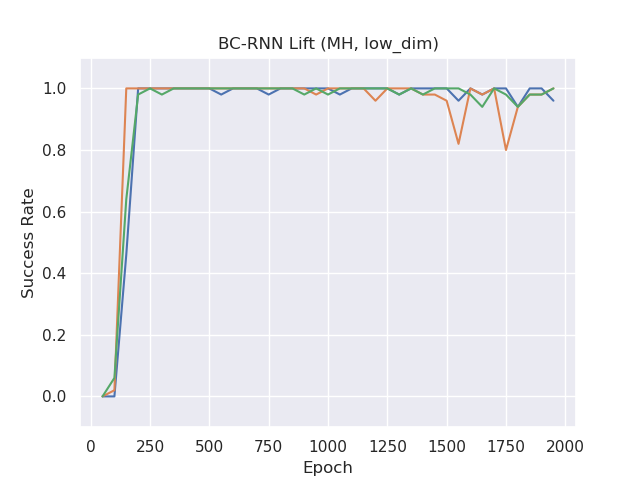}
\caption{Lift (MH, ld)} 
\label{fig:lc-lift-mh-ld}
\end{subfigure}
\hfill
\begin{subfigure}{0.24\linewidth}
\centering
\includegraphics[width=1.0\textwidth]{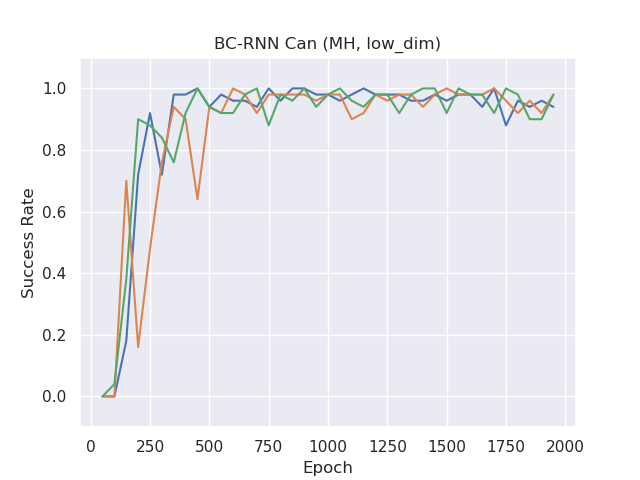}
\caption{Can (MH, ld)} 
\label{fig:lc-can-mh-ld}
\end{subfigure}
\hfill
\begin{subfigure}{0.24\linewidth}
\centering
\includegraphics[width=1.0\textwidth]{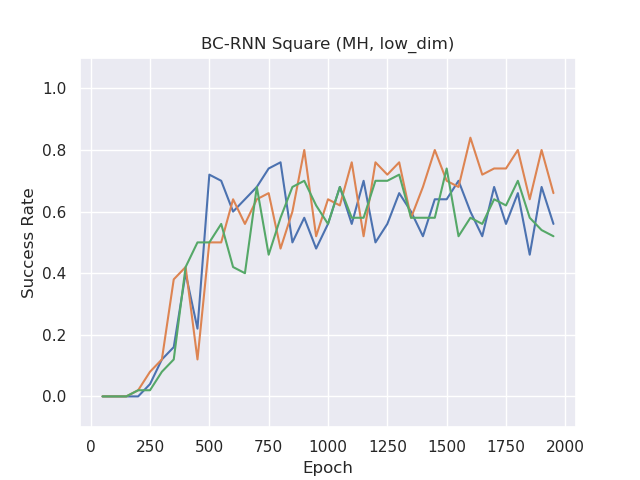}
\caption{Square (MH, ld)} 
\label{fig:lc-square-mh-ld}
\end{subfigure}
\hfill
\begin{subfigure}{0.24\linewidth}
\centering
\includegraphics[width=1.0\textwidth]{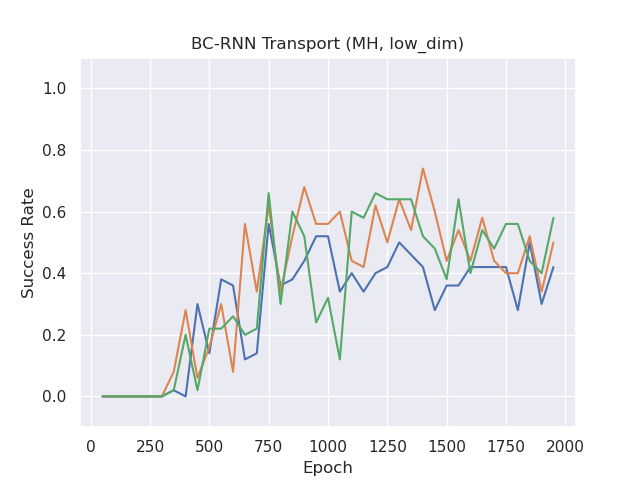}
\caption{Transport (MH, ld)} 
\label{fig:lc-transport-mh-ld}
\end{subfigure}
\hfill

\begin{subfigure}{0.24\linewidth}
\centering
\includegraphics[width=1.0\textwidth]{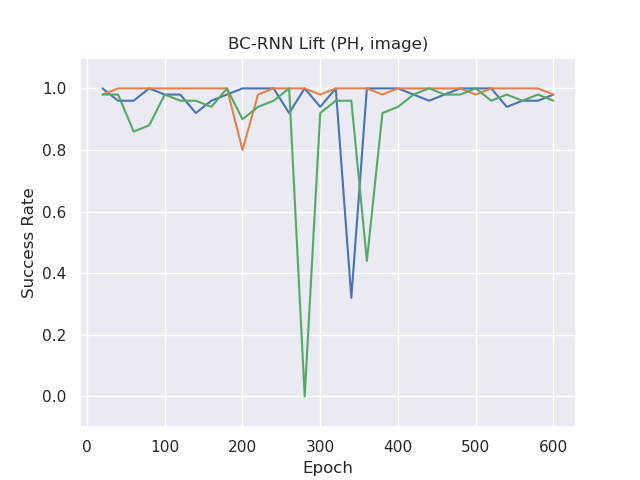}
\caption{Lift (PH, im)} 
\label{fig:lc-lift-ph-image}
\end{subfigure}
\hfill
\begin{subfigure}{0.24\linewidth}
\centering
\includegraphics[width=1.0\textwidth]{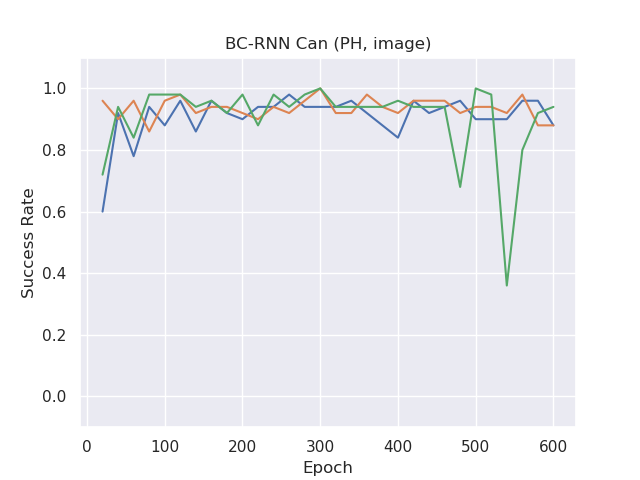}
\caption{Can (PH, im)} 
\label{fig:lc-can-ph-image}
\end{subfigure}
\hfill
\begin{subfigure}{0.24\linewidth}
\centering
\includegraphics[width=1.0\textwidth]{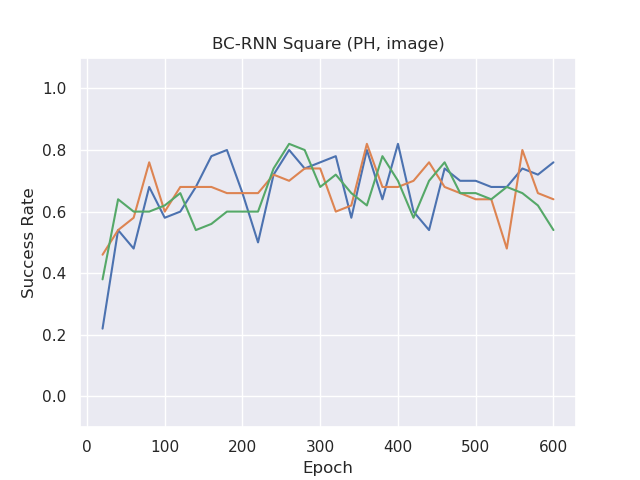}
\caption{Square (PH, im)} 
\label{fig:lc-square-ph-image}
\end{subfigure}
\hfill
\begin{subfigure}{0.24\linewidth}
\centering
\includegraphics[width=1.0\textwidth]{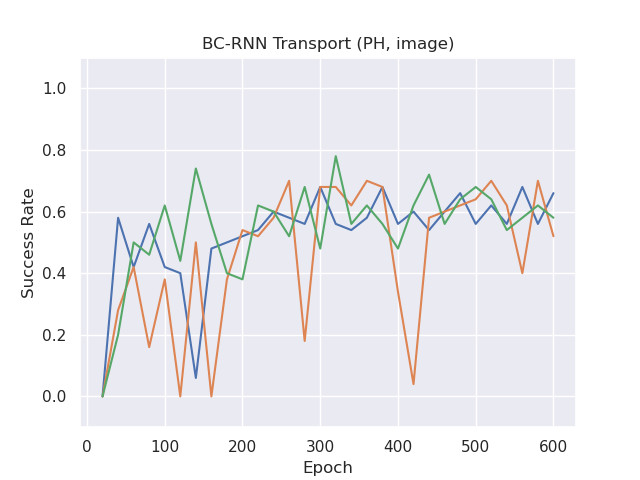}
\caption{Transport (PH, im)} 
\label{fig:lc-transport-ph-image}
\end{subfigure}
\hfill

\begin{subfigure}{0.24\linewidth}
\centering
\includegraphics[width=1.0\textwidth]{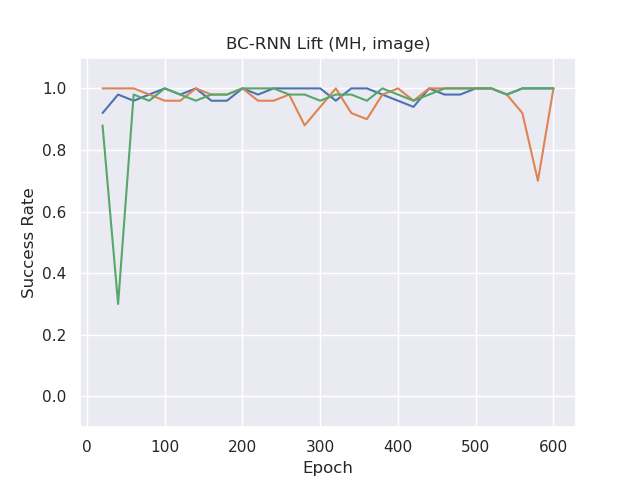}
\caption{Lift (MH, im)} 
\label{fig:lc-lift-mh-image}
\end{subfigure}
\hfill
\begin{subfigure}{0.24\linewidth}
\centering
\includegraphics[width=1.0\textwidth]{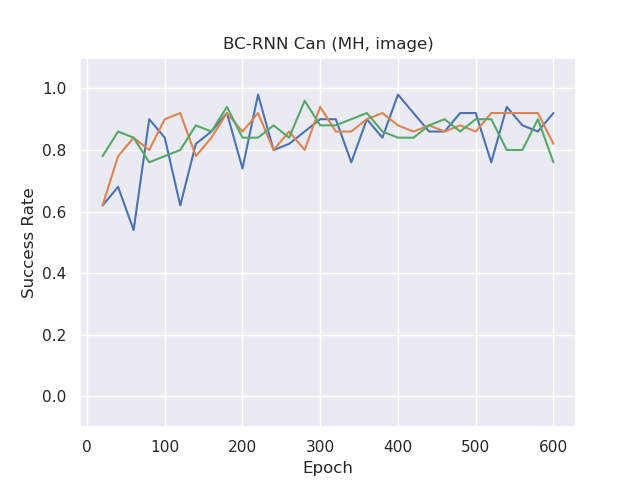}
\caption{Can (MH, im)} 
\label{fig:lc-can-mh-image}
\end{subfigure}
\hfill
\begin{subfigure}{0.24\linewidth}
\centering
\includegraphics[width=1.0\textwidth]{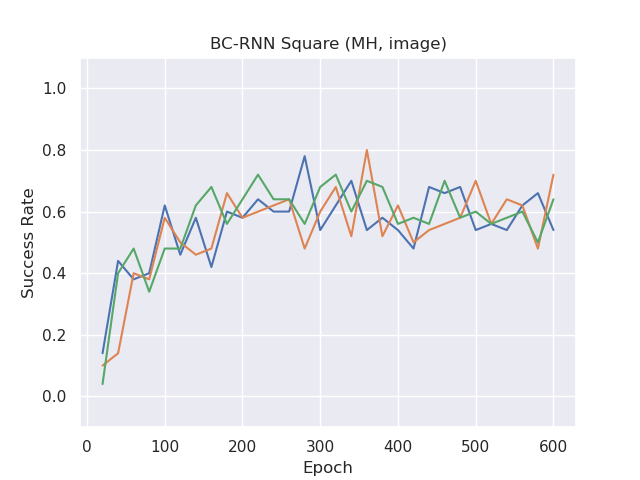}
\caption{Square (MH, im)} 
\label{fig:lc-square-mh-image}
\end{subfigure}
\hfill
\begin{subfigure}{0.24\linewidth}
\centering
\includegraphics[width=1.0\textwidth]{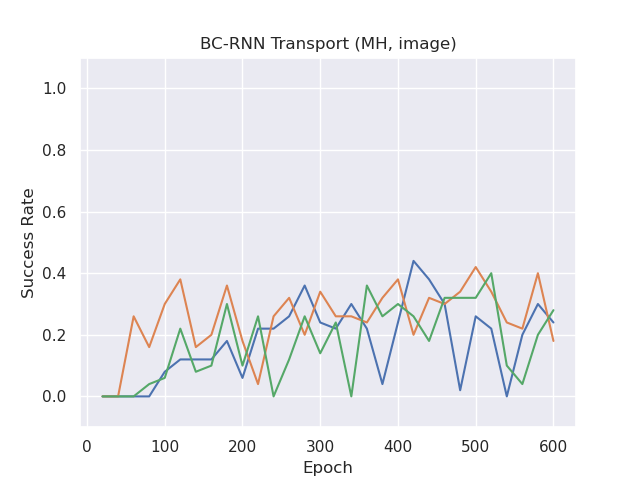}
\caption{Transport (MH, im)} 
\label{fig:lc-transport-mh-image}
\end{subfigure}
\hfill

\caption{\textbf{Learning Curves.} We show the success rate versus epoch for BC-RNN on the Proficient-Human (PH) and Multi-Human (MH) datasets, across 3 seeds. Notice that epoch-to-epoch performance can vary drastically.}
\label{fig:lr-plots}
\vspace{-10pt}
\end{figure}

\newpage
\section{Additional Results on Policy Selection}
\label{app:policy-selection}

In this section, we present some more results and discussion on offline policy selection related to Sec~\ref{exp:mismatch} and Fig~\ref{fig:valid}. We first show that policy performance can vary substantially during a training run -- this is why policy selection is non-trivial. One might expect that adding more validation data could help improve its use as a selection criteria for selecting a good policy. We empirically show that this is not necessarily true. Finally, we show that success rate can keep climbing, even while the validation loss increases substantially.

\textbf{Policy checkpoints can vary substantially in performance during training, even when performance appears to converge.} Fig~\ref{fig:lr-plots} shows several different learning curves for BC-RNN agents on our human datasets. Low-dimensional agents exhibit significant variance in policy success rate even in later stages of training, on harder tasks like Square (see Fig~\ref{fig:lc-square-ph-ld} and Fig~\ref{fig:lc-square-mh-ld}) and Transport (see Fig~\ref{fig:lc-transport-ph-ld} and Fig~\ref{fig:lc-transport-mh-ld}). While this is also true for image agents (see Fig~\ref{fig:lc-square-ph-image} and Fig~\ref{fig:lc-transport-ph-image}), even simpler tasks like Lift (Fig~\ref{fig:lc-lift-ph-image}) and Can (Fig~\ref{fig:lc-can-ph-image}) can suffer from such variance in performance.
This kind of variance in performance is problematic for real world settings where it's not feasible to run 50 rollouts per checkpoint for each training run, as we have done in simulation. This makes offline policy selection a difficult, but important problem to solve. 

\textbf{Increasing the amount of validation data does not improve policy selection using validation loss.} Fig~\ref{fig:valid} used a validation dataset size that was 10\% of the collected data. We also tried training low-dim BC-RNN policies on the Square (PH) and Transport (PH) datasets, where we used 30\% of the collected data for validation (and only 70\% for training). Across 3 seeds, the best policy on Square (PH) achieves $80.7\pm0.9$, while the policy achieving the lowest validation loss achieves $2.7\pm1.9$, and the best policy on Transport (PH) achieves $64.0\pm2.8$ while the policy achieving the lowest validation loss achieves $0.7\pm0.9$.

\begin{figure}[h!]

\begin{subfigure}{0.24\linewidth}
\centering
\includegraphics[width=1.0\textwidth]{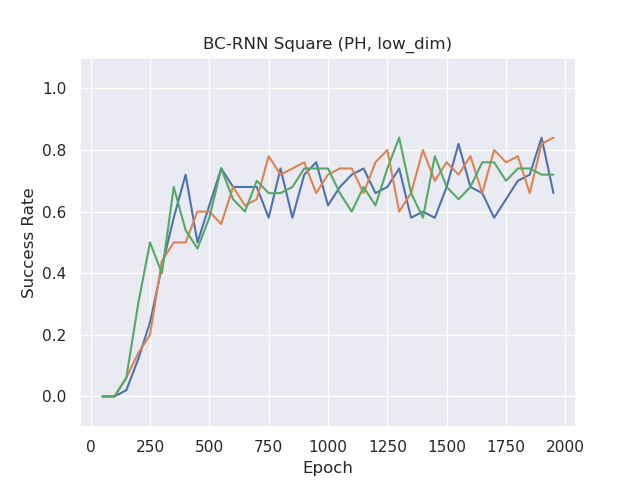}
\caption{Square (ld) -- SR} 
\label{fig:lcv-square-succ-ld}
\end{subfigure}
\hfill
\begin{subfigure}{0.24\linewidth}
\centering
\includegraphics[width=1.0\textwidth]{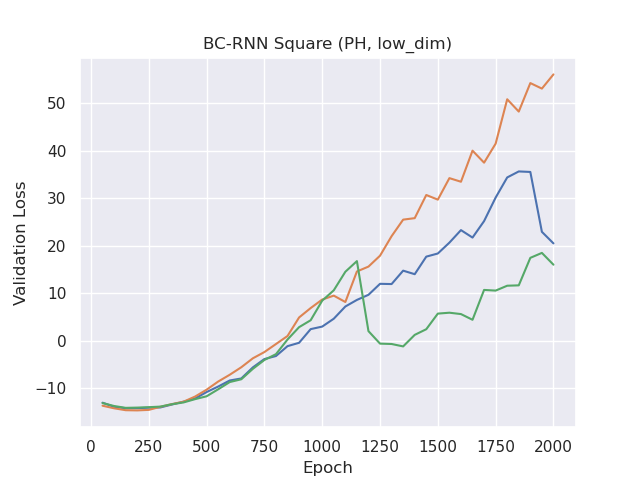}
\caption{Square (ld) -- Loss} 
\label{fig:lcv-square-val-ld}
\end{subfigure}
\hfill
\begin{subfigure}{0.24\linewidth}
\centering
\includegraphics[width=1.0\textwidth]{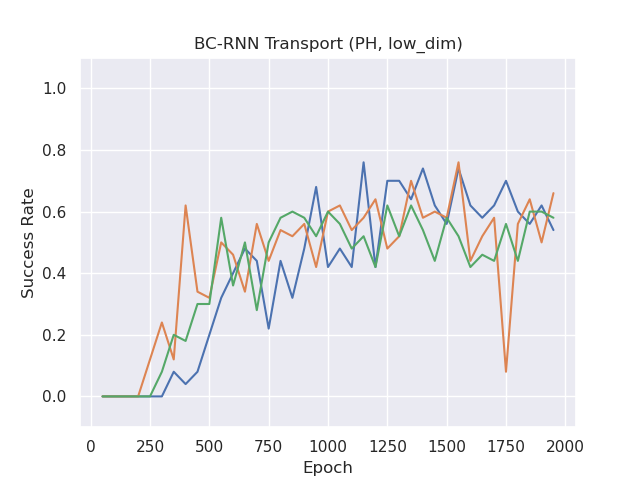}
\caption{Transport (ld) -- SR} 
\label{fig:lcv-transport-succ-ld}
\end{subfigure}
\hfill
\begin{subfigure}{0.24\linewidth}
\centering
\includegraphics[width=1.0\textwidth]{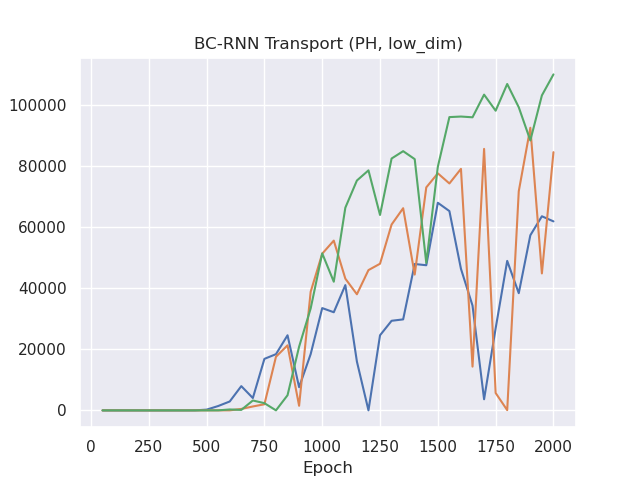}
\caption{Transport (ld) -- Loss} 
\label{fig:lcv-transport-val-ld}
\end{subfigure}
\hfill

\begin{subfigure}{0.24\linewidth}
\centering
\includegraphics[width=1.0\textwidth]{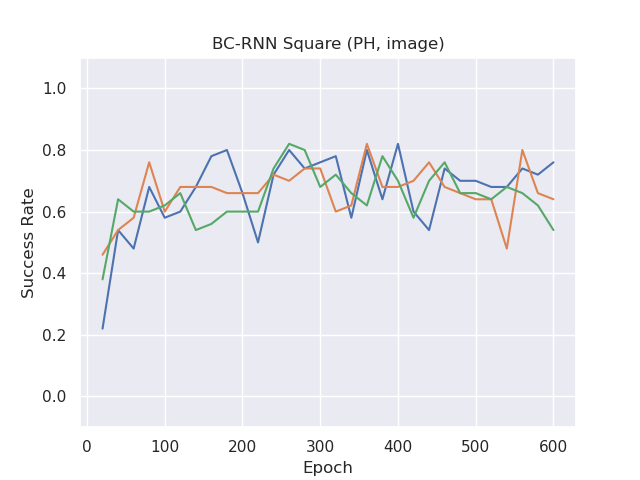}
\caption{Square (im) -- SR} 
\label{fig:lcv-square-succ-im}
\end{subfigure}
\hfill
\begin{subfigure}{0.24\linewidth}
\centering
\includegraphics[width=1.0\textwidth]{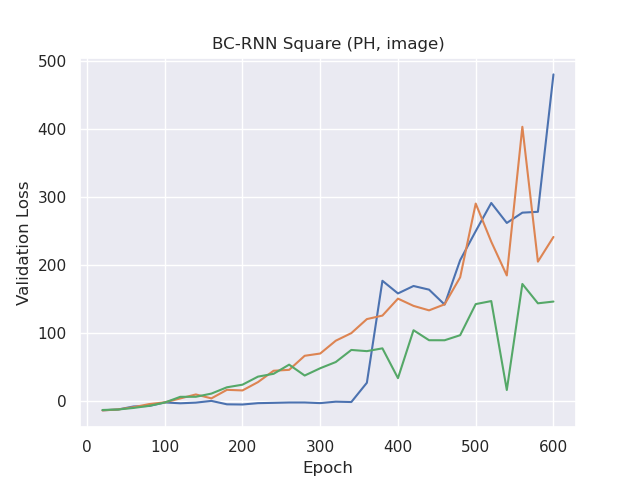}
\caption{Square (im) -- Loss} 
\label{fig:lcv-square-val-im}
\end{subfigure}
\hfill
\begin{subfigure}{0.24\linewidth}
\centering
\includegraphics[width=1.0\textwidth]{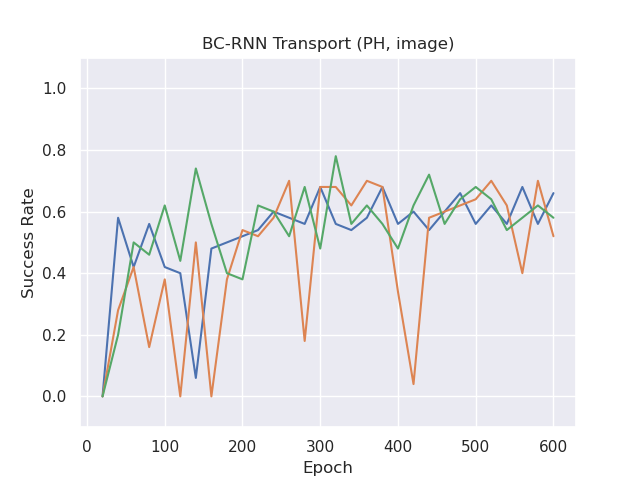}
\caption{Transport (im) -- SR} 
\label{fig:lcv-transport-succ-im}
\end{subfigure}
\hfill
\begin{subfigure}{0.24\linewidth}
\centering
\includegraphics[width=1.0\textwidth]{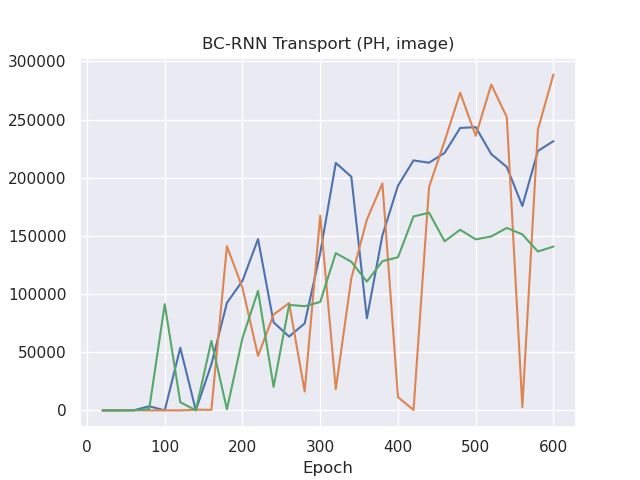}
\caption{Transport (im) -- Loss} 
\label{fig:lcv-transport-val-im}
\end{subfigure}
\hfill


\caption{\textbf{Success Rate and Validation Loss.} We show the success rate versus epoch and validation loss versus epoch side-by-side for BC-RNN on the Square (PH) and Transport (PH) datasets, across 3 seeds. The top row is low-dim observations, and the bottom is image observations. Notice that in many cases, the validation loss increases along with success rate.}
\label{fig:lr-valid-plots}
\vspace{-10pt}
\end{figure}

\textbf{Success rate can increase even while validation loss increases substantially.} Empirically, we found that best validation loss occurs relatively early in training (epoch 100-300) but the best performance occurs much later. In Fig~\ref{fig:lr-valid-plots}, we present selected plots of success rate and validation loss versus epoch, to show this. The plots also show that validation loss can keep increasing substantially in later epochs -- despite this, the success rate also keeps increasing. This further shows that validation loss is a poor measure of policy performance.

\newpage
\section{Additional Image Dataset Results}
\label{app:image}

\subsection{Machine-Generated Datasets}

\begin{table}[h!]
\small
\centering
\begin{tabular}{ccccc}
\toprule
\textbf{Dataset} & 
\begin{tabular}[c]{@{}c@{}}\textbf{BC}\end{tabular} &
\begin{tabular}[c]{@{}c@{}}\textbf{BC-RNN}\end{tabular} & 
\begin{tabular}[c]{@{}c@{}}\textbf{BCQ}\end{tabular} & 
\begin{tabular}[c]{@{}c@{}}\textbf{CQL}\end{tabular} \\
\midrule

Lift (MG) & $79.3\pm6.6$ & $81.3\pm5.7$ & $\mathbf{88.7\pm6.6}$ & $2.7\pm0.9$ \\
Can (MG) & $60.7\pm2.5$ & $63.3\pm2.5$ & $\mathbf{65.3\pm2.5}$ & $0.0\pm0.0$ \\

\bottomrule
\end{tabular}

\vspace{+3pt}
\caption{\textbf{Machine Generated Results (image).} We present success rates averaged over 3 seeds for each method across the image Machine-Generated (MG) datasets. BCQ outperforms the other methods; however, it is possible that recent batch RL methods~\cite{wang2020critic, yu2021combo} that have been shown to work on pixel observations might be able to perform even better.}
\label{table:rb_image}

\end{table}

Table~\ref{table:rb_image} shows results on the Machine-Generated (MG) datasets with image observations. BCQ outperforms the other methods; however, it is possible that recent batch RL methods~\cite{wang2020critic, yu2021combo} that have been shown to work on pixel observations might be able to perform even better.

\subsection{Suboptimal Human Datasets}

\begin{table}[h!]
\small
\centering
\begin{tabular}{ccccc}
\toprule
\textbf{Dataset} & 
\begin{tabular}[c]{@{}c@{}}\textbf{BC}\end{tabular} &
\begin{tabular}[c]{@{}c@{}}\textbf{BC-RNN}\end{tabular} &
\begin{tabular}[c]{@{}c@{}}\textbf{BCQ}\end{tabular} &
\begin{tabular}[c]{@{}c@{}}\textbf{CQL}\end{tabular} \\
\midrule

Can-Worse & $54.7\pm2.5$ & $\mathbf{70.0\pm3.3}$ & - & - \\
Can-Okay & $85.3\pm0.9$ & $\mathbf{90.0\pm3.3}$ & - & - \\
Can-Better & $\mathbf{96.0\pm0.0}$ & $\mathbf{96.0\pm2.8}$ & - & - \\
\midrule
Can-Worse-Okay & $72.7\pm1.9$ & $\mathbf{94.0\pm1.6}$ & - & - \\
Can-Worse-Better & $84.0\pm2.8$ & $\mathbf{92.7\pm1.9}$ & - & - \\
Can-Okay-Better & $94.7\pm0.9$ & $\mathbf{98.0\pm0.0}$ & - & - \\
\midrule
Square-Worse & $17.3\pm1.9$ & $\mathbf{36.7\pm0.9}$ & - & - \\
Square-Okay & $28.7\pm5.0$ & $\mathbf{44.0\pm1.6}$ & - & - \\
Square-Better & $49.3\pm1.9$ & $\mathbf{60.0\pm2.8}$ & - & - \\
\midrule
Square-Worse-Okay & $28.7\pm2.5$ & $\mathbf{52.7\pm6.2}$ & - & - \\
Square-Worse-Better & $38.7\pm0.9$ & $\mathbf{57.3\pm3.4}$ & - & - \\
Square-Okay-Better & $47.3\pm2.5$ & $\mathbf{62.0\pm5.7}$ & - & - \\
\midrule
Can-Paired & $56.7\pm0.9$ & $\mathbf{62.0\pm2.8}$ & $50.0\pm5.9$ & $0.0\pm0.0$ \\

\bottomrule
\end{tabular}
\vspace{+3pt}
\caption{\textbf{Results on Suboptimal Human Data (Image).} We present success rates averaged over 3 seeds for each method across different subsets of the Multi-Human datasets, corresponding to mixtures of demonstrations from ``Better'', ``Adequate'', and ``Worse'' human operators, and finally on a diagnostic dataset with paired success and failure human trajectories for each starting initialization. We omitted Batch RL methods (except for Can-Paired) due to their poor performance on the other human datasets with image observations.}
\label{table:subopt-image}
\end{table}

Table~\ref{table:subopt-image} shows results on our multi-human data subsets with image observations. We excluded batch RL methods due to their poor performance on human datasets with image observations. BC-RNN improves over BC on all datasets, especially datasets with lower quality data.

\newpage
\section{Additional Batch (Offline) RL Results}
\label{app:offline-rl}





\subsection{Hyperparameter Sensitivity}

\subsubsection{BCQ}
\begin{table}[h!]
\small
\centering
\begin{tabular}{ccc}
\toprule
\textbf{Dataset} & 
\begin{tabular}[c]{@{}c@{}}\textbf{Default}\end{tabular} &
\begin{tabular}[c]{@{}c@{}}\textbf{Perturbation}\\\textbf{Actor}\end{tabular} \\
\midrule

Lift (PH) & $100.0\pm0.0$ & $72.0\pm4.3$ \\
Can (PH) & $88.7\pm0.9$ & $8.0\pm4.3$ \\
Square (PH) & $50.0\pm4.9$ & $3.3\pm0.9$ \\
Transport (PH) & $7.3\pm3.3$ & $0.0\pm0.0$ \\

\bottomrule
\end{tabular}

\vspace{+3pt}
\caption{\textbf{BCQ Hyperparameter Sensitivity - Actor.} The perturbation actor causes large performance drops on human datasets.}
\label{table:hyper-bcq-actor}

\end{table}

\begin{table}[h!]
\small
\centering
\begin{tabular}{cccc}
\toprule
\textbf{Dataset} & 
\begin{tabular}[c]{@{}c@{}}\textbf{Default}\\\textbf{(BCQ)}\end{tabular} &
\begin{tabular}[c]{@{}c@{}}\textbf{Default}\\\textbf{(BC)}\end{tabular} &
\begin{tabular}[c]{@{}c@{}}\textbf{BCQ}\\\textbf{(BC param)}\end{tabular} \\
\midrule

Can (PH) & $88.7\pm0.9$ & $95.3\pm0.9$ & $32.0\pm1.6$ \\
Square (PH) & $50.0\pm4.9$ & $78.7\pm1.9$ & $22.7\pm6.6$ \\
Can (MH) & $62.7\pm8.2$ & $86.0\pm4.3$ & $12.0\pm2.8$ \\
Square (MH) & $14.0\pm4.3$ & $52.7\pm6.6$ & $4.0\pm0.0$ \\

\bottomrule
\end{tabular}

\vspace{+3pt}
\caption{\textbf{BCQ Hyperparameter Sensitivity - matching parameters to BC.} We find that matching the hyperparameters of the BCQ action sampler to the ones we used for BC is not sufficient to improve performance.}
\label{table:hyper-bcq-gmm}

\end{table}

In Table~\ref{table:hyper-bcq-actor}, we show that using the BCQ perturbation actor (see Appendix~\ref{app:algorithm-bcq}) can have a catastrophic effect when training on human datasets. The results show that there is a large performance drop after enabling the perturbation actor (over $80\%$ on Can-PH, for example). This result showcases BCQ is highly sensitive to the perturbation actor.

In Sec~\ref{exp:human} and Sec~\ref{exp:suboptimal}, we empirically saw BCQ consistently underperform compared to BC. To further investigate this problem, we tried matching the hyperparameters of the BCQ action sampler to the BC model we used (the same learning rate, MLP architecture, and using a Gaussian Mixture Model). We present the results in Table~\ref{table:hyper-bcq-gmm}. BCQ still underperforms -- since the only difference between BC and this version of BCQ at test-time is selecting a GMM action sample uniformly at random versus using the Q-function to select one, this indicates that the Q-function is responsible for poor performance.




\subsubsection{CQL}
\begin{table}[h!]
\small
\centering
\begin{tabular}{cccccc}
\toprule
\textbf{Dataset} & 
\begin{tabular}[c]{@{}c@{}}\textbf{Default}\end{tabular} &
\begin{tabular}[c]{@{}c@{}}\textbf{smaller}\\\textbf{LR}\end{tabular} & 
\begin{tabular}[c]{@{}c@{}}\textbf{no}\\\textbf{DBackup}\end{tabular} & 
\begin{tabular}[c]{@{}c@{}}\textbf{smaller}\\\textbf{Batch Size}\end{tabular} & 
\begin{tabular}[c]{@{}c@{}}\textbf{no}\\\textbf{Lagrange}\end{tabular} \\ 
\midrule

Lift (MG) & $\mathbf{64.0\pm2.8}$ & $30.0\pm13.4$ & $44.0\pm15.0$ & $36.7\pm9.0$ & $8.0\pm7.1$ \\
Lift (PH) & $\mathbf{92.7\pm5.0}$ & $21.3\pm5.2$ & $\mathbf{90.7\pm5.0}$ & $60.7\pm36.5$ & $\mathbf{90.7\pm3.8}$ \\

\bottomrule
\end{tabular}

\vspace{+3pt}
\caption{\textbf{CQL Hyperparameter Sensitivity.} We find that CQL (1) is highly sensitive to learning, (2) benefits greatly from larger batch sizes, and (3) benefits from the deterministic backup and Lagrange variants with significant improvements on some datasets.}
\label{table:hyper-cql}

\end{table}
We investigate the effect of various hyperparameters on CQL in Table~\ref{table:hyper-cql} for the low-dim Lift MG and PH datasets.
First, we see that a smaller learning rate (specifically $10\times$ smaller) for the Q and policy networks leads to a decrease in success rate of over 50\%, indicating that CQL is highly sensitive to learning rate.
Next, we see that a smaller batch size (specifically $100$ instead of $1024$) leads to a performance decrease of over 30\%, indicating the CQL can greatly benefit from higher batch sizes.
The results for excluding the Lagrange variant and the deterministic backup also indicate that these components can help, with substantial improvements in performance for the MG dataset yet marginal improvements for the PH dataset.

\newpage
\section{GMM Policy Details}
\label{app:gmm}


\begin{table}[h!]
\small
\centering
\begin{tabular}{ccc}
\toprule
\textbf{Dataset} & 
\begin{tabular}[c]{@{}c@{}}\textbf{Default}\end{tabular} &
\begin{tabular}[c]{@{}c@{}}\textbf{No}\\\textbf{Low Noise Eval}\end{tabular} \\
\midrule

Square (PH, ld) & $\mathbf{84.0\pm0.0}$ & $\mathbf{84.0\pm3.3}$ \\
Transport (PH, ld) & $\mathbf{71.3\pm6.6}$ & $64.7\pm5.2$ \\
Square (MH, ld) & $\mathbf{78.0\pm4.3}$ & $\mathbf{79.3\pm3.4}$ \\
Transport (MH, ld) & $\mathbf{65.3\pm7.4}$ & $54.7\pm4.1$ \\

\midrule

Square (PH, im) & $\mathbf{82.0\pm0.0}$ & $\mathbf{78.0\pm4.3}$ \\
Transport (PH, im) & $\mathbf{72.0\pm4.3}$ & $\mathbf{71.3\pm1.9}$ \\
Square (MH, im) & $\mathbf{76.7\pm3.4}$ & $68.0\pm0.0$ \\
Transport (MH, im) & $\mathbf{42.0\pm1.6}$ & $\mathbf{43.3\pm2.5}$ \\

\bottomrule
\end{tabular}

\vspace{+3pt}
\caption{\textbf{GMM Low Noise Evaluation Trick.} This table shows the effect of sampling from the GMM instead of using the low-noise-eval trick that we used by default.}
\label{table:gmm-lne}

\end{table}

As in Acme~\cite{hoffman2020acme}, when using Gaussian Mixture Model (GMM) policies during rollouts, we ignore the learned standard deviations of each mode and instead set it to 1e-4. This amounts to sampling one of the GMM modes instead of sampling from the full GMM distribution. A similar trick is often used when learning Gaussian policies, where at test-time, the mean action of the distribution is used instead of sampling from the distribution. In the table above, we present an ablation for not using this "low-noise-evaluation" (LNE) trick. In most cases, the performance decreases slightly. In early experiments (where the minimum learned std for each Gaussian mode was set to 1e-2 instead of 1e-4), this trick made a more substantial difference. We suggest using this trick by default in all experiments involving GMM policies.
\newpage
\section{Additional Results on Multi-Human Datasets}
\label{app:mh-datasets}

This section contains additional results on Multi-Human datasets that were excluded from the main text for space reasons. This includes conducting the Observation Space study on Multi-Human datasets (the main text included results on the Single-Human datasets in Fig.~\ref{fig:obs} and Table~\ref{table:obs}), and results on the Lift and Transport Multi-Human data subsets (the main text only included results on the low-dim Can and Square subsets in Table~\ref{table:subopt}, and the image subsets in Appendix~\ref{app:image}).

\begin{table}[h!]
\small
\centering
\begin{tabular}{cccccc}
\toprule
\textbf{Dataset} & 
\begin{tabular}[c]{@{}c@{}}\textbf{Default}\end{tabular} &
\begin{tabular}[c]{@{}c@{}}\textbf{+ EEF Vel}\end{tabular} & 
\begin{tabular}[c]{@{}c@{}}\textbf{+ Joint}\end{tabular} & 
\begin{tabular}[c]{@{}c@{}}\textbf{- Rand}\end{tabular} & 
\begin{tabular}[c]{@{}c@{}}\textbf{- Wrist}\end{tabular} \\ 
\midrule

Square(MH, ld) & $78.0\pm4.3$ & $16.0\pm3.3$ & $15.3\pm0.9$ & - & - \\
Transport(MH, ld) & $65.3\pm7.4$ & $2.0\pm0.0$ & $2.0\pm0.0$ & - & - \\
Square(MH, im) & $76.7\pm3.4$ & $46.7\pm2.5$ & $47.3\pm4.1$ & $29.3\pm3.8$ & $59.3\pm3.4$ \\
Transport(MH, im) & $42.0\pm1.6$ & $10.0\pm3.3$ & $16.7\pm0.9$ & $18.0\pm1.6$ & $28.7\pm6.8$ \\

\bottomrule
\end{tabular}

\vspace{+3pt}
\caption{\textbf{Observation Space Study (Multi-Human).} This table presents the same observation space study as conducted in Fig~\ref{fig:obs}, but on the multi-human datasets instead of the proficient-human datasets. The results and conclusions are consistent.}
\label{table:obs-comp}

\end{table}
\begin{table}[h!]
\centering
\resizebox{0.8\linewidth}{!}{
\begin{tabular}{ccccccc}
\toprule
\textbf{Dataset} & 
\begin{tabular}[c]{@{}c@{}}\textbf{BC}\end{tabular} &
\begin{tabular}[c]{@{}c@{}}\textbf{BC-RNN}\end{tabular} & 
\begin{tabular}[c]{@{}c@{}}\textbf{BCQ}\end{tabular} & 
\begin{tabular}[c]{@{}c@{}}\textbf{CQL}\end{tabular} & 
\begin{tabular}[c]{@{}c@{}}\textbf{HBC}\end{tabular} & 
\begin{tabular}[c]{@{}c@{}}\textbf{IRIS}\end{tabular} \\ 
\midrule

Lift-Worse & $100.0\pm0.0$ & $100.0\pm0.0$ & $97.3\pm0.9$ & $13.3\pm9.0$ & $100.0\pm0.0$ & $100.0\pm0.0$ \\
Lift-Okay & $96.0\pm1.6$ & $100.0\pm0.0$ & $100.0\pm0.0$ & $67.3\pm10.5$ & $100.0\pm0.0$ & $100.0\pm0.0$ \\
Lift-Better & $98.7\pm1.9$ & $100.0\pm0.0$ & $98.0\pm1.6$ & $88.0\pm5.9$ & $100.0\pm0.0$ & $100.0\pm0.0$ \\
\midrule
Lift-Worse-Okay & $98.7\pm1.9$ & $100.0\pm0.0$ & $100.0\pm0.0$ & $64.7\pm2.5$ & $99.3\pm0.9$ & $100.0\pm0.0$ \\
Lift-Worse-Better & $100.0\pm0.0$ & $100.0\pm0.0$ & $98.7\pm0.9$ & $75.3\pm25.6$ & $100.0\pm0.0$ & $100.0\pm0.0$ \\
Lift-Okay-Better & $99.3\pm0.9$ & $100.0\pm0.0$ & $100.0\pm0.0$ & $86.0\pm6.5$ & $100.0\pm0.0$ & $100.0\pm0.0$ \\
\midrule
Transport-Worse-Worse & $0.6\pm0.9$ & $4.7\pm0.9$ & $0.0\pm0.0$ & $0.0\pm0.0$ & $4.0\pm1.6$ & $6.0\pm0.0$ \\
Transport-Okay-Okay & $0.7\pm0.9$ & $6.7\pm0.9$ & $0.0\pm0.0$ & $0.0\pm0.0$ & $7.3\pm1.9$ & $7.7\pm1.9$ \\
Transport-Better-Better & $3.3\pm0.9$ & $18.7\pm3.8$ & $2.0\pm0.0$ & $0.0\pm0.0$ & $24.0\pm3.3$ & $22.0\pm3.3$ \\
\midrule
Transport-Worse-Okay & $1.3\pm0.9$ & $5.3\pm0.9$ & $0.0\pm0.0$ & $0.0\pm0.0$ & $4.0\pm0.0$ & $3.3\pm0.9$ \\
Transport-Worse-Better & $7.3\pm0.9$ & $22.7\pm2.5$ & $0.7\pm0.9$ & $0.0\pm0.0$ & $35.3\pm2.5$ & $25.3\pm1.9$ \\
Transport-Okay-Better & $2.7\pm0.9$ & $7.3\pm2.5$ & $0.0\pm0.0$ & $0.0\pm0.0$ & $10.0\pm1.6$ & $11.3\pm3.4$ \\

\bottomrule
\end{tabular}
}
\vspace{+3pt}
\caption{\textbf{Results on Suboptimal Lift and Transport Human Data Subsets.} We present success rates averaged over 3 seeds for each method across different subsets of the Multi-Human datasets, corresponding to mixtures of demonstrations from ``Better'', ``Adequate'', and ``Worse'' human operators.}
\label{table:subopt_comp}
\end{table}
\begin{table}[h!]
\small
\centering
\begin{tabular}{ccccc}
\toprule
\textbf{Dataset} & 
\begin{tabular}[c]{@{}c@{}}\textbf{BC}\end{tabular} &
\begin{tabular}[c]{@{}c@{}}\textbf{BC-RNN}\end{tabular} &
\begin{tabular}[c]{@{}c@{}}\textbf{BCQ}\end{tabular} &
\begin{tabular}[c]{@{}c@{}}\textbf{CQL}\end{tabular} \\
\midrule

Lift-Worse & $98.0\pm0.0$ & $100.0\pm0.0$ & - & - \\
Lift-Okay & $97.3\pm0.9$ & $100.0\pm0.0$ & - & - \\
Lift-Better & $100.0\pm0.0$ & $100.0\pm0.0$ & - & - \\
\midrule
Lift-Worse-Okay & $99.3\pm0.9$ & $100.0\pm0.0$ & - & - \\
Lift-Worse-Better & $100.0\pm0.0$ & $100.0\pm0.0$ & - & - \\
Lift-Okay-Better & $100.0\pm0.0$ & $100.0\pm0.0$ & - & - \\
\midrule
Transport-Worse-Worse & $3.3\pm0.9$ & $4.0\pm0.0$ & - & - \\
Transport-Okay-Okay & $8.7\pm0.9$ & $6.7\pm0.9$ & - & - \\
Transport-Better-Better & $32.0\pm3.7$ & $39.3\pm5.0$ & - & - \\
\midrule
Transport-Worse-Okay & $5.3\pm0.9$ & $4.0\pm0.0$ & - & - \\
Transport-Worse-Better & $21.3\pm4.1$ & $30.7\pm13.6$ & - & - \\
Transport-Okay-Better & $4.7\pm2.5$ & $8.7\pm2.5$ & - & - \\

\bottomrule
\end{tabular}
\vspace{+3pt}
\caption{\textbf{Results on Suboptimal Lift and Transport Human Data Subsets (Image).} We present success rates averaged over 3 seeds for each method across different subsets of the Multi-Human datasets, corresponding to mixtures of demonstrations from ``Better'', ``Adequate'', and ``Worse'' human operators.}
\label{table:subopt-image-comp}
\end{table}

\newpage
\section{Full Tables}
\label{app:full-tables}

This section contains more detailed tables that were excluded in the main text. These tables correspond to the results in Fig~\ref{fig:obs-hyper}, Fig.~\ref{fig:dataset_size}, and Fig~\ref{fig:valid}.

\begin{table}[h!]
\small
\centering
\begin{tabular}{cccccc}
\toprule
\textbf{Dataset} & 
\begin{tabular}[c]{@{}c@{}}\textbf{Default}\end{tabular} &
\begin{tabular}[c]{@{}c@{}}\textbf{+ EEF Vel}\end{tabular} & 
\begin{tabular}[c]{@{}c@{}}\textbf{+ Joint}\end{tabular} & 
\begin{tabular}[c]{@{}c@{}}\textbf{- Rand}\end{tabular} & 
\begin{tabular}[c]{@{}c@{}}\textbf{- Wrist}\end{tabular} \\ 
\midrule

Square (ld) & $84.0\pm0.0$ & $42.7\pm3.4$ & $39.3\pm2.5$ & - & - \\
Transport (ld) & $71.3\pm6.6$ & $8.7\pm0.9$ & $10.0\pm3.3$ & - & - \\
Square (im) & $82.0\pm0.0$ & $64.7\pm0.9$ & $58.0\pm11.4$ & $43.3\pm5.0$ & $74.7\pm3.8$ \\
Transport (im) & $72.0\pm4.3$ & $64.7\pm3.8$ & $70.7\pm2.5$ & $46.7\pm0.9$ & $41.3\pm7.5$ \\

\bottomrule
\end{tabular}

\vspace{+3pt}
\caption{\textbf{Observation Space Study.} This table corresponds to the results presented in Fig~\ref{fig:obs}.}
\label{table:obs}

\end{table}
\begin{table}[h!]
\small
\centering
\begin{tabular}{ccccccc}
\toprule
\textbf{Dataset} & 
\begin{tabular}[c]{@{}c@{}}\textbf{Default}\end{tabular} &
\begin{tabular}[c]{@{}c@{}}\textbf{larger}\\\textbf{LR}\end{tabular} & 
\begin{tabular}[c]{@{}c@{}}\textbf{no}\\\textbf{GMM}\end{tabular} & 
\begin{tabular}[c]{@{}c@{}}\textbf{larger}\\\textbf{MLP}\end{tabular} & 
\begin{tabular}[c]{@{}c@{}}\textbf{Shallow}\\\textbf{Conv}\end{tabular} & 
\begin{tabular}[c]{@{}c@{}}\textbf{smaller}\\\textbf{RNN dim}\end{tabular} \\ 
\midrule

Square (PH, ld) & $84.0\pm0.0$ & $86.0\pm2.8$ & $82.0\pm0.0$ & $82.0\pm0.0$ - & & $81.3\pm0.9$ \\
Transport (PH, ld) & $71.3\pm6.6$ & $64.0\pm5.9$ & $69.3\pm3.4$ & $58.7\pm6.8$ & - &  $47.3\pm2.5$ \\
Square (MH, ld) & $78.0\pm4.3$ & $76.7\pm2.5$ & $58.0\pm1.6$ & $73.3\pm3.4$ & - & $58.7\pm7.4$ \\
Transport (MH, ld) & $65.3\pm7.4$ & $49.3\pm2.5$ & $27.3\pm10.9$ & $46.0\pm3.3$ & - & $27.3\pm12.4$ \\

\midrule

Square (PH, im) & $82.0\pm0.0$ & $41.3\pm7.7$ & $84.0\pm3.3$ & - & $50.0\pm2.8$ & $74.0\pm3.3$ \\
Transport (PH, im) & $72.0\pm4.3$ & $46.7\pm20.4$ & $74.0\pm4.3$ & - & $54.0\pm3.2$ & $65.3\pm5.2$ \\
Square (MH, im) & $76.7\pm3.4$ & $28.7\pm4.1$ & $61.3\pm0.9$ & - & $48.0\pm3.3$ & $58.0\pm4.3$ \\
Transport (MH, im) & $42.0\pm1.6$ & $23.3\pm4.1$ & $41.0\pm0.3$ & - & $16.0\pm0.0$ & $34.0\pm0.0$ \\

\bottomrule
\end{tabular}

\vspace{+3pt}
\caption{\textbf{BC-RNN Hyperparameter Sensitivity.} This table corresponds to the results presented in Fig~\ref{fig:hyper-low-dim} and Fig~\ref{fig:hyper-image}.}
\label{table:hyper}

\end{table}
\begin{table}[h!]
\small
\centering
\begin{tabular}{cccc}
\toprule
\textbf{Dataset} & 
\begin{tabular}[c]{@{}c@{}}\textbf{20\%}\end{tabular} &
\begin{tabular}[c]{@{}c@{}}\textbf{50\%}\end{tabular} & 
\begin{tabular}[c]{@{}c@{}}\textbf{100\%}\end{tabular} \\ 
\midrule

Lift (ld) & $96.7\pm2.5$ & $100.0\pm0.0$ & $100.0\pm0.0$ \\
Can (ld) & $76.7\pm5.2$ & $97.3\pm0.9$ & $100.0\pm0.0$ \\
Square (ld) & $38.7\pm6.2$ & $67.3\pm7.7$ & $84.0\pm0.0$ \\
Transport (ld) & $6.7\pm0.9$ & $44.0\pm5.9$ & $71.3\pm6.6$ \\
\midrule
Lift (im) & $100.0\pm0.0$ & $100.0\pm0.0$ & $100.0\pm0.0$ \\
Can (im) & $83.3\pm1.9$ & $97.3\pm0.9$ & $98.0\pm0.9$ \\
Square (im) & $29.3\pm4.1$ & $64.7\pm4.1$ & $82.0\pm0.0$ \\
Transport (im) & $30.7\pm4.1$ & $60.1\pm4.1$ & $72.0\pm4.3$ \\

\bottomrule
\end{tabular}

\vspace{+3pt}
\caption{\textbf{Proficient-Human Dataset Size Ablation.} This table corresponds to the results presented in Fig~\ref{fig:dataset_size}.}
\label{table:ds_size_low_dim_se}

\end{table}
\begin{table}[h!]
\small
\centering
\begin{tabular}{cccc}
\toprule
\textbf{Dataset} & 
\begin{tabular}[c]{@{}c@{}}\textbf{20\%}\end{tabular} &
\begin{tabular}[c]{@{}c@{}}\textbf{50\%}\end{tabular} & 
\begin{tabular}[c]{@{}c@{}}\textbf{100\%}\end{tabular} \\ 
\midrule

Lift (ld) & $100.0\pm0.0$ & $100.0\pm0.0$ & $100.0\pm0.0$ \\
Can (ld) & $79.3\pm5.0$ & $97.3\pm0.9$ & $100.0\pm0.0$ \\
Square (ld) & $32.0\pm3.3$ & $60.7\pm0.9$ & $78.0\pm4.3$ \\
Transport (ld) & $7.3\pm2.5$ & $33.3\pm7.5$ & $65.3\pm7.4$ \\
\midrule
Lift (im) & $98.0\pm0.0$ & $100.0\pm0.0$ & $100.0\pm0.0$ \\
Can (im) & $77.3\pm2.5$ & $87.3\pm1.9$ & $96.0\pm1.6$ \\
Square (im) & $27.3\pm1.9$ & $50.7\pm3.8$ & $76.7\pm3.4$ \\
Transport (im) & $8.0\pm3.3$ & $25.3\pm3.8$ & $42.0\pm1.6$ \\


\bottomrule
\end{tabular}

\vspace{+3pt}
\caption{\textbf{Multi-Human Dataset Size Ablation.} This table corresponds to the results presented in Fig~\ref{fig:dataset_size}.}
\label{table:ds_size_low_dim_mh}

\end{table}
\begin{table}[h!]
\small
\centering
\begin{tabular}{ccccccc}
\toprule
\textbf{Dataset} & 
\begin{tabular}[c]{@{}c@{}}\textbf{Valid} \\ \textbf{(BC)}\end{tabular} &
\begin{tabular}[c]{@{}c@{}}\textbf{Last} \\ \textbf{(BC)}\end{tabular} & 
\begin{tabular}[c]{@{}c@{}}\textbf{Max} \\ \textbf{(BC)}\end{tabular} & 
\begin{tabular}[c]{@{}c@{}}\textbf{Valid} \\ \textbf{(BC-RNN)}\end{tabular} & 
\begin{tabular}[c]{@{}c@{}}\textbf{Last} \\ \textbf{(BC-RNN)}\end{tabular} & 
\begin{tabular}[c]{@{}c@{}}\textbf{Max} \\ \textbf{(BC-RNN)}\end{tabular} \\ 
\midrule

Square (PH, ld) & $20.0\pm10.2$ & $65.3\pm0.9$ & $\mathbf{78.7\pm1.9}$ & $7.3\pm5.0$ & $74.0\pm7.5$ & $\mathbf{84.0\pm0.0}$ \\
Transport (PH, ld) & $0.0\pm0.0$ & $11.3\pm3.8$ & $\mathbf{17.3\pm2.5}$ & $4.0\pm5.7$ & $59.3\pm5.0$ & $\mathbf{71.3\pm6.6}$ \\
\midrule
Square (PH, im) & $20.6\pm14.0$ & $44.7\pm4.1$ & $\mathbf{62.0\pm4.9}$ & $35.3\pm10.0$ & $64.7\pm9.0$ & $\mathbf{82.0\pm0.0}$ \\
Transport (PH, im) & $16.0\pm12.3$ & $38.7\pm11.5$ & $\mathbf{55.3\pm6.2}$ & $0.0\pm0.0$ & $58.7\pm5.7$ & $\mathbf{72.0\pm4.3}$ \\

\bottomrule
\end{tabular}
\vspace{+3pt}
\caption{\textbf{Effect of Policy Selection Criteria.} We compare how performance decreases when choosing the policy to evaluate by using the lowest validation loss, or when using the final trained checkpoint, compared to the best performing policy. Corresponds to results in Fig~\ref{fig:valid}.}
\label{table:valid-full}
\end{table}

\end{document}